%% file: main.tex
\def\year{2021}\relax
\documentclass[letterpaper]{article} % DO NOT CHANGE THIS
\usepackage{aaai21}  % DO NOT CHANGE THIS
\usepackage{times}  % DO NOT CHANGE THIS
\usepackage{helvet} % DO NOT CHANGE THIS
\usepackage{courier}  % DO NOT CHANGE THIS
\usepackage[hyphens]{url}  % DO NOT CHANGE THIS
\usepackage{graphicx} % DO NOT CHANGE THIS
\usepackage{natbib}  % DO NOT CHANGE THIS AND DO NOT ADD ANY OPTIONS TO IT
\usepackage{caption} % DO NOT CHANGE THIS AND DO NOT ADD ANY OPTIONS TO IT
\nocopyright % -- Your paper will not be published if you use this command
\usepackage{comment}
\usepackage{amsmath} 

\usepackage{epsfig}
\usepackage{algorithmic}
\usepackage{array}

\usepackage{sidecap}

\usepackage[utf8]{inputenc}
\usepackage{subcaption}
\usepackage{multirow}
\usepackage[flushleft]{threeparttable}

\usepackage{lipsum}
\usepackage{mwe}

\usepackage[belowskip=-8pt,aboveskip=13pt]{caption}
\usepackage{amssymb}% http://ctan.org/pkg/amssymb
\usepackage{pifont}

\newcommand{\tablestyle}[2]{\setlength{\tabcolsep}{#1}\renewcommand{\arraystretch}{#2}\centering\footnotesize}

\newcolumntype{x}[1]{>{\centering\arraybackslash}p{#1pt}}
\newcommand{\app}{\raise.17ex\hbox{$\scriptstyle\sim$}}

\newlength\savewidth\newcommand\shline{\noalign{\global\savewidth\arrayrulewidth
  \global\arrayrulewidth 1pt}\hline\noalign{\global\arrayrulewidth\savewidth}}
  
\makeatletter\renewcommand\paragraph{\@startsection{paragraph}{4}{\z@}
  {.5em \@plus1ex \@minus.2ex}{-.5em}{\normalfont\normalsize\bfseries}}\makeatother

\def\fig#1{Fig.~\ref{fig:#1}}

\def\imwh#1#2#3{\includegraphics[clip,width=#2\linewidth,height=#3\textheight]{#1}}

\newcommand{\tb}[3]{\setlength{\tabcolsep}{#2mm}\begin{tabular}{#1}#3\end{tabular}}

{%
\centering\addtocounter{figure}{1}% if caption at bottom
\begin{enumerate}[%
itemsep=2pt,parsep=0em,
label={(\alph*)},
ref={\thefigure.(\alph*)}
]}%

\usepackage[switch]{lineno}
\usepackage[usestackEOL]{stackengine}
\usepackage{color, colortbl}
\definecolor{Gray}{gray}{0.9}

\usepackage{multicol}
\usepackage{lipsum}
\usepackage{mathtools}
\usepackage{cuted}
\title{Tied Block Convolution: \\Leaner and Better CNNs with Shared Thinner Filters}
\author {
        Xudong Wang \textsuperscript{\rm 1,\rm 2} and
        Stella X. Yu \textsuperscript{\rm 1,\rm 2} \\
}
\affiliations {
    \textsuperscript{\rm 1} University of California, Berkeley \\
    \textsuperscript{\rm 2} International Computer Science Institute \\
    xdwang@eecs.berkeley.edu, stellayu@berkeley.edu
}
\begin{document}
% \linenumbers

\maketitle

% \section{Introduction}
\input{0abstract}
\input{1introduction}
\input{2related_work}

\input{3methods}
\input{4experiments}
\input{5summary}
\clearpage

% \begin{strip}
% \centering \huge {\bf{
% Tied Block Convolution: \\
% Leaner and Better CNNs with Shared Thinner Filters}} \\
% Supplementary Materials
% \end{strip}

% \bibliographystyle{Natbib} 
\clearpage
\bibliography{egbib}

\vspace{20pt}
\begin{strip}
\centering \LARGE {\bf{
Tied Block Convolution: \\
Leaner and Better CNNs with Shared Thinner Filters}} \\\vspace{10pt}
Supplementary Materials
\end{strip}
\vspace{20pt}

\input{arxiv_sup}

\end{document}

%% file: 0abstract
\begin{abstract}
Convolution is the main building block of convolutional neural networks (CNN).  We observe that an optimized CNN often has highly correlated filters as the number of channels increases with depth, reducing the expressive power of feature representations.   
We propose {\it Tied Block Convolution} (TBC) that shares the same thinner filters over equal blocks of channels and produces multiple responses with a single filter. The concept of TBC can also be extended to group convolution and fully connected layers, and can be applied to various backbone networks and attention modules. % With ResNet as an example, we apply our TBC concept to individual components of ResNet and design a TiedResNet backbone network.  

Our extensive experimentation on classification, detection, instance segmentation, and attention demonstrates TBC’s significant across-the-board gain over standard convolution and group convolution. The proposed TiedSE attention module can even use 64$\times$ fewer parameters than the SE module to achieve comparable performance.
In particular, standard CNNs often fail to accurately aggregate information in the presence of occlusion and result in multiple redundant partial object proposals.  By sharing filters across channels, TBC reduces correlation and can effectively handle highly overlapping instances.  TBC increases the average precision for object detection on MS-COCO by $6\%$ when the occlusion ratio is 80\%.  Our code will be released.
\end{abstract}

%% file: 1introduction
\section{Introduction}

\def\figConvComp#1{
\begin{figure*}[#1]
\begin{subfigure}{.33\linewidth}
  \centering
  \includegraphics[width=.9\linewidth]{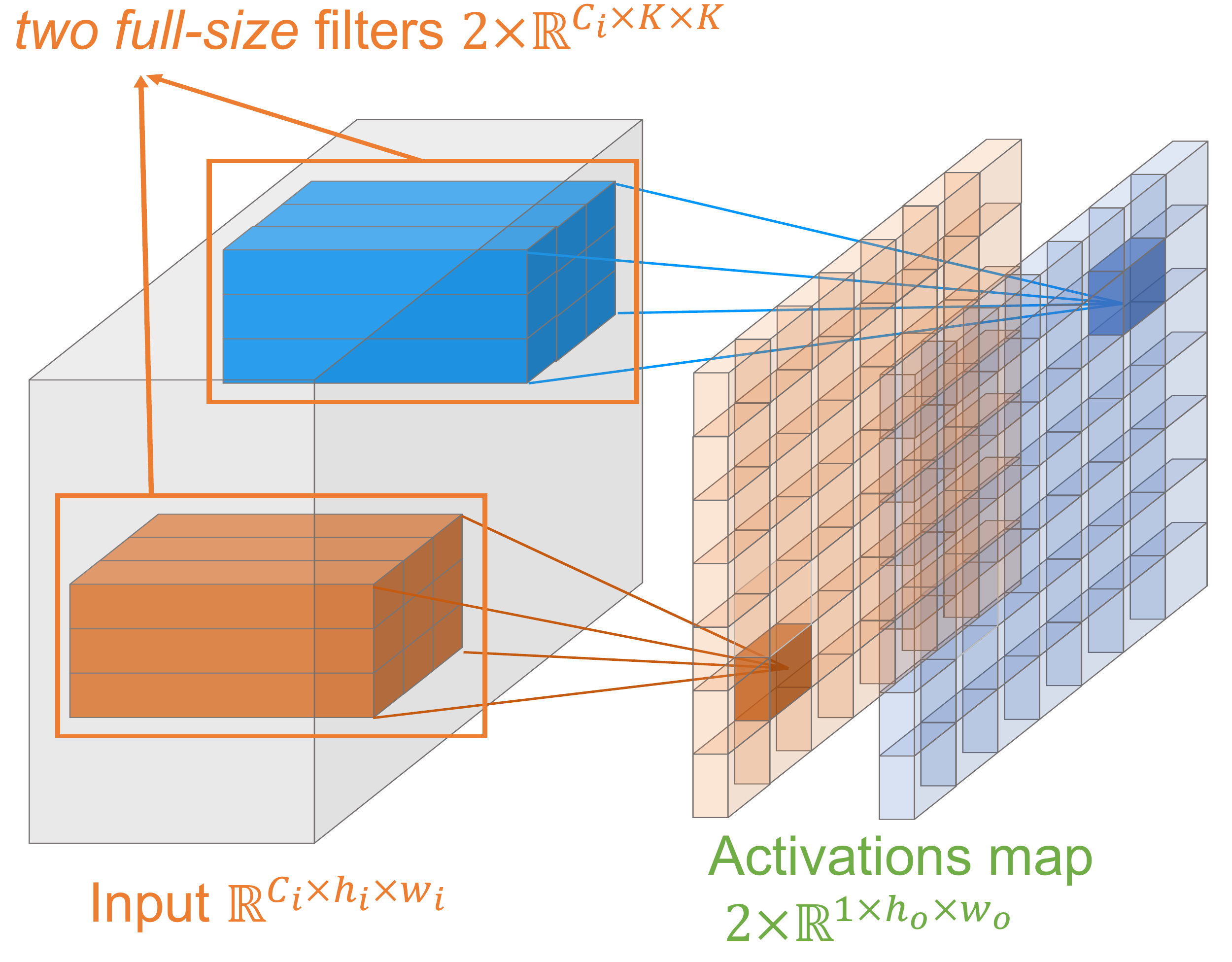}\vspace{-3pt}
  \caption{Standard Convolution}
  \label{fig:sc}
\end{subfigure}%
\begin{subfigure}{.33\linewidth}
  \centering
  \includegraphics[width=.9\linewidth]{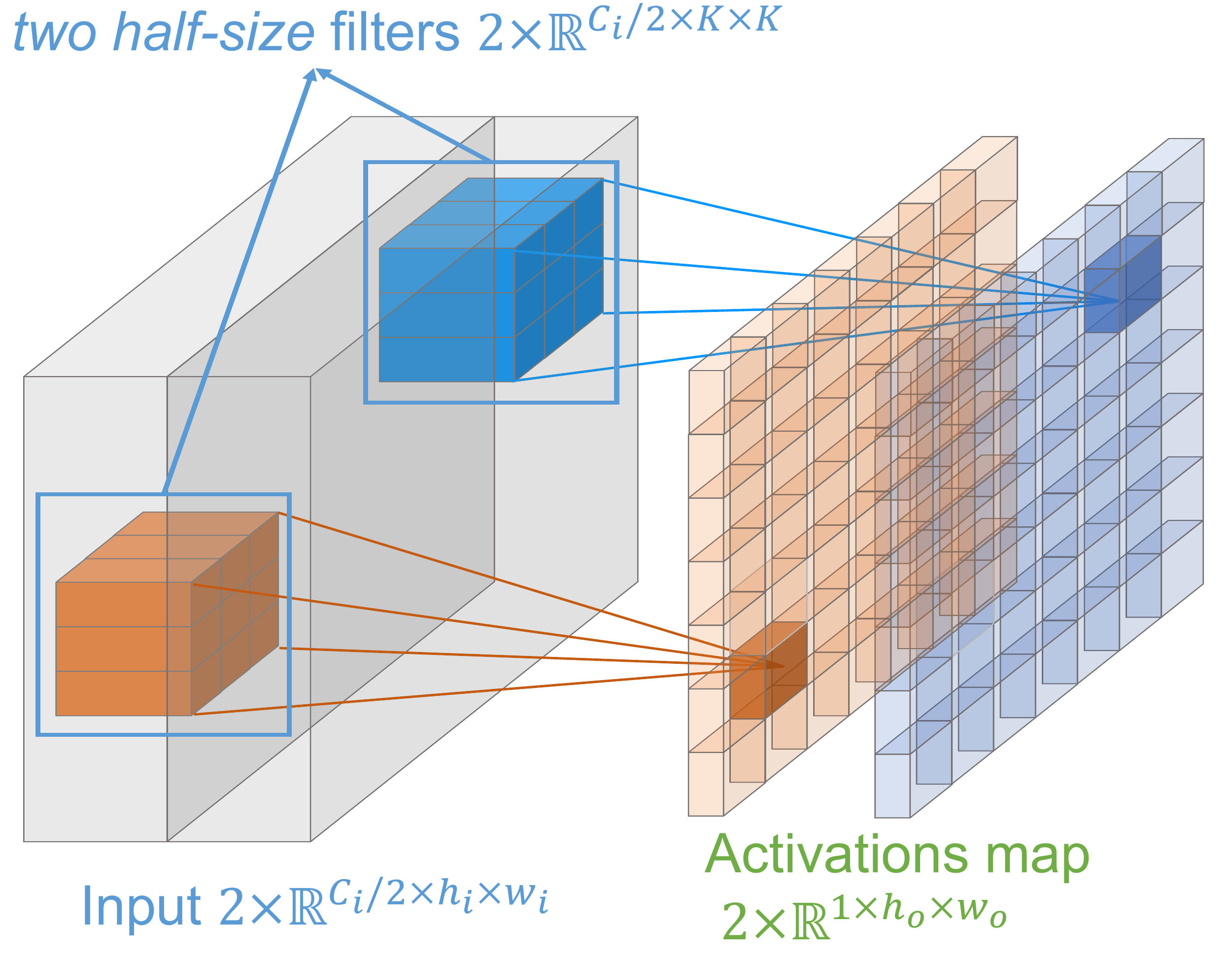}\vspace{-3pt}
  \caption{Group Convolution}
  \label{fig:gc}
\end{subfigure}
\begin{subfigure}{.33\linewidth}
  \centering
  \includegraphics[width=.9\linewidth]{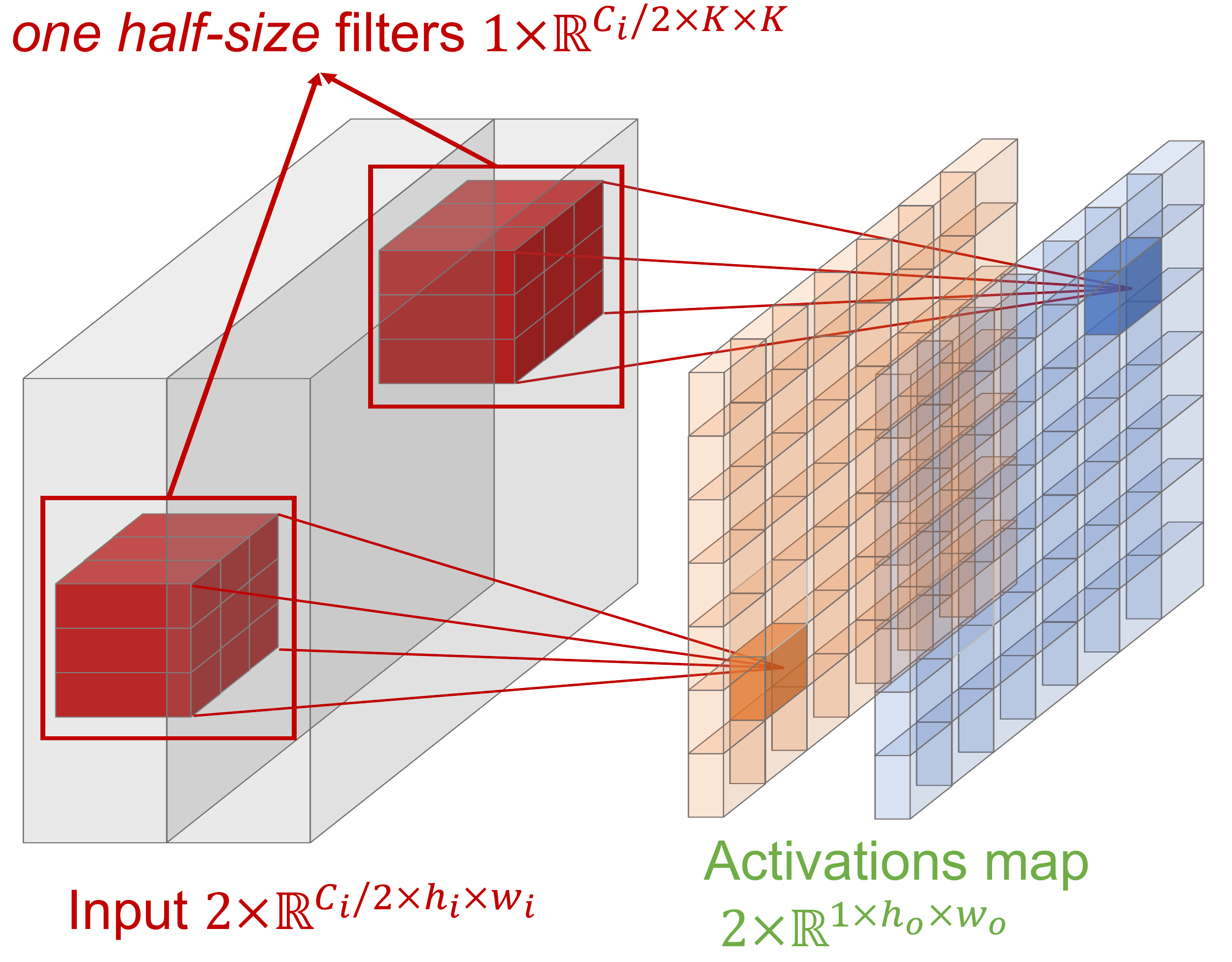}\vspace{-3pt}
  \caption{Tied Block Convolution}
  \label{fig:tbc}
\end{subfigure}\vspace{-4pt}
\caption{\textbf{Convolution operators.} To generate two activation maps, standard convolution requires \textit{two full-size} filters and group convolution requires \textit{two half-size} filters, however, our tied block convolution only requires \textit{one half-size} filter, that is, the parameters are reduced by 4$\times$. The idea of TBC can also be applied to fully connected and group convolutional layers.}
\label{fig:ConvComp}
\end{figure*}
}

\def\fighistvgg#1{
\begin{figure}[#1]
\begin{subfigure}{.69\linewidth}
  \centering
  \includegraphics[width=.99\linewidth]{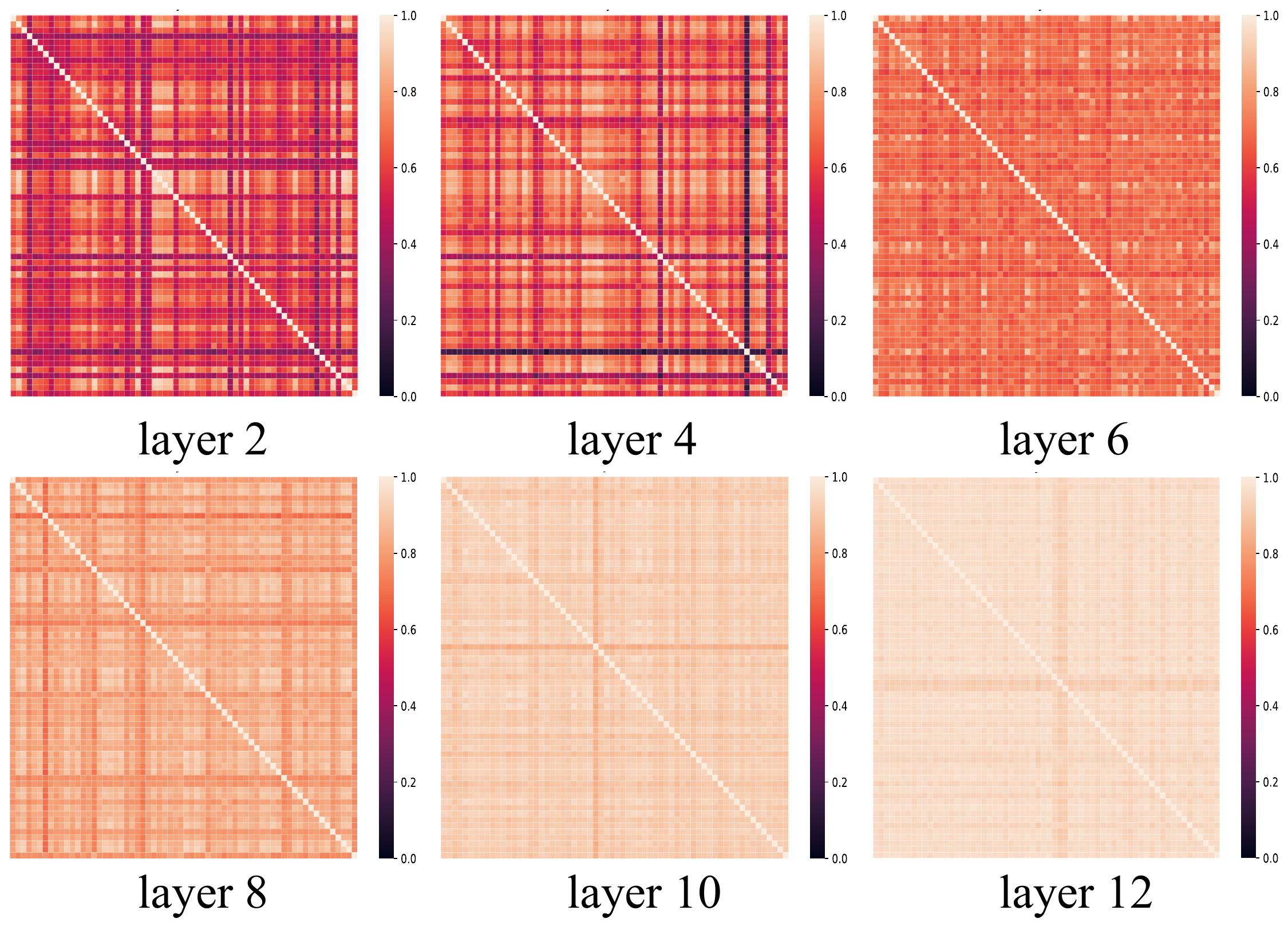}\vspace{-2pt}
  \caption{Correlation matrix}
  \label{fig:sfig2}
\end{subfigure}
\begin{subfigure}{.3\linewidth}
  \centering
  \includegraphics[width=.99\linewidth]{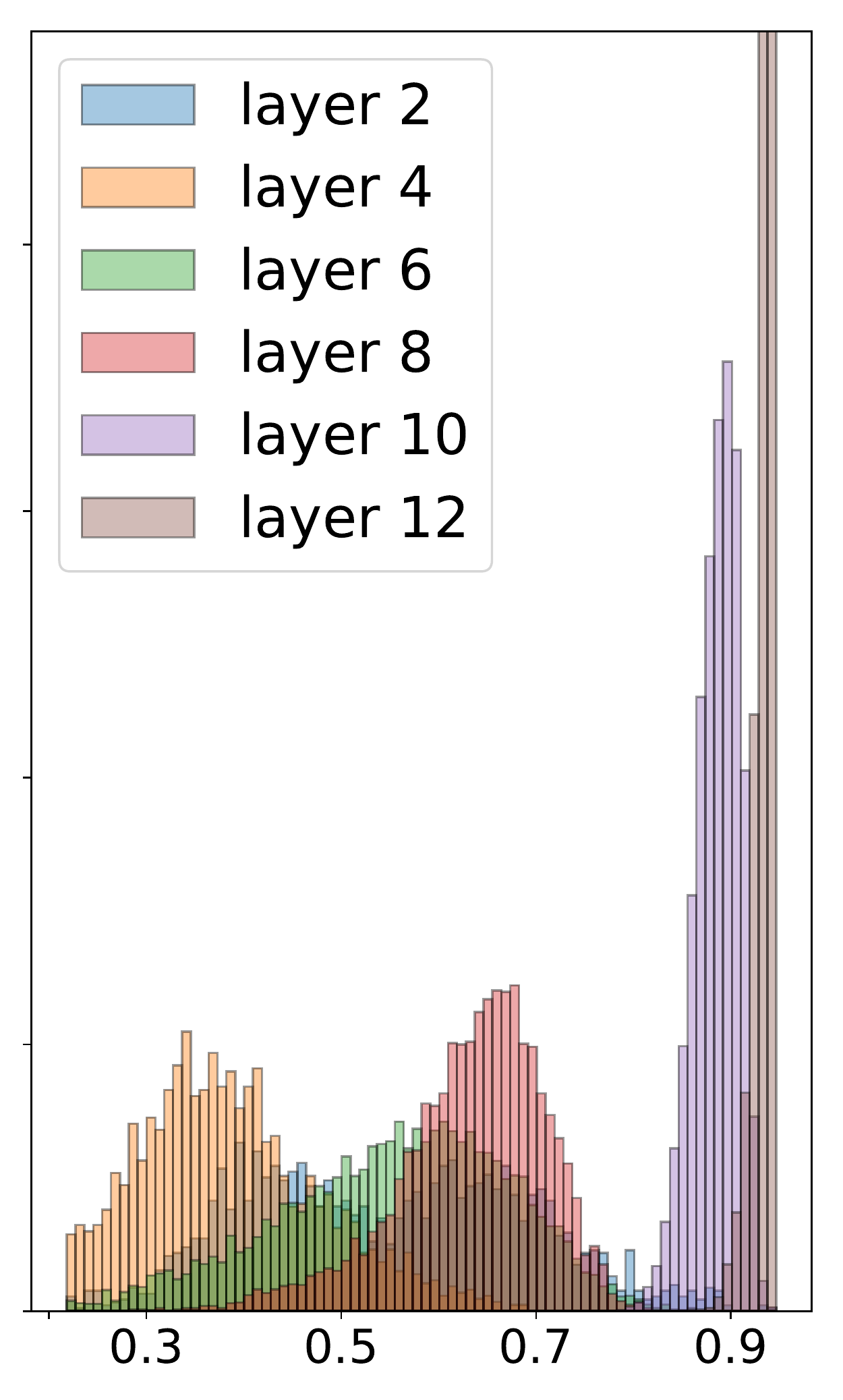}\vspace{-2pt}
  \caption{Histograms}
  \label{fig:sfig2}
\end{subfigure}\vspace{-2pt}
\caption{(a) \textbf{Correlation matrix of 64 randomly selected filters} from layer 2 to layer 12 of VGG16. At depth layer $d$ of VGG16 for ImageNet classification, we compute the similarity between two filters based on their guided back-propagation patterns \cite{springenberg2014striving} averaged on a set of images. As the layers get deeper, it becomes easier to find a set of filters that have a high similarity score to each other. (b) \textbf{Normalized histograms of pairwise filter similarities} of various VGG16 layers. As the number of channels increases with depth from 64 to 128 to 256, the curve shifts right and becomes far narrower, i.e., more filters become similar. Motivated by this, \textit{can we eliminate redundancy in the convolution layer by reusing similar filters?}}
\label{fig:histvgg}\vspace{-1pt}
\end{figure}
}

\fighistvgg{!t}

Convolution is the main building block of convolutional neural networks (CNN), which are widely successful on image classification \cite{krizhevsky2012imagenet,he2016deep,xie2017aggregated,simonyan2014very}, object detection \cite{girshick2015fast,ren2015faster,he2017mask}, image segmentation \cite{kirillov2019panoptic,long2015fully,chen2017deeplab,chen2018encoder} and action recognition \cite{ji20123d,wang2016temporal,carreira2017quo,wang2018non}.  However, standard convolution is still costly in terms of computation, storage, and memory access. More importantly, an optimized CNN often develops highly correlated filters.  

We can evaluate pairwise filter similarity  in standard convolution (SC), using the cosine similarity of guided back-propagation patterns \cite{springenberg2014striving} averaged over a set of ImageNet images.  \fig{histvgg} shows that, as the depth of layer increases, the filter correlation also increases. That is, filters become more similar from early to late layers, reducing the expressive power of feature representations. 

\figConvComp{!t}

How to optimize a CNN with less redundancy has been studied \cite{howard2017mobilenets,zhang2018shufflenet,ma2018shufflenet,xie2017aggregated}, often by exploring dependencies across space and channel dimensions.  In SC, while each filter has a limited size spatially, it extends to the full set of input features, whereas 
in group convolution (GC) \cite{krizhevsky2012imagenet}, a filter only convolves with a subset of input features.  Therefore, if there are $B$ groups of input features, each GC layer reduces the number of parameters $B$ times by reducing the size of each filter by $B$ times.  Depth-wise convolution (DW) is an extreme case of GC, where each group only contains one channel, maximally reducing the parameter count.

% For example, in order to get a network with better representation learning capability meanwhile reducing redundancy, ResNeXt \cite{xie2017aggregated} utilized group convolution \cite{krizhevsky2012imagenet} to reduce the size of each filter, and increased the total number of filters to enhance model capacity with comparable computing budget. 

While GC and DW are effective at reducing the model size, they do not look into the correlation between filters and their isolated representations cannot capture cross-channel relationships.  Instead of removing redundancy by reducing the size of each filter as in GC and DW, we explore another way of eliminating redundancy by exploring the potential of each filter. Directly reducing the number of filters is known to reduce the model capacity \cite{he2016deep}.  However, since SC filters become similar (\fig{histvgg}), we can reduce the {\it effective number} of filters by reusing them across different feature groups. We propose such a simple alternative called {\it tied block convolution} (TBC):
We split $C$ input feature channels into $B$ equal blocks, and use a single block filter defined only on $\frac{C}{B}$ channels to produce $B$ responses.  \fig{ConvComp} shows that an SC filter spans the entire $C$ channels, whereas at $B=2$, our TBC spans only $\frac{C}{2}$ channels and yet it also produces 2 filter responses.  TBC is simply GC shared across groups, and TBC is reduced to SC when $B\!=\!1$.  The tied block group convolution (TGC) and tied block fully connected layer (TFC) can be straightforwardly obtained by extending this concept to fully connected layer and group convolution layer.

Our TBC utilizes each filter, memory access, and samples more effectively.  {\bf 1)} At $B=2$, TBC obtains the same number of responses with one half-sized thin filter, approaching the same SC-size output with 4 times of model reduction.  {\bf 2)} As the same thin filter is applied to each of the $B$ blocks, TBC has more efficient memory access by utilizing GPU parallel processing.   {\bf 3)} Since each thin filter is trained on $B$ times more samples, learning also becomes more effective. {\bf 4)} Since each set of TBC filters are applied to all input channels, TBC could aggregate global information across channels and better model cross-channel dependencies.

% \figMaskrcnnComp{tp}

While TBC seems to be an appealing concept in theory, whether we can demonstrate its advantage in practice over SC or GC would be critically dependent upon neural network architectures.  We apply TBC/TGC/TFC to various backbone networks, including ResNet \cite{he2016deep}, ResNeXt \cite{xie2017aggregated}, SENet \cite{hu2018squeeze} and ResNeSt \cite{zhang2020resnest}, and propose their tied version: \textit{TiedResNet, TiedResNeXt, TiedSENet} and \textit{TiedResNeSt.} Extensive experimentation on classification, detection, segmentation, and attention are conducted, which demonstrate TBC/TGC/TFC's significant across-the-board performance improvement over \textit{standard convolution}, \textit{group convolution} and fully connected layer.  For example, Fig. \ref{fig:mscoco_params} shows TiedResNet consistently outperforms ResNet, ResNeXt and HRNetV2 \cite{wang2019deep} by a larger margin with a much leaner model.  Similar performance boost and model reduction are also obtained in various frameworks, tasks and datasets.

Lastly, learned filter redundancy not only reduces the model capacity at a bloated size, but also renders the CNN unable to capture diversity, resulting in inferior performance.  For object detection on MS-COCO, standard CNNs often fail to accurately locate the target object regions and aggregate useful information from the foreground.  Consequently, there are multiple overlapping partial object proposals, preventing a single full object proposal to emerge from the proposal pool.   TiedResNet can  handle high overlapping instances much better and increase the average precision (AP) by $6\%$ and AP at IoU = 0.75 by $8.3\%$, when the occlusion ratio is $0.8$.

%% file: 2related_work
\section{Related works}
\label{related_work}
% \textcolor{red}{TODO: Add detection related section.}

\noindent\textbf{Backbone Networks.}   AlexNet \cite{krizhevsky2012imagenet} is the first CNN success with significant accuracy gain on the ILSVRC competition. However, large kernels and fully connected layers greatly increase the model size.  With smaller kernels, GoogleNet \cite{szegedy2015going} and VGGNet \cite{simonyan2014very} only need 12 times fewer parameters to outperform \cite{krizhevsky2012imagenet,zeiler2014visualizing}.  However, the large network depth causes vanishing gradient problems, which is later solved by the residual connection design in ResNet \cite{he2016deep}.  Since the depth of model is no longer an issue, researchers have begun to explore how to use parameters more efficiently. With comparable model complexity, ResNeXt \cite{xie2017aggregated} outperformes ResNet in many major tasks, mainly due to the use of efficient group convolution.  Through careful design of the architecture, HRNetV2 \cite{wang2019deep} achieves the state-of-the-art performance on multiple major tasks. Compared with these works using either GC or SC, Our TBC further utilizes the full potential of each thinner filter.  We provide comparisons with these networks in remaining sections.

% In standard convolution, each kernel filter convolves on all the feature maps obtained on previous layer, resulting in lots of convolutions, some of which maybe redundant. 
\noindent\textbf{Group-wise Convolution.} Group convolution (GC) \cite{krizhevsky2012imagenet} is proposed to remove filter redundancy.  Since each GC filter only convolves with features in its group, with the same number of channels, this mechanism can reduce the number of parameters in each layer by a factor of $B$, where $B$ is the number of  groups.  
%Given the fixed model size, GC allows more filters and thus increases the network capacity.
%, leading to better accuracy.  
When the number of groups is the same as the number of input features, GC becomes identical to depth-wise convolution (DW) \cite{howard2017mobilenets}. Both GC and DW greatly reduce the model redundancy by reducing the size of each filter. \textit{However, they never look into the correlation between (learned) filters. }

As each filter in GC and DW only responds to partial input feature map, the ability to incorporate global information across channel dimensions is damaged in the GC and completely lost in the DW. In contrast, our TBC filter is shared across all input channels, and the long-range dependencies can be aggregated. This mechanism also introduces another benefit, our TBC only has one fragmentation. Therefore, TBC can take full advantage of the powerful parallel computing capabilities of the GPU.

\noindent\textbf{Attention Modules.} \cite{hu2018squeeze} introduces the squeeze-and-excitation (SE) module to adaptatively recalibrate channel-wise feature responses.  \cite{cao2019gcnet} unifies SE and a non-local \cite{wang2018non} module into a global context  block (GCB).  While SE and GCB are relatively light, SE (GCB) still counts for 10\% (25\%) of the  model size. Our tied block convolution and tied fully connected layers can be integrated into various attention modules and significantly reduce the number of parameters: 2.53M vs 0.04M for SE and 10M vs 2.5M for GCB.

%% file: 3methods
%****************** Methods ***********************
%**************************************************
% \def\figtiedBottleneck#1{
% \begin{figure}[#1]
%     \centering
%     \includegraphics[width=0.5\columnwidth]{figures/tiedbottleneck.pdf}
%     \caption{Diagram of (a) ResNet bottleneck module and (b) TiedResNet version with 4 splits. Each having its own TBC. Each tied block convolution has a specific block number.}
%     \label{fig:tiedBottleneck}
% \end{figure}
% }

\def\figtiedbottleneck#1{
\begin{figure*}[#1]
    \centering
    \begin{minipage}[t][][b]{.32\linewidth}
      \includegraphics[width=\linewidth, height=0.75\linewidth]{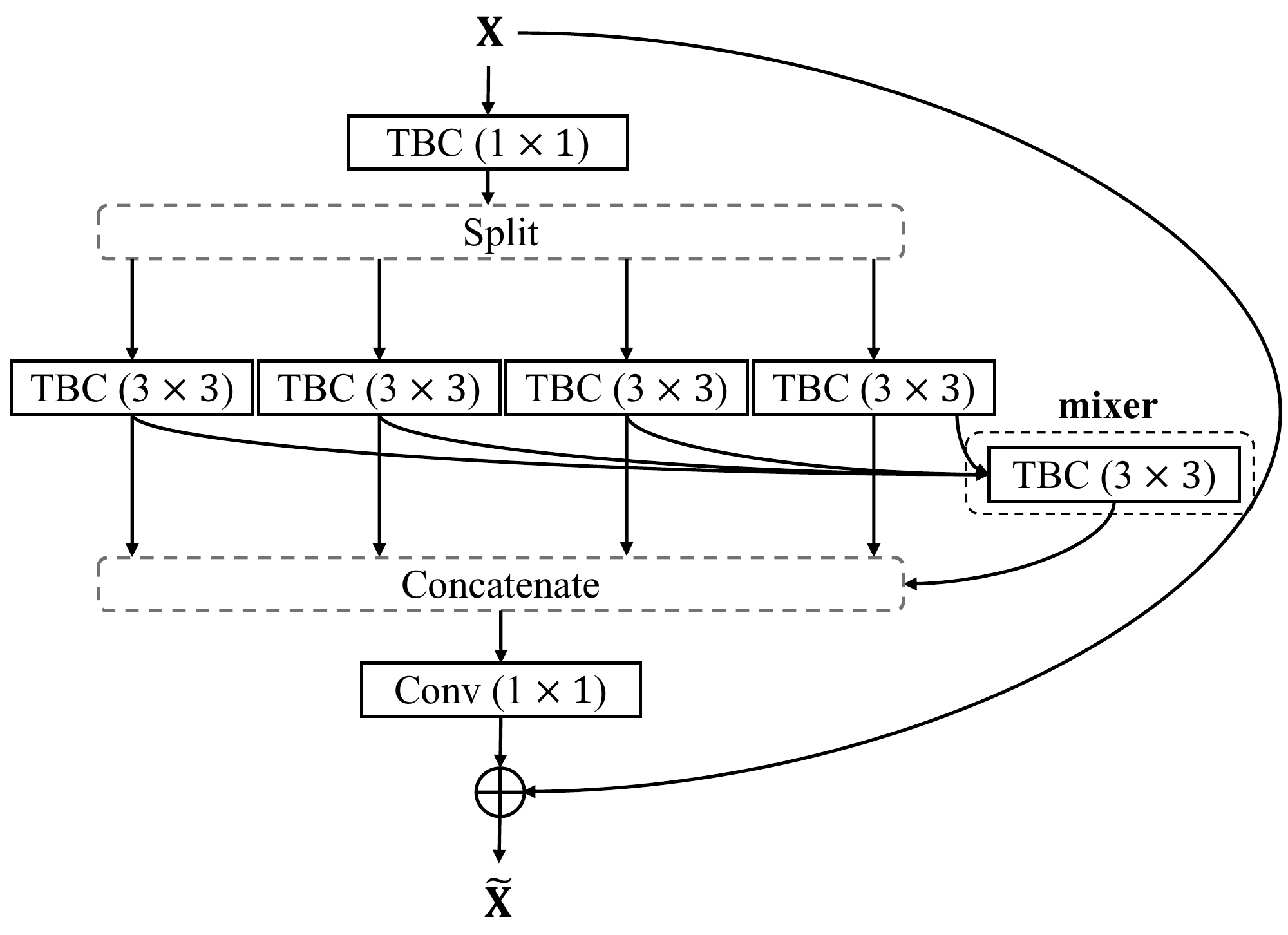}\vspace{-3pt}
      % \captionof{figure}{Second caption}
      \subcaption{TiedResNet}
      \label{TiedResNet}
    \end{minipage}
    % \hspace{.1\linewidth}
    \begin{minipage}[t][][b]{.32\linewidth}
      \includegraphics[width=\linewidth,height=0.75\linewidth]{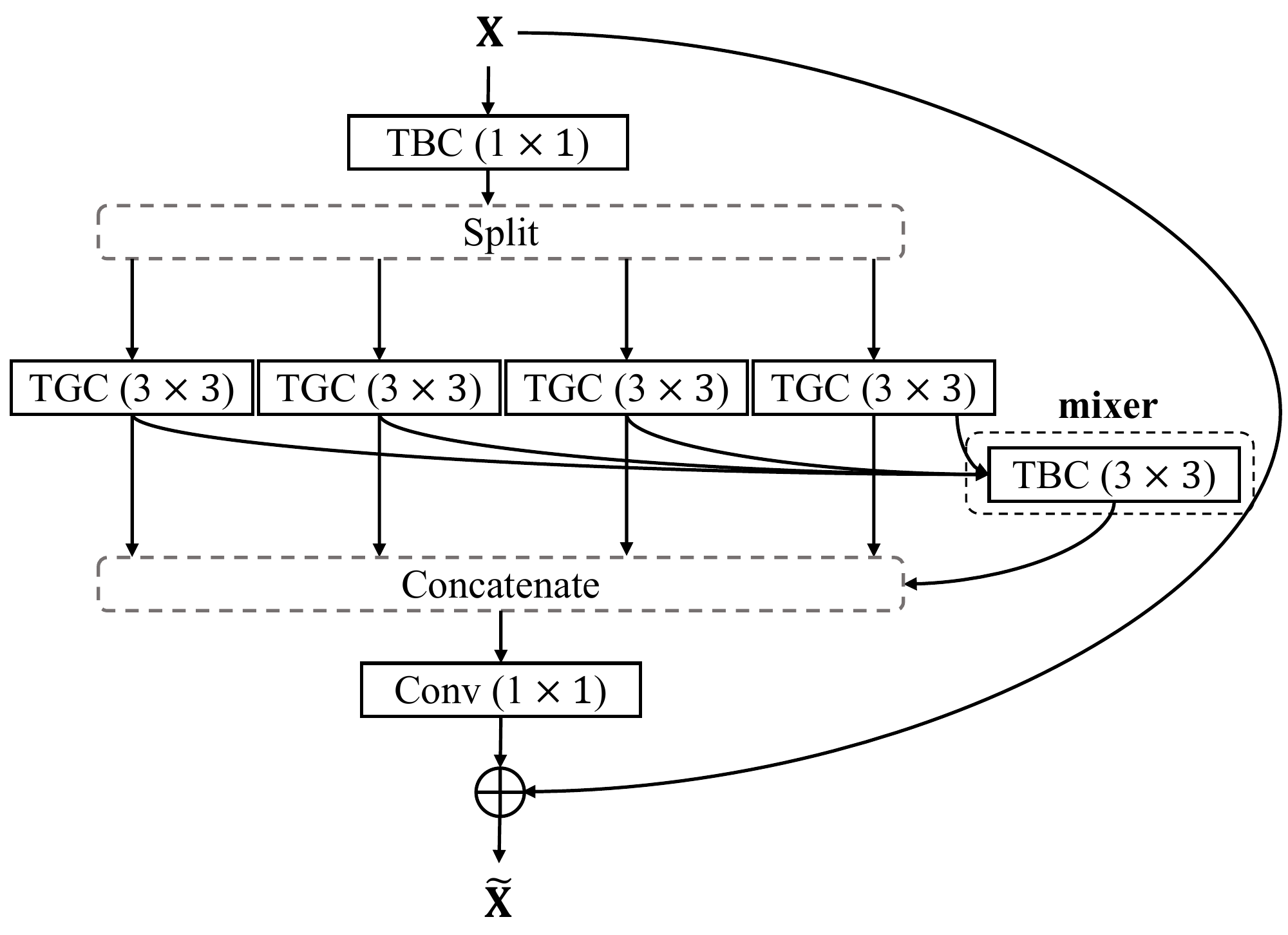}\vspace{-3pt}
      % \captionof{figure}{Second caption}
      \subcaption{TiedResNeXt}
      \label{TiedResNeXt}
    \end{minipage}
    \begin{minipage}[t][][b]{.32\linewidth}
      \includegraphics[width=\linewidth,height=0.75\linewidth]{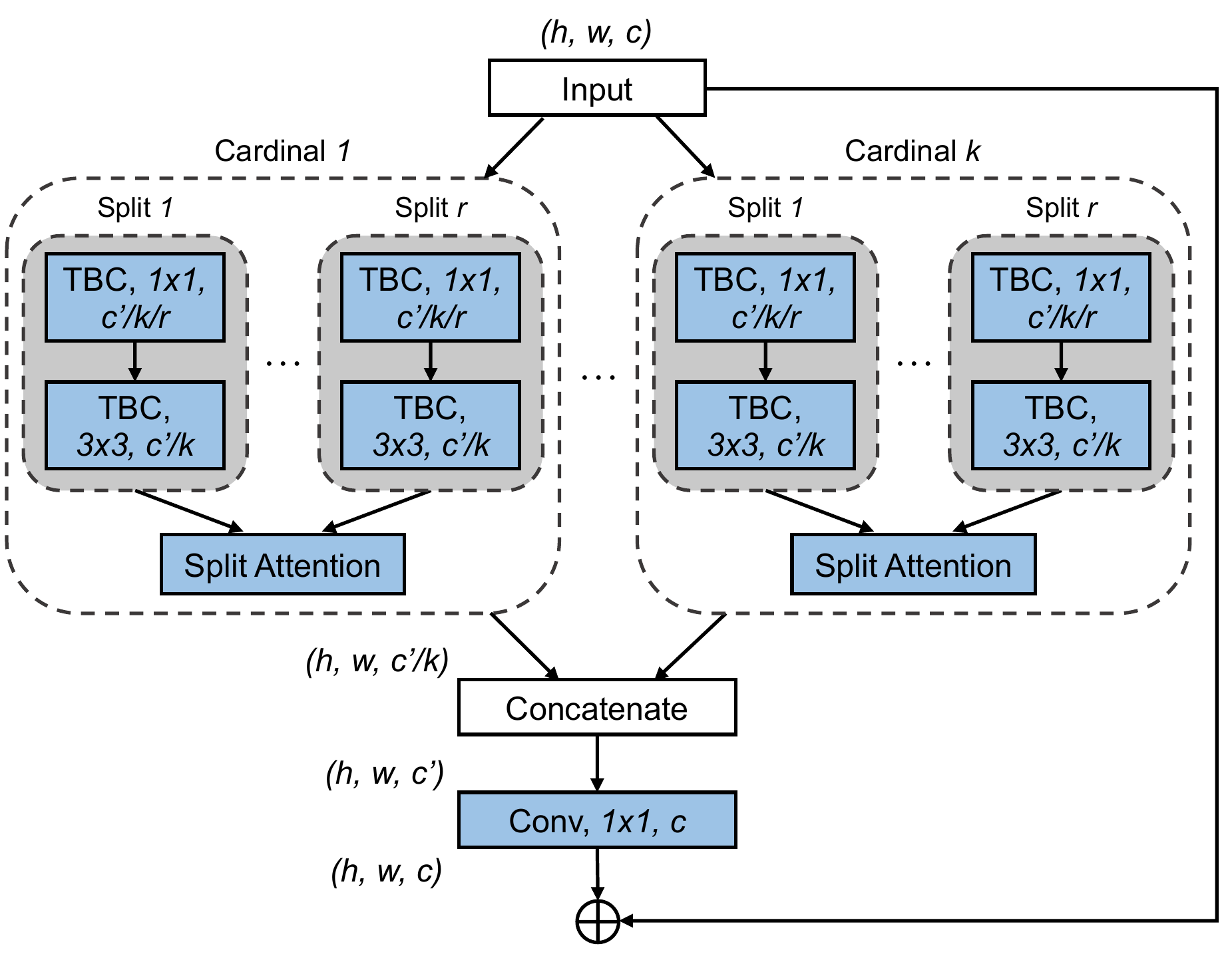}\vspace{-3pt}
      \subcaption{TiedResNeSt}
      \label{TiedResNeSt}
    \end{minipage}
    % Samples from Mask R-CNN
    \vspace{-3pt}
    \caption{\textbf{Diagram of bottleneck modules} for (a) TiedResNet with 4 splits (b) TiedResNeXt with 4 splits and (c) TiedResNeSt. Each tied block convolution (TBC) and tied block group convolution (TGC) has a specific block number.}
    \label{fig:tiedBottleneck}
\end{figure*}
}

\def\figtfc#1{
\begin{figure}[#1]
    \centering
    \begin{minipage}{.4\linewidth}
      \includegraphics[width=\linewidth]{figures/standard_fc.pdf}
      % \captionof{figure}{Second caption}
      \subcaption{Standard fc}
      \label{standard_conv}
    \end{minipage}
    \hspace{.1\linewidth}
    \begin{minipage}{.4\linewidth}
      \includegraphics[width=\linewidth]{figures/tied_fc.pdf}
      % \captionof{figure}{Second caption}
      \subcaption{Tied block fc(TFC)}
      \label{tied_conv}
    \end{minipage}
    % Samples from Mask R-CNN
    \caption{Diagram of (a) standard fully connected layer. (b) our proposed tied block fully connected layer.}
    \label{fig:tfc}
\end{figure}
}

\def\figatten#1{
\begin{figure}[#1]
% \begin{subfigure}{.24\linewidth}
%   \centering
%   \includegraphics[width=.95\linewidth]{figures/SE.pdf}
%   \caption{SE}
%   \label{fig:sfig2}
% \end{subfigure}
\begin{subfigure}{.48\linewidth}
  \centering
  \includegraphics[width=\linewidth]{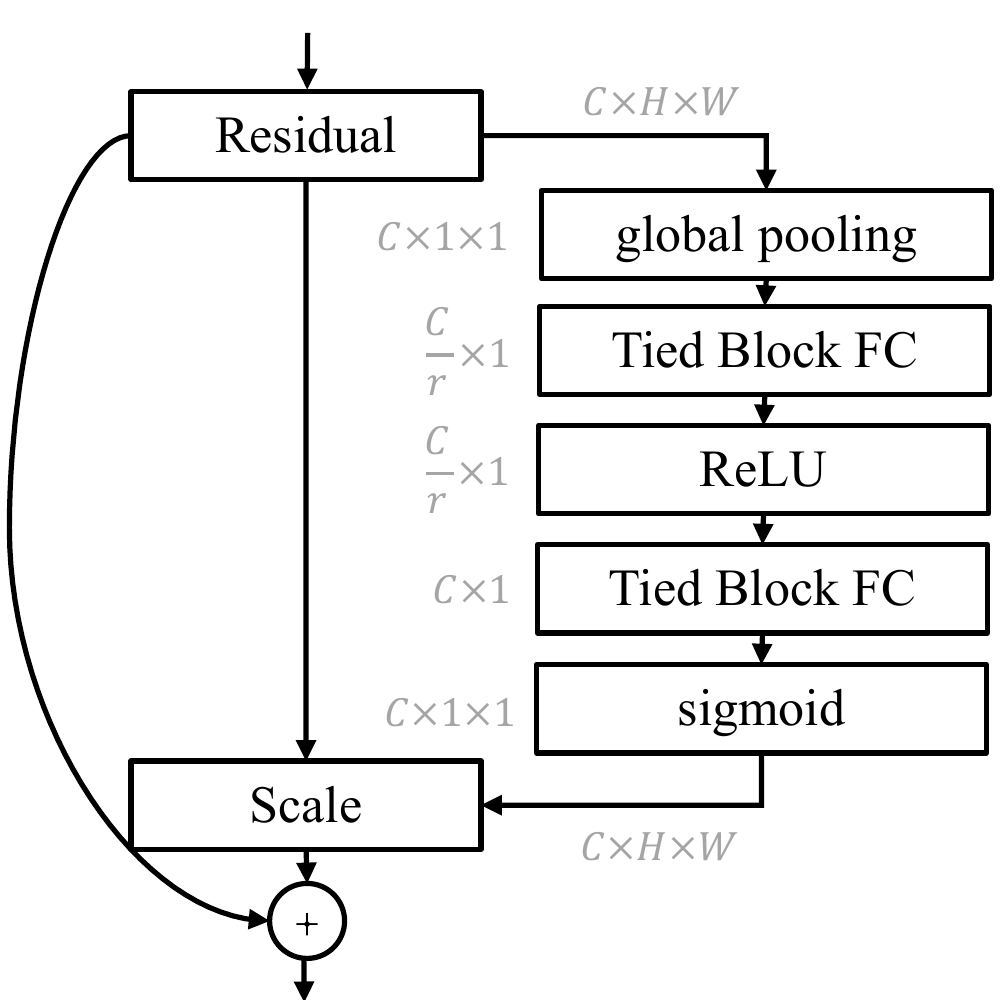}\vspace{-2pt}
  \caption{TiedSE}
  \label{fig:sfig2}
\end{subfigure}
% \begin{subfigure}{.24\linewidth}
%   \centering
%   \includegraphics[width=.95\linewidth]{figures/GC.pdf}
%   \caption{GCB}
%   \label{fig:sfig2}
% \end{subfigure}
\begin{subfigure}{.48\linewidth}
  \centering
  \includegraphics[width=\linewidth]{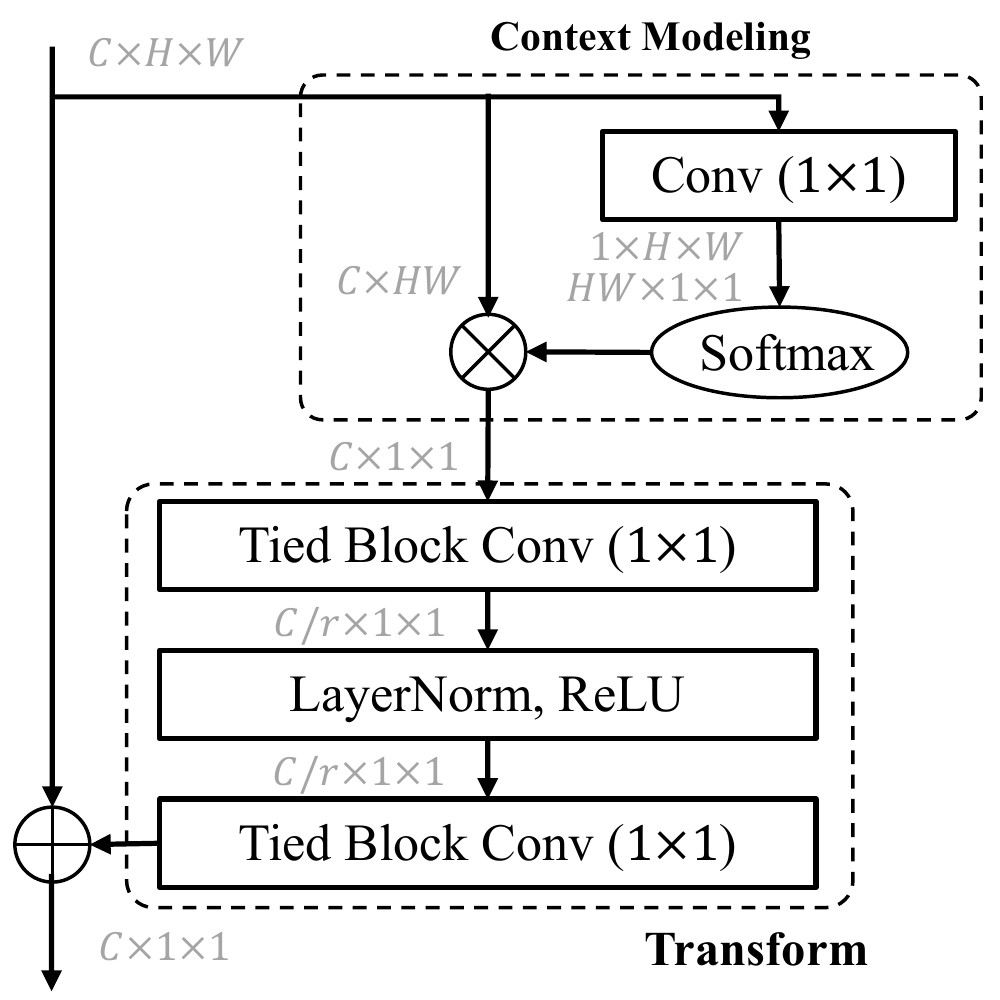}\vspace{-2pt}
  \caption{TiedGCB}
  \label{fig:sfig2}
\end{subfigure}\vspace{-3pt}
\caption{\textbf{Diagram of Tied attention modules.} (a) TiedSE module replaces FC in the original squeeze-and-excitation (SE) module \cite{hu2018squeeze} to be TFC. (b) TiedGCB module replaces standard convolution in global context block (GCB) \cite{cao2019gcnet} with TBC.}
\label{fig:atten}
\end{figure}
}

\def\figtimeCost#1{
\begin{figure}[#1]
\centering
\includegraphics[width=0.95\linewidth]{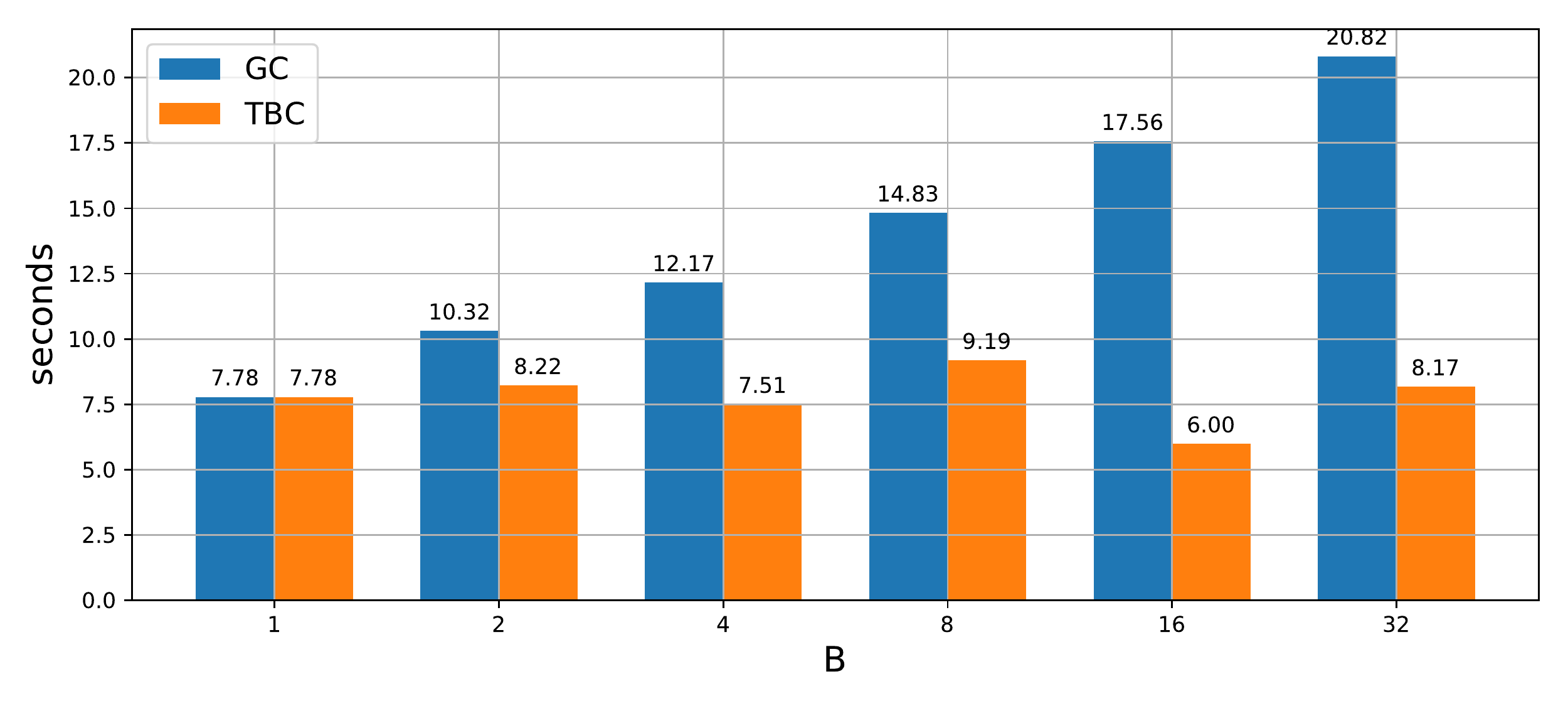}
\vspace{-9pt}
\caption{\textbf{The time cost of processing 1k iterations} of each feature map using the RTX 2080Ti GPU. When group number increases, GC increases the time cost almost linearly. In contrast, when using a larger $B$, TBC keeps a similar time cost. Different block numbers $B$ were tested for GC and TBC, the total FLOPs at these values were fixed by changing the total filter number. When $B=1$, GC and TBC are equal to SC. Input feature map size is 56$\times$56$\times$2048.}
\label{fig:timeCost}
\end{figure}
}

\section{Tied Block Convolution Network Design}
\label{batch_convolution}

We first analyze TBC and TGC to guide us in network design.  
% Among many options for replacing SC with TBC, we focus on ResNet and introduce our experimentally validated choices for its key bottleneck modules. 
We also develop TFC and apply to attention modules.
%We use the following acronyms throughout:
%standard convolution (SC),
%tied block convolution (TBC), 
%group convolution (GC), 
%depthwise separable convolution (DWS),
%fully connected (FC),
%tied block fully connected (TFC).

\subsection{TBC Formulation}

Let the input feature be denoted by $X \in \mathbb{R}^{c_i\times h_i\times w_i}$ and the output feature  $\tilde{X} \in \mathbb{R}^{{c_o\times h_o\times w_o}}$, where $c, h, w$ are the number of channels, the  height and width of feature maps respectively. The kernel size is $k \times k$ and the bias term is ignored for clarity.

\noindent{\bf Standard Convolution}, denoted by $*$, can be formulated as:
\begin{equation}
    \tilde{X} = X * W
\end{equation}
\noindent{}where $W \in \mathbb{R}^{c_o \times c_i \times k \times k}$ is the SC kernel.  The parameters for SC is thus: $c_o \times c_i \times k \times k.$
% \begin{equation}
%     c_o \times c_i \times k \times k.
% \end{equation}

% Group Convolution
\noindent{\bf Group Convolution} first divides input feature $X$ into $G$ equal-sized groups $X_1,...,X_G$ with size ${c_i/G \times h_i \times w_i}$ per group.  Each group  shares the same convolutional filters $W_g$. The output of GC is computed as:
\begin{equation}
    \tilde{X} = X_1 * W_1 \oplus X_2 * W_2 \oplus \cdots \oplus X_G * W_G
\end{equation}
\noindent{}where $\oplus$ is the concatenation operation along the channel dimension, $W_g$ is the convolution filters for group $g$, where $g \in \{1,\ldots, G\}$, $W_g\in \mathbb{R}^{\frac{c_o}{G} \times \frac{c_i}{G} \times k \times k}$. The number of parameters for GC is: $G \times \frac{c_o}{G} \times \frac{c_i}{G} \times k \times k.$
% \begin{equation}
%     G \times \frac{c_o}{G} \times \frac{c_i}{G} \times k \times k.
% \end{equation}

% Tied Block Convolution
\noindent{\bf Tied Block Convolution} reduces the \textit{effective number} of filters by reusing filters across different feature groups with the following formula:
% is simply GC shared across groups, i.e. $G_1=\cdots=G_B$:
\begin{equation}
    \tilde{X} = X_1 * W' \oplus X_2 * W' \oplus \cdots \oplus X_B * W'
\end{equation}
\noindent{}where $W' \in \mathbb{R}^{\frac{c_o}{B} \times \frac{c_i}{B} \times k \times k}$ is the TBC filters shared among all the groups. The parameter number is: $\frac{c_o}{B} \times \frac{c_i}{B} \times k \times k.$
% \begin{equation}
%     \frac{c_o}{B} \times \frac{c_i}{B} \times k \times k.
% \end{equation}

% Not only does TBC reduce the size of each filter by a factor of $B$, but it also reduces the number of filters by a factor of $B$, which results in a $B^2 \times$ parameter reduction. In addition, TBC's single fragmentation property also enables it to take full advantage of GPU parallel computing capabilities as in Fig. \ref{fig:timeCost}. As the number of groups in the GC increases, the processing time increases linearly. On the contrary, our TBC maintains almost the same processing time. This shows that TBC can make full use of the GPU parallel computing capabilities that GC does not do.

\figtimeCost{}

% \subsection{Relation to Prior Work}

\noindent{\bf TBC vs. GC}. While TBC is GC with filters shared across groups, it has several major distinctions from GC in practical consequences (assume $B = G$).
\begin{enumerate}
\setlength{\itemsep}{0pt}
%{\bf 1)} 
\item TBC has $B\times$ fewer parameters than GC. 
% where $B$ is the number of blocks (or groups) for GC / TBC.  
% GC reduces \#parameters by  $B\times$ only through reducing the \#parameters per filter, whereas TBC further reduces the \#parameters by \textit{another} $B\times$ through sharing the weights among the filters.

%{\bf 2)} 
\item TBC only has one fragmentation on GPU utilization, whereas GC has G fragmentations, greatly reducing the degree of parallelism.  Fig.\ref{fig:timeCost} shows that the processing time increases linearly with the number of groups in GC, whereas our TBC keeps almost the same processing time.%Multiple fragmentation also causes overhead on kernel launching and synchronization.
% TBC can thus fully utilize GPU parallel computing power, which GC fail to do. 

%{\bf 3)} 
\item TBC can better model cross-channel dependencies.  Since each set of GC filters are only convolved on subsets of channels, GC has trouble aggregating global information across channels.  However, each set of TBC filters are applied to all input channels and can better model cross-channel dependencies.

%{\bf 4)} 
\item
TBC-based TiedResNet greatly surpasses GC-integrated ResNeXt in object detection and instance segmentation tasks. TiedResNet-S can even outperform ResNeXt with 2$\times$ model size reduction, demonstrating that TiedResNet make more effective use of model parameters.
\end{enumerate}

% \noindent{\bf TBC \textit{v.s.} DWS} Depthwise convolution is GC at the extreme, where \#groups is the same as \#channels.  TBC differs from DWS on three accounts.

% {\bf 1)} The kernel size cannot be too small for DWS, whereas TBC has no such restrictions and can easily be $1\times 1$ kernels and straightforwardly extended to FC layer.

% {\bf 2)} DWS has the maximal number of fragmentations, hindering its speed on a GPU.  By comparison, TBC doesn't have such problems since it only has one fragmentation.

% {\bf 3)} It is much easier to adopt TBC for practical performance gain, as it only needs minor changes on popular architectures.  In  stark contrast,
% DWS would require a brand new architecture, many tricks and a much longer training schedule, in order to significantly reduce \#parameters with a relatively small (around 5\%) performance drop.
% As in [24], directly applying DWS to existing architectures leads to significant performance drop.

\noindent{\bf Tied Block Group Convolution (TGC)}
The idea of tied block filtering can also be directly applied to group convolution, formulated as: 
\begin{equation}
    \begin{split}
    \tilde{X} = &(X_{11} * W'_1 \oplus \cdots \oplus X_{1B} * W'_1) \oplus \cdots \oplus \\
    & (X_{G1} * W'_G \oplus \cdots \oplus X_{GB} * W'_G)
    \end{split}
\end{equation}
\noindent{}where $W'_g \in \mathbb{R}^{\frac{c_o}{BG} \times \frac{c_i}{BG} \times k \times k}$, $X_{gb} \in \mathbb{R}^{\frac{c_i}{BG} \times h_i \times w_i}$ is the divided feature map, $g \in [1, G]$ and $b \in [1, B]$.

\figtiedbottleneck{!t}
\figatten{!t}

\noindent{\bf Tied Block Fully Connected Layer (TFC)}
% \figtfc{h}
Convolution is a special case of fully connected (FC) layer, just as FC is a special case of convolution.  We apply the same tied block filtering idea to FC.  Tied block fully connected layer (TFC) shares the FC connections between equal blocks of input channels.  Like TBC, TFC could reduce \(B^2\) times parameters and $B$ times computational cost.

\subsection{TBC/TGC in Bottleneck Modules}

The ResNet/ResNeXt/ResNeSt bottleneck modules have $1\times1$ and $3\times3$ convolutional filters.  We apply TBC/TGC differently as in Fig.\ref{fig:tiedBottleneck}. For \(3\times3\) in ResNet and ResNeXt, we split all the filters into groups; each group has its own TBC/TGC setting. This choice allows different levels of sharing and is motivated by network visualization works \cite{zeiler2014visualizing,bau2017network}:  Filters assume different roles at different layers and some are unique concept detectors \cite{agrawal2015learning,bau2017network}.  For the \(1\times1\) convolutions at the entry and the exit of bottlenecks, we replace the entry one by TBC with $B\!=\!2$ to allow filter sharing, while maintaining the exit convolution to aggregate information across channels. Since ResNeSt replaces \(3\times3\) convolutions to be multi-path and split attention modules with $k$ cardinals, \(3\times3\) convolutions occupy less proportion of the overall model complexity. Therefore, we only replace all \(3\times3\) convolution to be TBC with $B=2$ as in \(1\times1\) convolution. Further increase of $B$ will only marginally reduce the model parameters, but will greatly reduce performance.

The default setting for TiedResNet-50 (TiedResNeXt-50) is 4 splits with base width of 32 (64), i.e. 4s$\times$32w (4s$\times$64w), and the default setting for TiedResNet-S (TiedResNeXt-50-S) is 4s$\times$18w (4s$\times$36w). Our TiedBottleNeck reaches more than 1\% performance improvement in term of top-1 accuracy on ImageNet-1K. However, losing cross-channel integration could weaken the model.  To add it back, we introduce a mixer that fuses outputs of multiple splits.  Introducing the mixer increases performance by another 0.5\%.  The input to the mixer can be either concatenation or element-wise sum of split outputs. Table \ref{table:ablation_bconv_mixer} shows that element-wise sum has a better trade-off.

\subsection{TBC and TFC in Attention Modules}

We apply TBC and TFC to attention modules such as SE \cite{hu2018squeeze} and  GCB \cite{cao2019gcnet}, by simply replacing SC and FC with their tied block counterparts (\fig{atten}).  Both designs significantly reduce the number of parameters without dropping performance.  
% Tables \ref{table:imagenet_se_settings} and \ref{table:gcnet} show that, despite model reduction by 16$\times$ and 4$\times$, our TiedSE and TiedGCB maintain on-par performance with the benchmark.

%% file: 4experiments
%%%%%%%%% Splts Number and Fusion Methods
\def\tabSplitsFusion#1{
\begin{table*}[#1]
\tablestyle{2.5pt}{1.05}
\begin{minipage}[t][][b]{.56\textwidth}
    % \tablestyle{2.0pt}{1.05}
    \centering
    \setlength{\tabcolsep}{5pt}
    \begin{tabular}{l||c|c|c|c|c}
    \shline
    model & setting & params & GFlops & top-1 & top-5 \\ [.1em]
    \hline\hline
    % ResNet-50\cite{he2016deep} & - & 76.15 & 92.87 & 25.6 & 4.2 \\
    % TiedResNet-50 & 1s\(\times\)80w & 76.75 & 93.21 & 23.8 & \bf{3.9} \\
    TiedResNet-50 & 2s\(\times\)48w & 23.8 & 4.4 & 77.27 & 93.53 \\
    TiedResNet-50 & 4s\(\times\)32w & \bf{22.0} & 4.4 & \bf{77.61} & \bf{93.62} \\
    TiedResNet-50 & 6s\(\times\)24w & 23.0 & 4.6 & 77.37 & 93.66 \\
    TiedResNet-50 & 8s\(\times\)18w & 23.8 & 4.4 & 77.21 & 93.54 \\
    \shline
    \end{tabular}\vspace{-8pt}
    \caption{Ablation study on \textbf{splits number and base width of each split}. Accuracies (\%) on ImageNet-1k are listed.}
    \label{table:ablation_bconv_settings}
\end{minipage}
\begin{minipage}[t][][b]{.36\textwidth}
    % \tablestyle{2.0pt}{1.05}
    \centering
    \setlength{\tabcolsep}{5pt}
    \begin{tabular}{l||c|c|c}
    \shline
    mixer & top-1 & top-5 & params \\ [.1em]
    \hline\hline
    baseline & 76.15\% & 92.87\% & 25.6 \\
    \hline
    none & 77.17\% & 93.27\% & 20.8 \\
    \hline
    element-sum & 77.61\% & 93.62\% & 22.0 \\
    concatenate & \bf{77.65\%} & \bf{93.64\%} & 26.7 \\
    \shline
    \end{tabular}\vspace{-5pt}
    \caption{Ablation study on \textbf{fusion methods} of mixer module.}
    \label{table:ablation_bconv_mixer}
\end{minipage}
\end{table*}
}

\def\tabSplits#1{
\begin{table}[#1]
\tablestyle{2.5pt}{1.05}
    % \tablestyle{2.0pt}{1.05}
    \centering
    \setlength{\tabcolsep}{5pt}
    \begin{tabular}{l||c|c|c|c|c}
    \shline
    model & setting & params & GFlops & top-1 & top-5 \\ [.1em]
    \hline\hline
    % ResNet-50\cite{he2016deep} & - & 76.15 & 92.87 & 25.6 & 4.2 \\
    % TiedResNet-50 & 1s\(\times\)80w & 76.75 & 93.21 & 23.8 & \bf{3.9} \\
    TiedResNet-50 & 2s\(\times\)48w & 23.8 & 4.4 & 77.27 & 93.53 \\
    TiedResNet-50 & 4s\(\times\)32w & \bf{22.0} & 4.4 & \bf{77.61} & \bf{93.62} \\
    TiedResNet-50 & 6s\(\times\)24w & 23.0 & 4.6 & 77.37 & 93.66 \\
    TiedResNet-50 & 8s\(\times\)18w & 23.8 & 4.4 & 77.21 & 93.54 \\
    \shline
    \end{tabular}\vspace{-5pt}
    \caption{Ablation study on \textbf{splits number} and \textbf{base width of each split}. Accuracies (\%) on ImageNet-1k are listed.}
    \label{table:ablation_bconv_settings}
\end{table}
}

\def\tabFusion#1{
\begin{table}[#1]
\tablestyle{2.5pt}{1.05}
    % \tablestyle{2.0pt}{1.05}
    \centering
    \setlength{\tabcolsep}{7pt}
    \begin{tabular}{l||c|c|c}
    \shline
    mixer & top-1 acc. & top-5 acc. & \#params (M) \\ [.1em]
    \hline\hline
    % baseline & 76.15 & 92.87 & 25.6 \\
    % \hline
    % none & 77.17 & 93.27 & 20.8 \\
    % \hline
    element-sum & 77.61\% & 93.62\% & \textbf{22.0} \\
    concatenate & \bf{77.65\%} & \bf{93.64\%} & 26.7 \\
    \shline
    \end{tabular}\vspace{-5pt}
    \caption{Ablation study on \textbf{fusion method} of mixer module.}
    \label{table:ablation_bconv_mixer}
\end{table}
}

\def\tabImageNet#1{
\begin{table}[#1]
\tablestyle{2.5pt}{1.02}
% \scriptsize
\centering
\begin{threeparttable}
\begin{minipage}{1\linewidth}
\begin{tabular}{l||l|l|c|c}
\shline
model & params(M) & GFlops & top-1(\%) & top-5(\%) \\ [.1em]
% \hline\hline
% % DLA60 \cite{yu2018deep} & 22.8 & 4.3 & 76.68 & 93.40\\
% ResNeXt50 & 25.5 & 4.1 & 77.4 & 93.5 \\
% SEResNet50 & 28.1 & 4.2 & 76.7 & 93.4\\
% InceptionV3 & 23.8 & 5.0 & 77.4 & 93.6 \\
% % Densenet201 & 20.2 & 4.3 & 77.2 & 93.6 \\
% ResNeXt101-32$\times4$d & 44.2 & 8.1 & 78.5 & 94.1 \\
% Densenet161 & 28.7 & 7.8 & 77.7 & 93.8 \\
\hline\hline
\multicolumn{5}{l}{\textbf{\textit{ResNet50}} \cite{he2016deep}} \\
\hline
baseline & 25.6 & 4.2 & 76.2 & 92.9 \\
TiedResNet50-S & \bf{13.9} {\scriptsize(54\%)} & \bf{2.5} {\scriptsize(60\%)} & 76.3 & 92.9\\
TiedResNet50 & 22.0 {\scriptsize(86\%)} & 4.4 {\scriptsize(105\%)} & \bf{77.6} & \bf{93.6}\\
\hline\hline
\multicolumn{5}{l}{\textbf{\textit{ResNet101}} \cite{he2016deep}} \\
\hline
baseline & 44.6 & 7.9 & 77.4 & 93.6 \\
TiedResNet101-S & \bf{24.0} {\scriptsize(54\%)} & \bf{4.8} {\scriptsize(61\%)} & 77.7 & 93.8 \\
TiedResNet101 & 39.4 {\scriptsize(88\%)} & 8.6 {\scriptsize(109\%)} & \bf{78.8} & \bf{94.2} \\
\hline\hline
\multicolumn{5}{l}{\textbf{\textit{ResNeXt101-32$\times$8d}} \cite{xie2017aggregated}} \\
\hline
baseline & 88.8 & 16.5 & 79.3 & 94.5 \\
TiedResNeXt101-S & \bf{64.0} {\scriptsize(65\%)} & \bf{14.6} {\scriptsize(78\%)} & 79.3 & 94.5 \\
\hline\hline
\multicolumn{5}{l}{\textbf{\textit{SENet101}} \cite{hu2018squeeze}} \\
\hline
baseline & 49.1 & 7.9 & 77.6 & 93.9 \\
baseline $\ddagger$ & 49.1 & 7.9 & 78.3 & 94.2 \\
TiedSENet101-S & \bf{26.4} {\scriptsize(54\%)} & \bf{5.2} {\scriptsize(66\%)} & 79.0 & 94.5 \\
TiedSENet101-S $\dagger$ & \bf{26.4} {\scriptsize(54\%)} & \bf{5.2} {\scriptsize(66\%)} & \bf{80.9} & \bf{95.3} \\
TiedSENet101 & 41.8 {\scriptsize(85\%)} & 9.1 {\scriptsize(115\%)} & 79.8 & 94.8 \\
\hline\hline
\multicolumn{5}{l}{\textbf{\textit{ResNeSt-50-fast}} \cite{zhang2020resnest}} \\
\hline
baseline $\ddagger$ & 27.5 & 4.4 & 78.6 & 93.9 \\
TiedResNeSt50-S & \bf{16.5} {\scriptsize(60\%)} & \bf{3.6} {\scriptsize(82\%)} & \bf{78.8} & \bf{94.6} \\
\hline\hline
\multicolumn{5}{l}{\textbf{\textit{VS. pruning methods and Mobile nets (large model version)}}} \\
\hline
Taylor-FO-BN & 14.2 & 2.3 & 74.5 & - \\
ShuffleNet-50 $^\dagger$ & - & 2.3 & 74.8 & - \\
GhostNet-50 ($s$=2) & \bf{13.0} & \bf{2.2} & 75.0 & 92.3 \\
\hline
TiedResNet50-S & 13.9 & 2.5 & \bf{76.3} & \bf{92.9}\\
\shline
\end{tabular}
% \begin{tablenotes}
%     \small
%     \item[$\dagger$] trained with larger epochs label smoothing, cosine learning scheduler and heavier data augmentation 
%     \item[$\ddagger$] re-implemented results with released codes, standard data augmentations and 100 training epochs.
% \end{tablenotes}
\end{minipage}
\end{threeparttable}
\vspace{-5pt}
\caption{\textbf{Recognition accuracy and model size comparison on ImageNet-1k}. The integration of TBC/TFC/TGC can obtain consistent performance improvements to various backbone networks. TiedResNet-S even greatly surpasses current SOTA pruning methods Taylor-FO-BN-ResNet50 \cite{molchanov2019importance} and Mobile architecture GhostNet (large model version) \cite{han2020ghostnet}. These results prove that TBC makes more efficient use of parameters. Baselines are copied from Pytorch model zoo, their TBC versions are trained for 100 epochs on 8 2080Ti GPUs to make fair comparisons, unless otherwise noticed. $\dagger$ denotes: trained with larger epochs, label smoothing, cosine learning scheduler and heavier data augmentation. $\ddagger$ denotes: re-implemented results with released codes, standard data augmentations and 100 training epochs.}
\label{table:imagenet_models}
\end{table}
}

\def\tabDetectionTest#1{
\begin{table*}[#1]
\tablestyle{2.5pt}{1.05}
% \scriptsize
\centering
\begin{tabular}{l|x{36}|x{13}x{13}x{13}x{13}x{16}|x{13}x{13}x{13}x{13}x{16}}
\shline
\multirow{2}{*}{Backbone} & \multirow{2}{*}{Params} & \multicolumn{5}{c|}{Object Detection} &\multicolumn{5}{c}{Instance Segmentation} \\\cline{3-12}
&& AP & AP$_{S}$ & AP$_{M}$ & AP$_{L}$ & AP$_{50}$ & AP & AP$_{S}$ & AP$_{M}$ & AP$_{L}$ & AP$_{50}$\\ [0.1em]
\hline\hline
\multicolumn{12}{c}{Mask R-CNN} \\
\hline
ResNeXt101-32$\times$4d & 44.2M &41.9&24.2&45.1&53.0&63.3  & 37.7&17.9&40.1&53.5&60.1\\
ResNet101 &  44.6M &40.8&22.9&43.9&52.0&62.3  &37.0&16.9&39.4&53.1&59.1 \\
% \textcolor{gray}{ResNeXt101-64$\times$4d} & \textcolor{gray}{83.5M} & \textcolor{gray}{0}&\textcolor{gray}{0}&\textcolor{gray}{0}&\textcolor{gray}{0}&\textcolor{gray}{0} & \textcolor{gray}{0}&\textcolor{gray}{0}&\textcolor{gray}{0}&\textcolor{gray}{0}&\textcolor{gray}{0} \\
\hline
TiedResNet101-S &  \bf{24.0M} &\bf{42.2}&\bf{24.3}&\bf{45.4}&\bf{53.4}&\bf{63.6}  &\bf{38.0}&\bf{17.9}&\bf{40.3}&\bf{54.2}&\bf{60.5} \\
TiedResNet101 &  39.4M & \bf{43.0}&\bf{25.3}&\bf{46.4}&\bf{54.3}&\bf{64.4} & \bf{38.6}&\bf{18.7}&\bf{41.2}&\bf{54.8}&\bf{61.4} \\
\hline\hline
\multicolumn{12}{c}{Cascade Mask R-CNN} \\
\hline
ResNeXt101-32$\times$4d & 44.2M & 45.1&25.5&48.2&57.4&63.6 & 39.0&18.4&41.4&55.1&61.0 \\
ResNet101 & 44.6M & 44.0&24.3&46.9&56.7&62.3 & 38.0&17.5&40.3&54.4&59.6 \\
% \textcolor{gray}{ResNeXt101-64$\times$4d} & \textcolor{gray}{83.5M} & \textcolor{gray}{0}&\textcolor{gray}{0}&\textcolor{gray}{0}&\textcolor{gray}{0}&\textcolor{gray}{0} & \textcolor{gray}{0}&\textcolor{gray}{0}&\textcolor{gray}{0}&\textcolor{gray}{0}&\textcolor{gray}{0} \\
\hline
TiedResNet101-S & \bf{24.0M} & \bf{45.3}&\bf{25.7}&\bf{48.4}&\bf{57.9}&\bf{64.0} & \bf{39.3}&\bf{18.5}&\bf{41.8}&\bf{55.9}&\bf{61.3} \\
TiedResNet101 & 39.4M & \bf{46.1}&\bf{27.0}&\bf{49.2}&\bf{58.7}&\bf{64.7} & \bf{39.8}&\bf{19.4}&\bf{42.1}&\bf{56.2}&\bf{62.0} \\
\hline\hline
\multicolumn{12}{c}{Hybrid Task Cascade} \\
\hline
ResNeXt101-32$\times$4d & 44.2M & 46.4&26.8&49.4&59.6&65.8 & 40.7&19.3&43.1&58.2&63.2 \\
ResNet101 & 44.6M & 45.1&25.2&48.0&58.2&64.3 & 39.7&18.4&42.2&57.0&61.8 \\
% \textcolor{gray}{ResNeXt101-64$\times$4d} & \textcolor{gray}{83.5M} & \textcolor{gray}{47.2}&\textcolor{gray}{27.7}&\textcolor{gray}{50.1}&\textcolor{gray}{60.3}&\textcolor{gray}{66.5} & \textcolor{gray}{41.3}&\textcolor{gray}{20.0}&\textcolor{gray}{43.7}&\textcolor{gray}{58.9}&\textcolor{gray}{63.9} \\
\hline
TiedResNet101-S & \bf{24.0M} & \bf{46.6}&\bf{27.0}&\bf{49.5}&\bf{59.9}&\bf{66.0} & \bf{41.0}&\bf{19.5}&\bf{43.5}&58.1&\bf{63.4} \\
TiedResNet101 & 39.4M & \bf{47.1}&\bf{27.6}&\bf{50.3}&\bf{60.0}&\bf{66.4} & \bf{41.3}&\bf{20.3}&\bf{43.8}&\bf{58.7}&\bf{63.9} \\
\shline
\end{tabular}\vspace{-5pt}
\caption{Comparison on MS-COCO \textit{test-dev}. Our TiedResNet101-S not only reduces the parameters and computation costs by \textit{half} but also obtains \textit{1.3$\sim$1.4}\% better performance than \textit{ResNet101} and outperforms \textit{ResNeXt101} in \textit{all} experimented frameworks. \textit{Without} the training and inference time multi-scale augmentation, TiedResNet-101 with Hybrid Task Cascade reaches \textit{state-of-the-art} performance in MS-COCO. Baseline models are obtained from \cite{mmdetection}.}
\label{table:coco_detection_instance_seg_test}
\end{table*}
}

\def\tabDetectionVal#1{
\begin{table}[#1]
\tablestyle{2.5pt}{1.05}
% \scriptsize
\centering
\begin{tabular}{l|r|c|c}
\shline
\multirow{2}{*}{Backbone} & \multirow{2}{*}{$\#$param} & \multicolumn{1}{c|}{\footnotesize{Object Detection}} &\multicolumn{1}{c}{Segmentation} \\\cline{3-4}
&& AP/AP$_{S}$/AP$_{M}$/AP$_{L}$ & AP/AP$_{S}$/AP$_{M}$/AP$_{L}$\\ [0.1em]
\shline
% \multicolumn{4}{c}{Mask R-CNN \cite{he2017mask}} \\
% \hline
% R-50 & 25.6M & 38.5/22.6/42.0/50.5 & 35.1/16.7/37.7/52.0 \\
% TR-50-S & \bf{13.9M} & 39.6/23.0/43.3/51.3 & 36.2/17.0/38.8/52.8 \\
% \rowcolor{Gray}
% \textit{v baseline} & {$\downarrow$11.7M} & {{+1.1}/{+0.4}/{+1.3}/{+0.8}} & {{+1.1}/{+0.3}/{+1.1}/{+0.8}} \\
% TR-50 & 22.0M & \bf{40.9}/\bf{24.0}/\bf{44.7}/\bf{53.6} & \bf{37.0}/\bf{17.4}/\bf{39.9}/\bf{54.6} \\
% \rowcolor{Gray}
% \textit{v baseline}  & {$\downarrow$3.6M} & {{+2.4}/{+1.4}/{+2.7}/{+3.1}} & {{+1.9}/{+0.7}/{+2.2}/{+2.6}} \\
% \hline
% R-101 & 44.6M & 40.3/22.2/44.8/52.9 & 36.5/16.3/39.7/54.6 \\
% TR-101-S & \bf{24.0M} & \bf{41.7}/\bf{24.1}/\bf{45.8}/\bf{54.3} & \bf{37.5}/\bf{17.9}/\bf{40.5}/\bf{55.0} \\
% \rowcolor{Gray}
% \textit{v baseline} & {$\downarrow$20.6M} & {{+1.4}/{+1.9}/{+1.0}/{+1.4}} & {{+1.0}/{+1.6}/{+0.8}/{+0.4}}\\
% TR-101 & 39.4M & \bf{42.8}/\bf{24.2}/\bf{46.8}/\bf{57.2} & \bf{38.4}/\bf{18.2}/\bf{41.5}/\bf{57.0} \\
% \rowcolor{Gray}
% \textit{v baseline} & {$\downarrow$5.2M} & {{+2.5}/{+2.0}/{+2.0}/{+4.3}} & {{+1.9}/{+1.9}/{+0.8}/{+3.4}}\\
% \hline\hline
\multicolumn{4}{c}{Cascade Mask R-CNN \cite{cai2018cascade}} \\
\hline
R-50 & 25.6M & 42.3/23.7/45.7/56.4 & 36.6/17.3/39.0/53.9 \\
TR-50 & 22.0M & \bf{44.7}/\bf{25.8}/\bf{47.9}/\bf{59.3} & \bf{38.4}/\bf{18.7}/\bf{40.9}/\bf{56.5} \\
\rowcolor{Gray}
\textit{v baseline} & {$\downarrow$3.6M} & {{+2.4}/{+2.1}/{+2.2}/{+2.9}} & {{+1.8}/{+1.6}/{+1.9}/{+2.6}} \\
\hline
R-101 & 44.6M & 43.3/24.4/46.9/58.0 & 37.6/17.3/40.4/56.2 \\
TR-101-S & 24.0M & \bf{44.5}/\bf{25.2}/\bf{48.4}/\bf{59.0} & \bf{38.6}/\bf{18.4}/\bf{41.6}/\bf{56.8} \\
\rowcolor{Gray}
\textit{v baseline} & {$\downarrow$22.6M} & {{+1.2}/+0.8/+1.6/+1.0} & {{+1.0}/+1.1/+1.2/+0.6} \\
TR-101 & 39.4M & \bf{45.6}/\bf{26.4}/\bf{49.5}/\bf{60.7} & \bf{39.3}/\bf{19.1}/\bf{42.4}/\bf{58.0} \\
\rowcolor{Gray}
\textit{v baseline} & {$\downarrow$5.2M} & {{+2.3}/+2.0/+2.6/+2.7} & {{+1.7}/+1.8/+2.0/+2.2} \\
\shline
\end{tabular}\vspace{-2pt}
\caption{MS-COCO \textit{val-2017} instance segmentation and object detection comparison. 2$\times$ learning rate schedule is applied for both baselines and TiedResNet (TR) \textit{without} multi-scale augmentations. The parameter number for backbone networks are listed. Comparison with other detectors is carried out in the appendix materials.}
\label{table:coco_detection_instance_seg_val}
\end{table}
}

%%%%%%%%%%%%%%%% Filter similarity
\def\figFilterSim#1{
\begin{figure}[#1]
    \centering
    \begin{minipage}{.32\linewidth}
      \includegraphics[width=\linewidth]{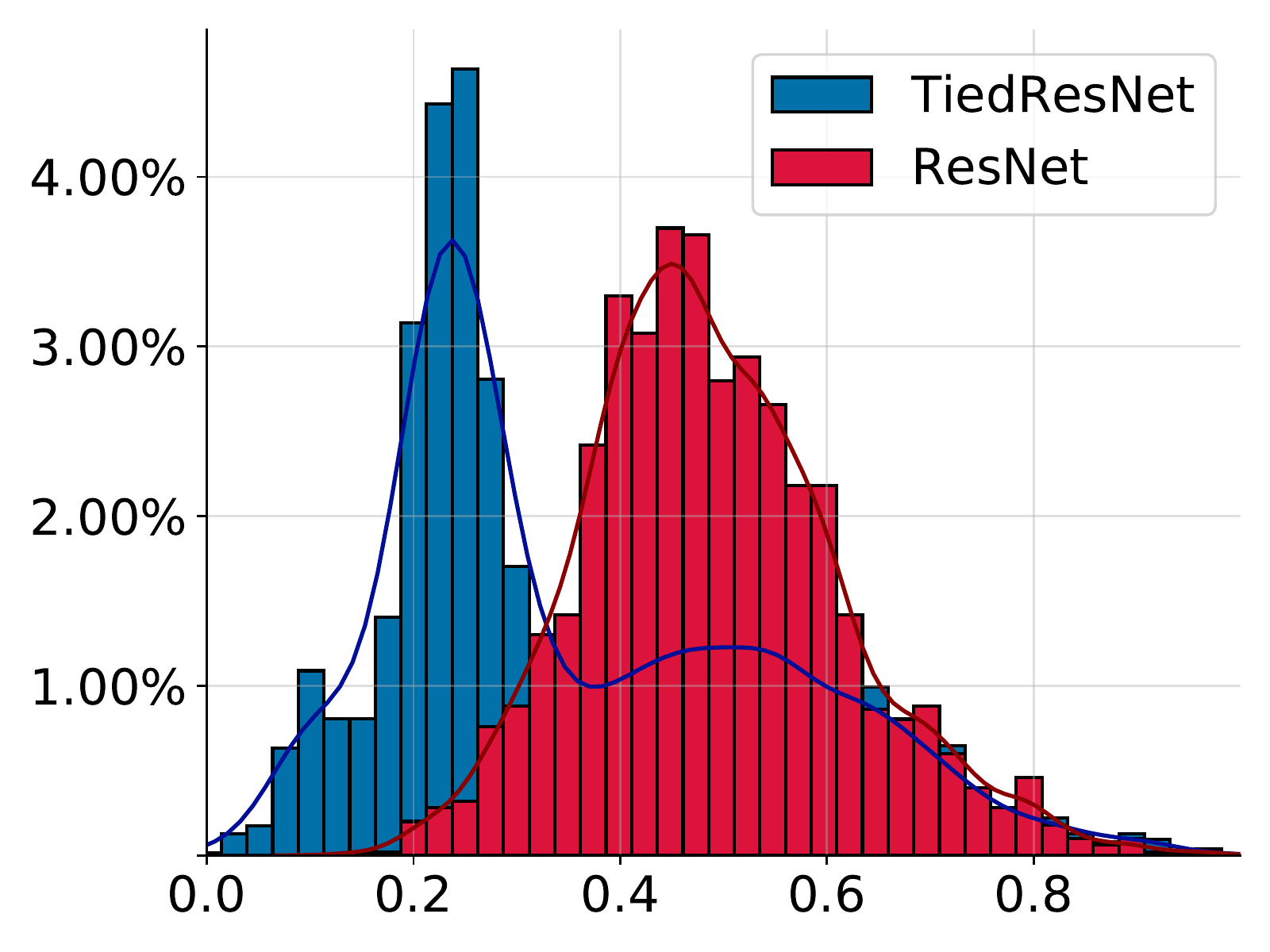}\vspace{-3pt}
      % \captionof{figure}{Second caption}
      \subcaption{Layer 9}
      \label{standard_conv}
    \end{minipage}
    % \hspace{.02\linewidth}
    \begin{minipage}{.32\linewidth}
      \includegraphics[width=\linewidth]{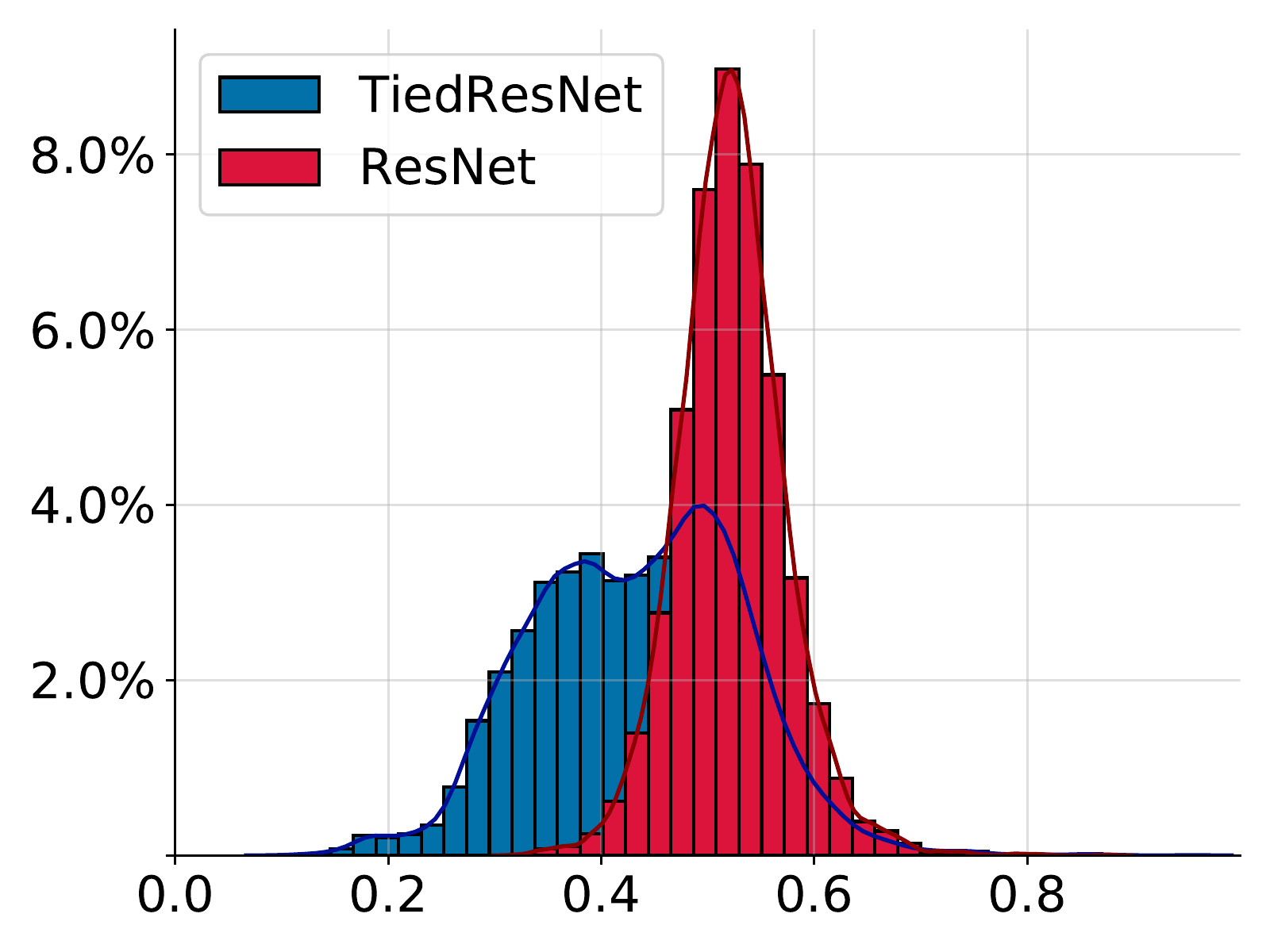}\vspace{-3pt}
      % \captionof{figure}{Second caption}
      \subcaption{Layer 18}
      \label{tied_conv}
    \end{minipage}
    \begin{minipage}{.32\linewidth}
      \includegraphics[width=\linewidth]{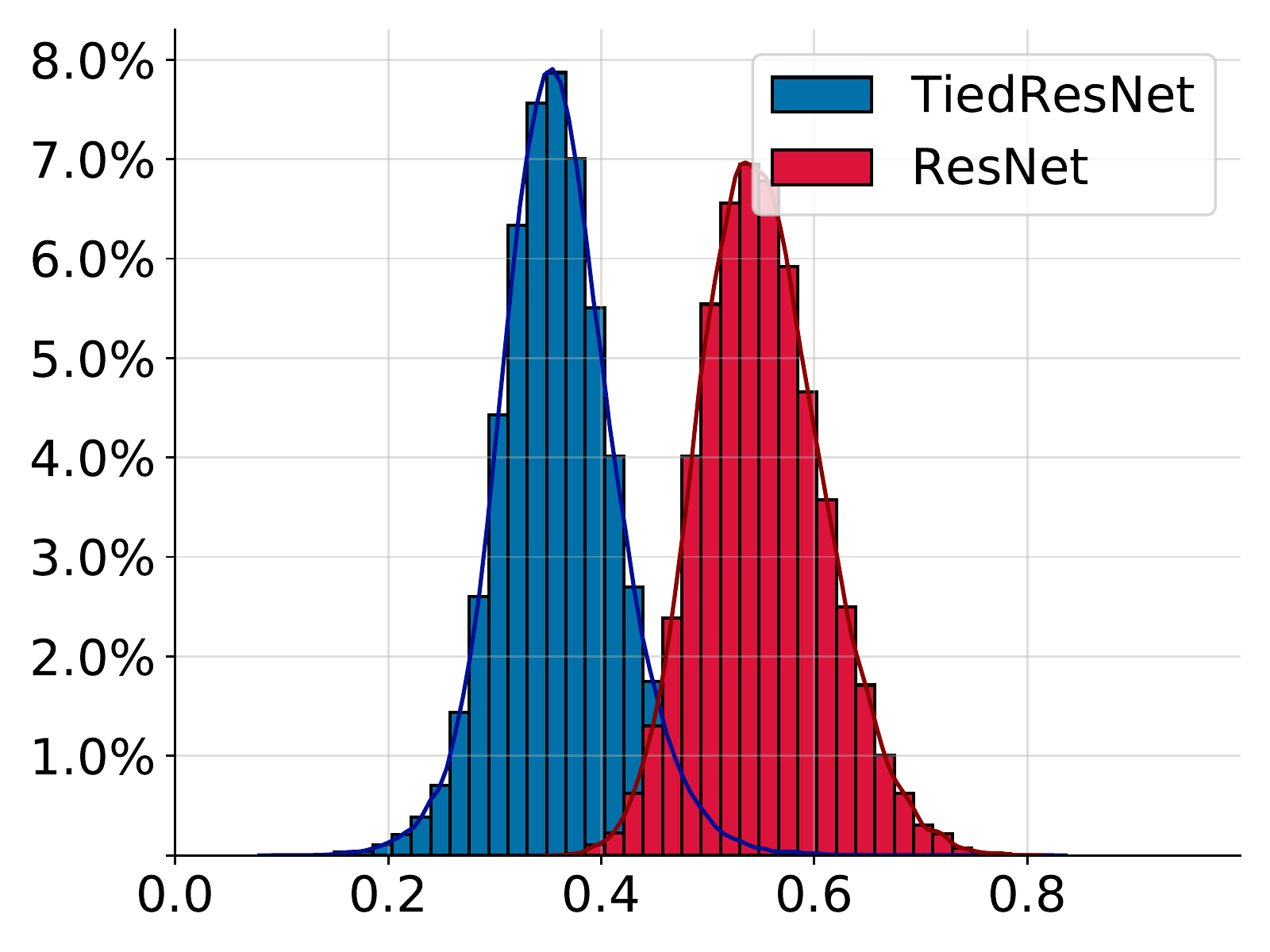}\vspace{-3pt}
      % \captionof{figure}{Second caption}
      \subcaption{Layer 36}
      \label{tied_conv}
    \end{minipage}
    % Samples from Mask R-CNN
    \vspace{-2pt}
    \caption{Histograms of pairwise filter similarity.}
    \label{fig:hist_tiedres}
\end{figure}
}

%%%%%%%%%%%%%%%% Grad-CAM Visualization 
\def\gradcam#1#2#3#4#5#6#7#8#9{
    \small
    \rotatebox[origin=c]{90}{#1}
    \begin{subfigure}{0.085\textwidth}
        \includegraphics[width=1\linewidth,height=\linewidth]{figures/gradcam/#1/#2.png}
    \end{subfigure}\hspace{1.5pt}%
    \begin{subfigure}{0.085\textwidth}
        \includegraphics[width=1\linewidth,height=\linewidth]{figures/gradcam/#1/#3.png}
    \end{subfigure}\hspace{1.5pt}%
    \begin{subfigure}{0.085\textwidth}
        \includegraphics[width=1\linewidth,height=\linewidth]{figures/gradcam/#1/#4.png}
    \end{subfigure}\hspace{1.5pt}%
   \begin{subfigure}{0.085\textwidth}
        \includegraphics[width=1\linewidth,height=\linewidth]{figures/gradcam/#1/#5.png}
    \end{subfigure}\hspace{1.5pt}%
    \begin{subfigure}{0.085\textwidth}
        \includegraphics[width=1\linewidth,height=\linewidth]{figures/gradcam/#1/#6.png}
    \end{subfigure}\hspace{1.5pt}%
    \begin{subfigure}{0.085\textwidth}
        \includegraphics[width=1\linewidth,height=\linewidth]{figures/gradcam/#1/#7.png}
    \end{subfigure}\hspace{1.5pt}%
    \begin{subfigure}{0.085\textwidth}
        \includegraphics[width=1\linewidth,height=\linewidth]{figures/gradcam/#1/#8.png}
    \end{subfigure}\hspace{1.5pt}%
    \begin{subfigure}{0.085\textwidth}
        \includegraphics[width=1\linewidth,height=\linewidth]{figures/gradcam/#1/#9.png}
    \end{subfigure}\hspace{1.5pt}%
    \begin{subfigure}{0.085\textwidth}
        \includegraphics[width=1\linewidth,height=\linewidth]{figures/gradcam/#1/tree_frog.png}
    \end{subfigure}\hspace{1.5pt}%
    \begin{subfigure}{0.085\textwidth}
        \includegraphics[width=1\linewidth,height=\linewidth]{figures/gradcam/#1/banana.png}
    \end{subfigure}\hspace{1.5pt}%
    \begin{subfigure}{0.085\textwidth}
        \includegraphics[width=1\linewidth,height=\linewidth]{figures/gradcam/#1/zebra.png}
    \end{subfigure}\hspace{1.5pt}%
}

%%%%%%%%%%%%%%%% Grad-CAM Visualization 
\def\gradcamSingle#1#2#3#4#5#6#7{
    \rotatebox[origin=c]{90}{\scriptsize{#1}}
    \begin{subfigure}{0.07\textwidth}
        \includegraphics[width=1\linewidth,height=\linewidth]{figures/gradcam/#1/#2.png}
    \end{subfigure}\hspace{1.5pt}%
    \begin{subfigure}{0.07\textwidth}
        \includegraphics[width=1\linewidth,height=\linewidth]{figures/gradcam/#1/#3.png}
    \end{subfigure}\hspace{1.5pt}%
    \begin{subfigure}{0.07\textwidth}
        \includegraphics[width=1\linewidth,height=\linewidth]{figures/gradcam/#1/#4.png}
    \end{subfigure}\hspace{1.5pt}%
   \begin{subfigure}{0.07\textwidth}
        \includegraphics[width=1\linewidth,height=\linewidth]{figures/gradcam/#1/#5.png}
    \end{subfigure}\hspace{1.5pt}%
    \begin{subfigure}{0.07\textwidth}
        \includegraphics[width=1\linewidth,height=\linewidth]{figures/gradcam/#1/#6.png}
    \end{subfigure}\hspace{1.5pt}%
    \begin{subfigure}{0.07\textwidth}
        \includegraphics[width=1\linewidth,height=\linewidth]{figures/gradcam/#1/#7.png}
    \end{subfigure}\hspace{1.5pt}%
}

\def\gradcamA#1#2#3#4#5#6#7#8#9{
\imwh{figures/gradcam/#1/#2.png}{0.082}{0.07}&
\imwh{figures/gradcam/#1/#3.png}{0.082}{0.07}&
\imwh{figures/gradcam/#1/#4.png}{0.082}{0.07}&
\imwh{figures/gradcam/#1/#5.png}{0.082}{0.07}&
\imwh{figures/gradcam/#1/#6.png}{0.082}{0.07}&
\imwh{figures/gradcam/#1/#7.png}{0.082}{0.07}&
\imwh{figures/gradcam/#1/#8.png}{0.082}{0.07}&
\imwh{figures/gradcam/#1/#9.png}{0.082}{0.07}&
\imwh{figures/gradcam/#1/tree_frog.png}{0.082}{0.07}&
\imwh{figures/gradcam/#1/banana.png}{0.082}{0.07}&
\imwh{figures/gradcam/#1/keeshond.png}{0.082}{0.07}&
\imwh{figures/gradcam/#1/zebra.png}{0.082}{0.07}\\
}

\def\figGradCam#1{
\begin{figure*}[#1]
    \centering
\tb{@{}cccccccccccc@{}}{0.15}{
\gradcam{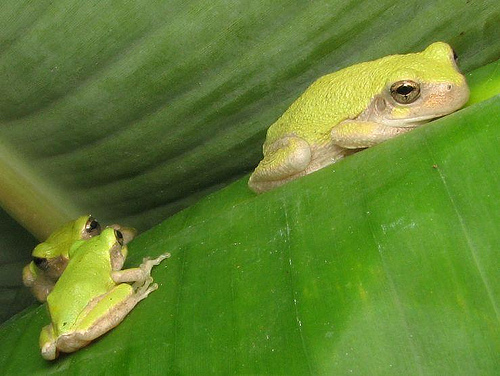}{tiger_cat}{kite}{racer}{koala}{studio_couch}{table_lamp}{television}{umbrella}\vspace{2pt}\\
\gradcam{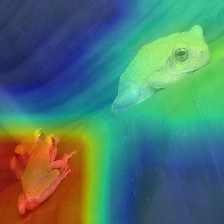}{tiger_cat}{kite}{racer}{koala}{studio_couch}{table_lamp}{television}{umbrella}\vspace{2pt}\\
\gradcam{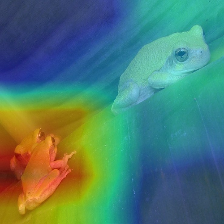}{tiger_cat}{kite}{racer}{koala}{studio_couch}{table_lamp}{television}{umbrella}\vspace{2pt}\\
\gradcam{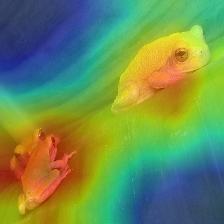}{tiger_cat}{kite}{racer}{koala}{studio_couch}{table_lamp}{television}{umbrella}
}\vspace{-4pt}
\caption{Grad-CAM \cite{selvaraju2017grad} visualization comparison among ResNet50, ResNeXt50 and TiedResNet50. The grad-CAM is calculated for the last convolutional output.}
\label{fig:gradcam}
\end{figure*}
}

\def\figGradCamSingle#1{
\begin{figure}[#1]
    \centering
\tb{@{}ccccccc@{}}{0.15}{
\gradcamSingle{Origin}{tiger_cat}{racer}{koala}{studio_couch}{tree_frog}{zebra}\vspace{2pt}\\
\gradcamSingle{ResNet}{tiger_cat}{racer}{koala}{studio_couch}{tree_frog}{zebra}\vspace{2pt}\\
\gradcamSingle{ResNeXt}{tiger_cat}{racer}{koala}{studio_couch}{tree_frog}{zebra}\vspace{2pt}\\
\gradcamSingle{TiedResNet}{tiger_cat}{racer}{koala}{studio_couch}{tree_frog}{zebra}
}\vspace{-6pt}
\caption{\textbf{Grad-CAM visualization} comparison among ResNet50, ResNeXt50 and TiedResNet50 for images in Row 1. The grad-CAM \cite{selvaraju2017grad} is calculated for the last convolutional output.}
\label{fig:gradcam}
\end{figure}
}

%%%%%%%%%%%%%%%% Detection and Instance Segmentation Comparison
\def\figMsCoCoDetector#1{
\begin{figure*}[#1]
    \centering
    \includegraphics[width=0.95\linewidth]{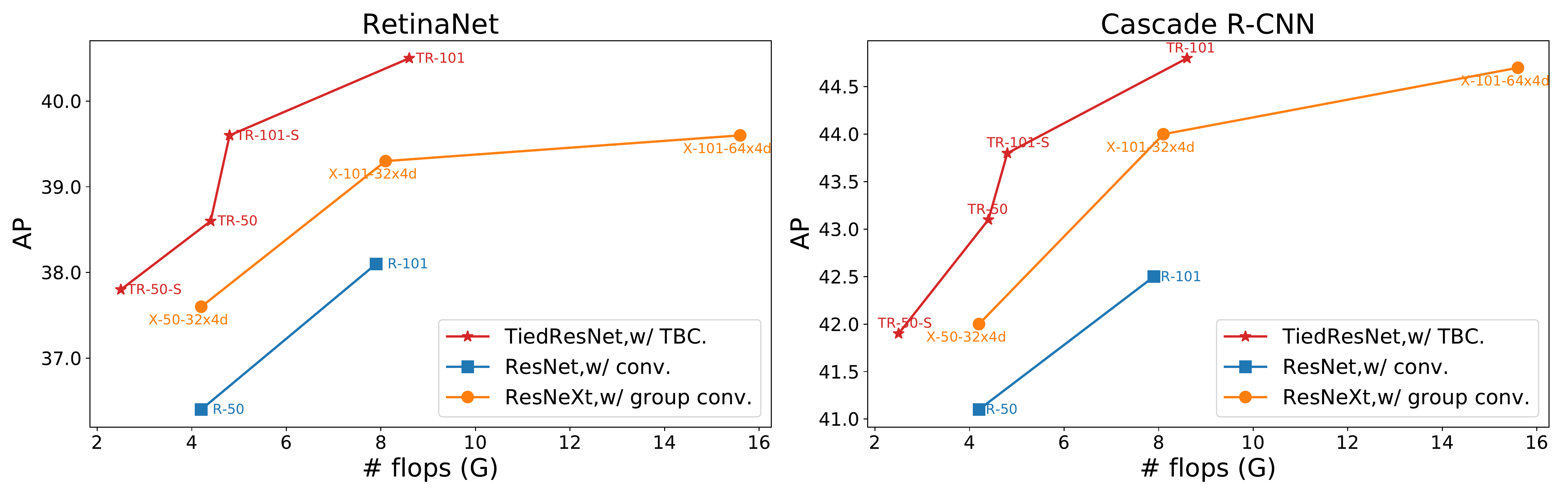}\vspace{-6pt}
    \caption{Average Precision vs. FLOPs of backbones on MS-COCO \textit{val-2017}. For state-of-the-art \textit{single-stage} detector RetinaNet and \textit{two-stage} detector Cascade R-CNN, TiedResNet consistently outperforms ResNet and ResNeXt.}
    \label{fig:mscoco_flops}
\end{figure*}
}

\def\figMaskrcnnComp#1{
\begin{figure*}[#1]
    \centering
    \includegraphics[width=0.95\linewidth]{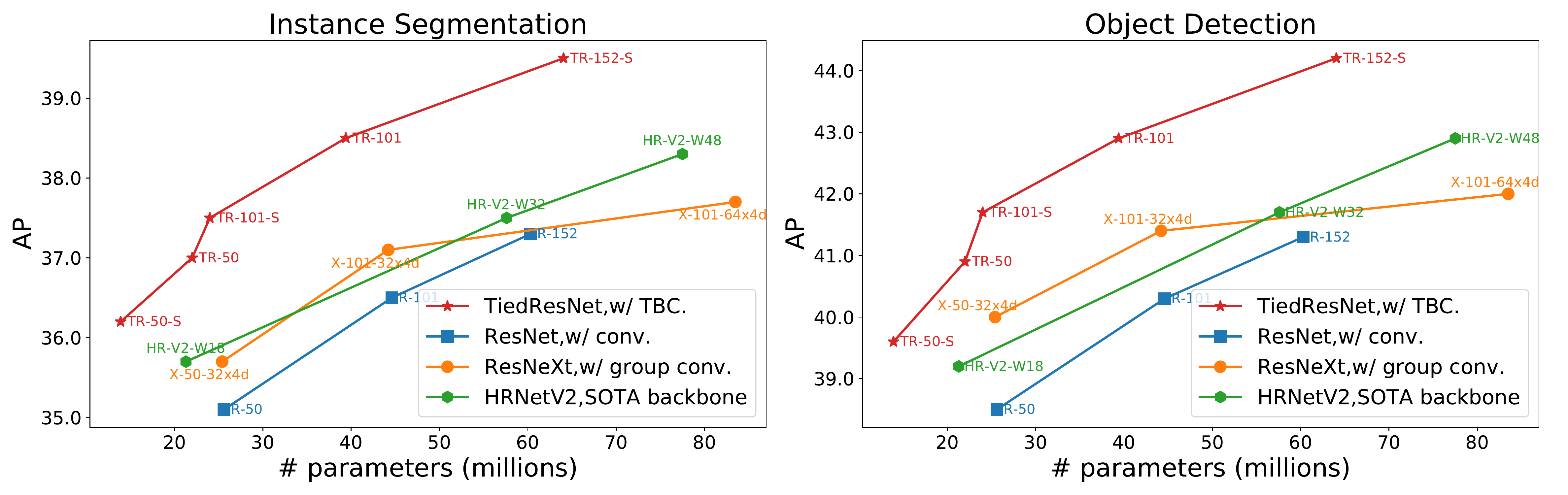}\vspace{-6pt}
    \caption{Average Precision (AP) vs. \#Parameter of backbones for various versions of Mask R-CNN on MS-COCO.  Our TiedResNet (using TBC) significantly outperforms ResNet (using SC),  ResNeXt \cite{xie2017aggregated} (using GC) and HRNetV2 \cite{wang2019deep} (current SOTA detection, segmentation and pose estimation backbone) in \textit{both} object detection and instance segmentation tasks with a much leaner model.  Baselines are copied from mmdetection \cite{mmdetection} model zoo.}
    \label{fig:MaskrcnnComp}
\end{figure*}
}

\def\figMsCoCoAll#1{
\begin{figure*}[#1]
    \centering
    \includegraphics[width=0.98\linewidth]{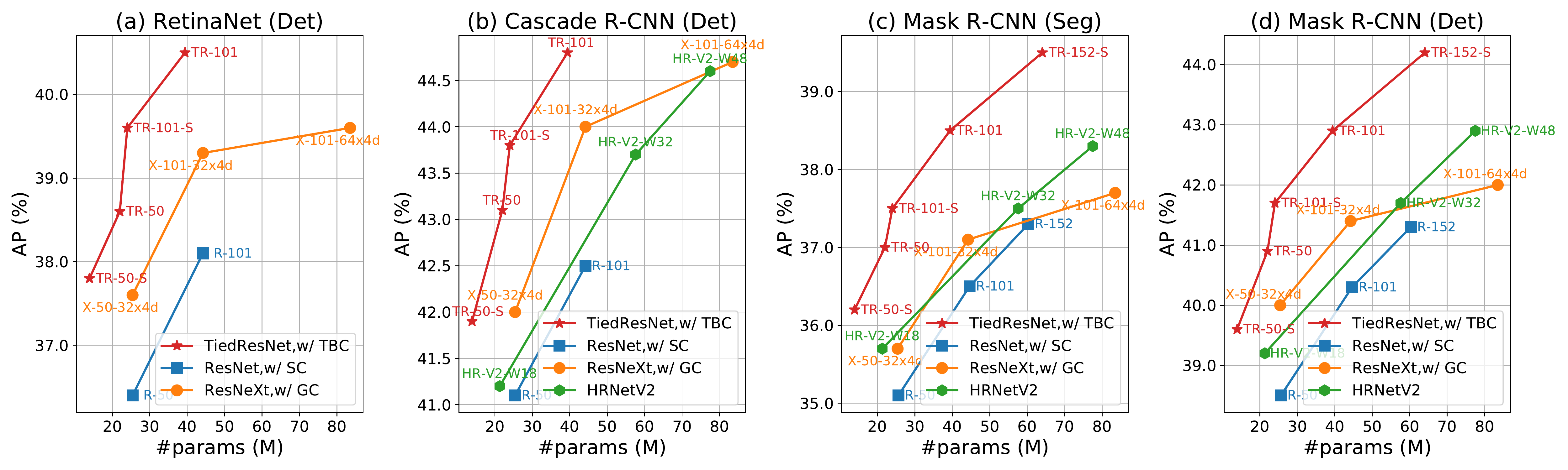}\vspace{-6pt}
    \caption{$\#$params of backbones vs. their Average Precision on \textbf{object detection and instance segmentation tasks of MS-COCO \textit{val-2017}}. For \textit{single-stage} detector RetinaNet and \textit{two-stage} detectors Cascade R-CNN and Mask R-CNN, TiedResNet consistently outperforms ResNet, ResNeXt and HRNetV2 with much fewer parameters. Detailed results are in appendix.}
    \label{fig:mscoco_params}
\end{figure*}
}

%%%%%%%%%%%%%%%% Occlusion
% \def\figOcclusion#1{
% \begin{figure}[#1]
%     \centering
%     % \begin{minipage}{.32\linewidth}
%     %   \includegraphics[width=\linewidth]{figures/Occlusion_hist.pdf}
%     %   \label{hist_occ}
%     % \end{minipage}
%     % \hspace{.02\linewidth}
%     \begin{minipage}{.49\linewidth}
%       \includegraphics[width=\linewidth]{figures/map_tiedres_vs_res.pdf}
%       \label{map_occ}
%     \end{minipage}
%     \begin{minipage}{.49\linewidth}
%       \includegraphics[width=\linewidth]{figures/ap75_tiedres_vs_res.pdf}
%       \label{ap75_occ}
%     \end{minipage}
%     \vspace{-22pt}
%     \caption{We evaluate TiedResNet and ResNet performance on images with different occlusion ratio $r$ in MS-COCO. AP and AP at IoU = 0.75 are reported. When $r$ = 0.8, TiedResNet increases by \textit{8.3\%} at AP$^{75}$ and \textit{5.9\%} at AP, much more effective at handling highly overlapping instances.}
%     \label{fig:occulusion}
% \end{figure}
% }

\def\figOcclusionWsamples#1{
\begin{figure*}[#1]
    \begin{subfigure}[][][b]{.172\textwidth}
        \centering
        \includegraphics[width=\linewidth,height=1.35\linewidth]{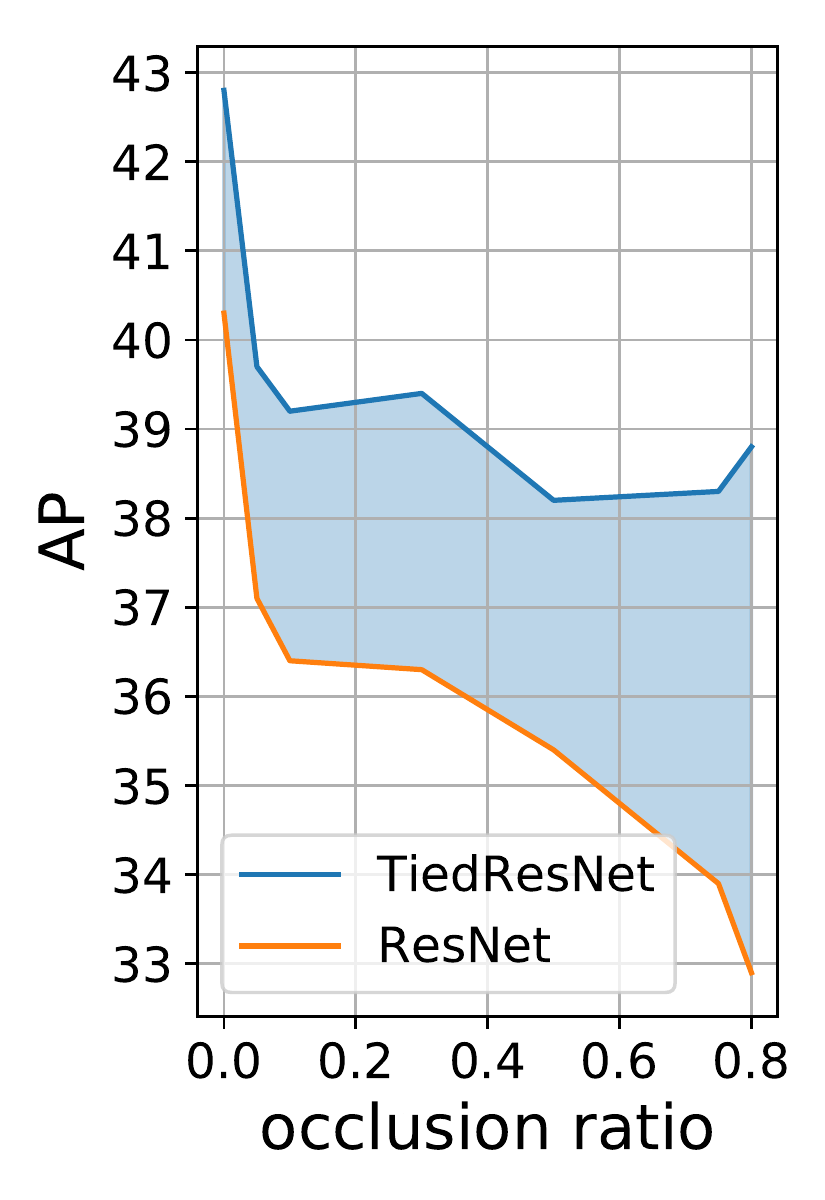}\vspace{-5pt}
        \caption{$r$ vs. AP}
        \label{fig:occulusion_ap}
    \end{subfigure}
    \begin{subfigure}[][][b]{.172\textwidth}
        \centering
        \includegraphics[width=\linewidth,height=1.35\linewidth]{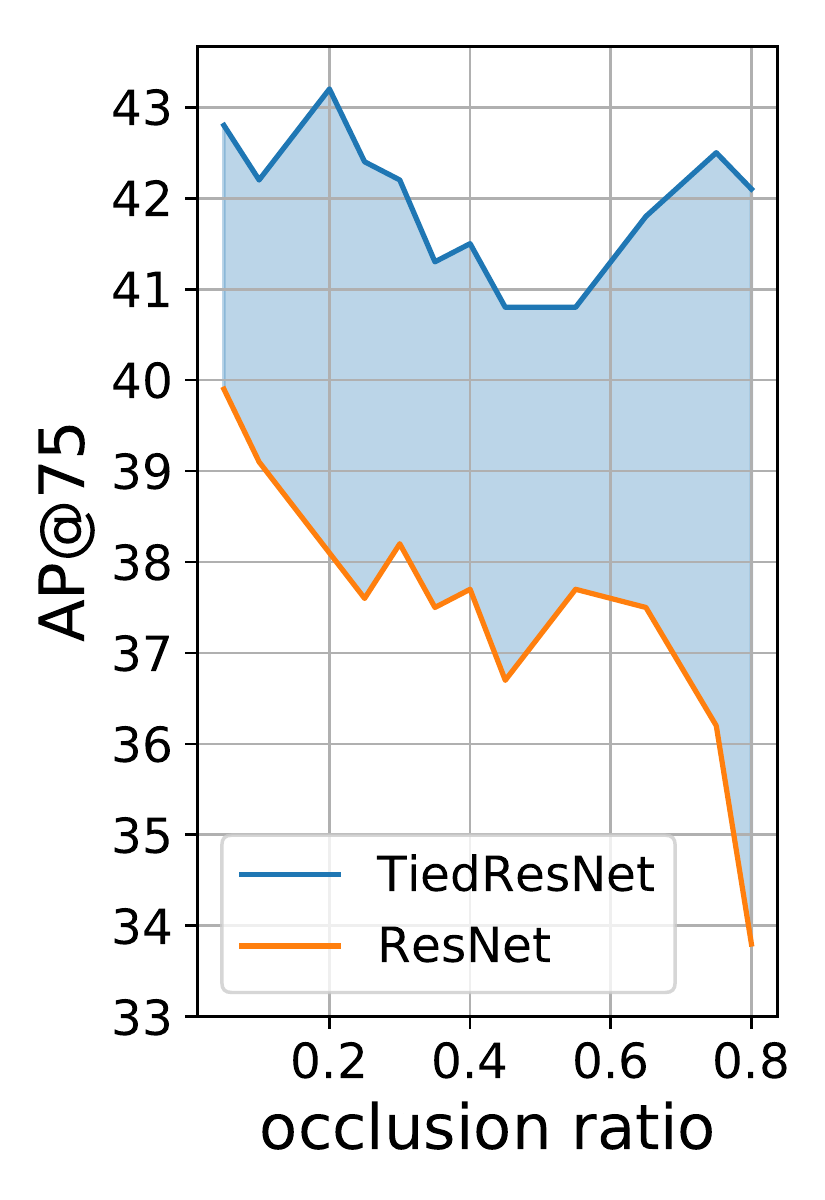}\vspace{-5pt}
        \caption{$r$ vs. AP$^{75}$}
        \label{fig:occulusion_ap75}
    \end{subfigure}
    \begin{subfigure}[][][b]{.645\textwidth}
      \centering
        \begin{subfigure}{.24\linewidth}
        \centering
        \includegraphics[width=1\linewidth,height=0.72\linewidth]{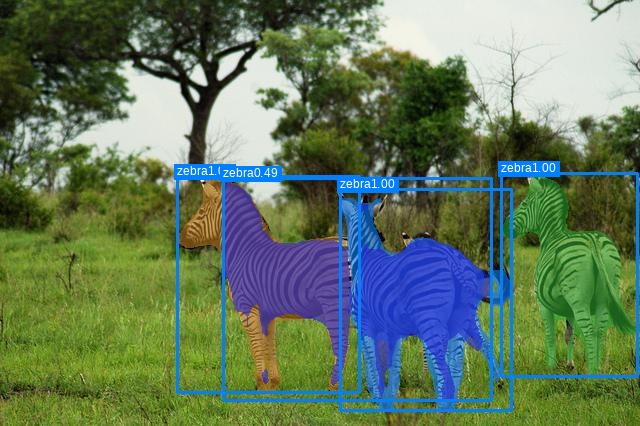}
        \end{subfigure}
        \begin{subfigure}{.24\linewidth}
          \centering
          \includegraphics[width=1\linewidth,height=0.72\linewidth]{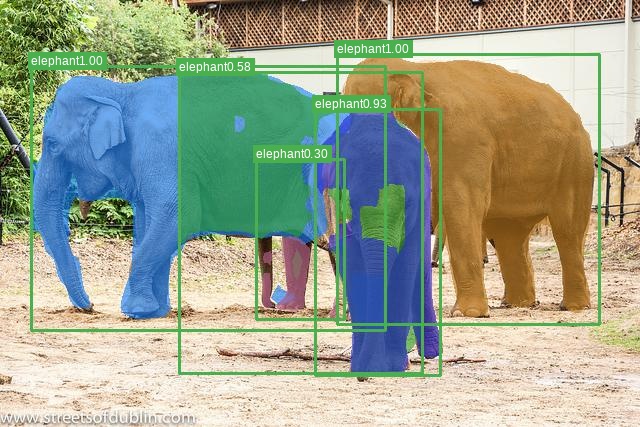}
         \end{subfigure}
        \begin{subfigure}{.24\linewidth}
          \centering
          \includegraphics[width=1\linewidth,height=0.72\linewidth]{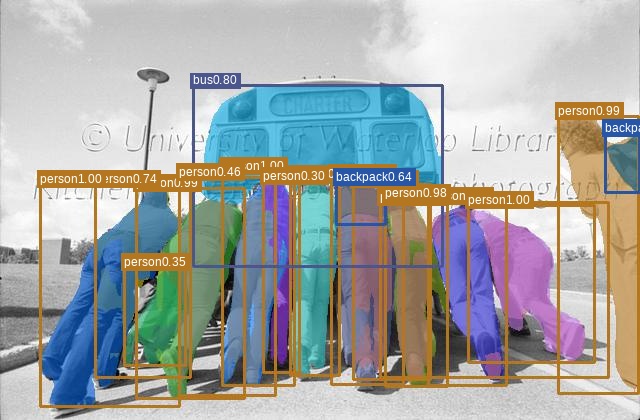}
        \end{subfigure}
        \begin{subfigure}{.24\linewidth}
          \centering
          \includegraphics[width=1\linewidth,height=0.72\linewidth]{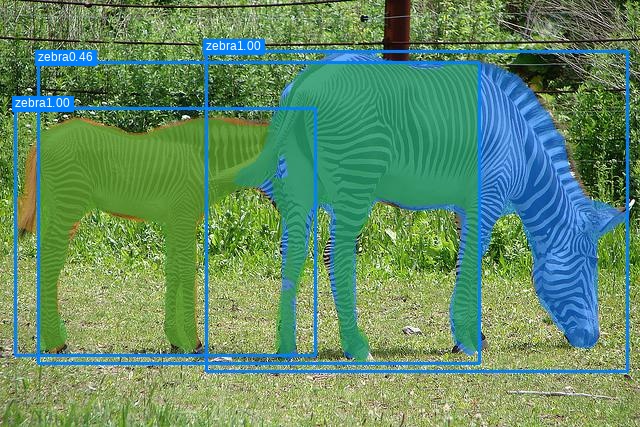}
        \end{subfigure}
        
        \begin{subfigure}{.24\linewidth}
          \centering
          \includegraphics[width=1\linewidth,height=0.72\linewidth]{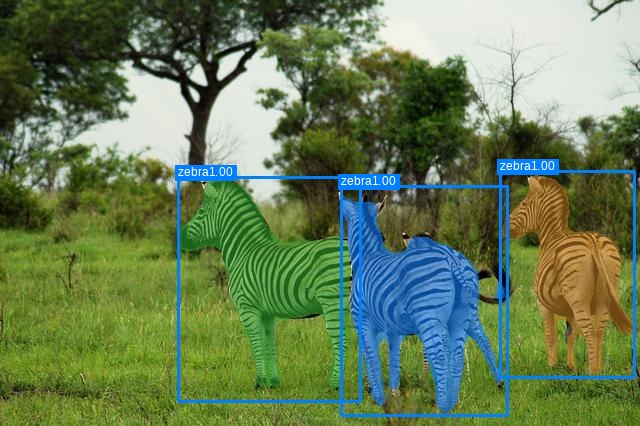}
        \end{subfigure}
        \begin{subfigure}{.24\linewidth}
          \centering
          \includegraphics[width=1\linewidth,height=0.72\linewidth]{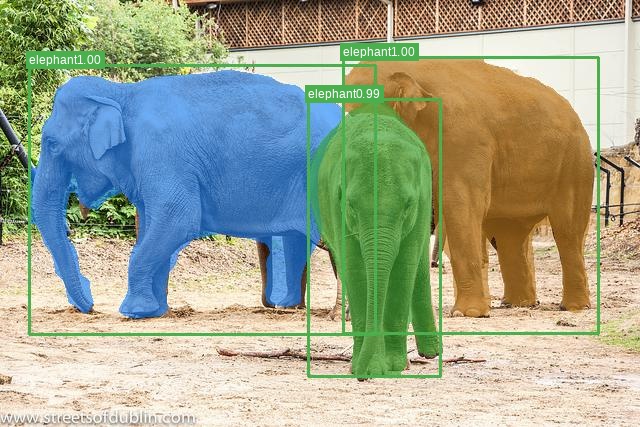}
        \end{subfigure}
        \begin{subfigure}{.24\linewidth}
          \centering
          \includegraphics[width=1\linewidth,height=0.72\linewidth]{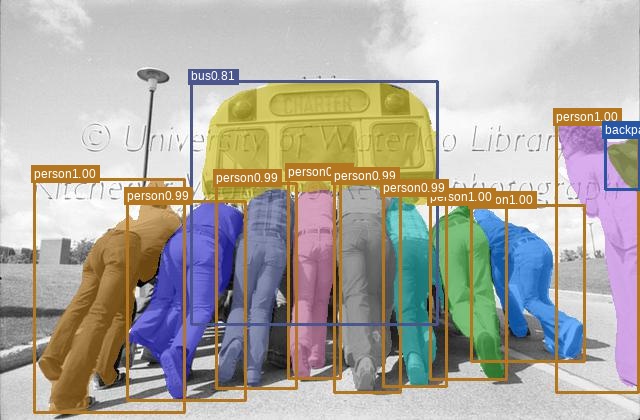}
        \end{subfigure}
        \begin{subfigure}{.24\linewidth}
          \centering
          \includegraphics[width=1\linewidth,height=0.72\linewidth]{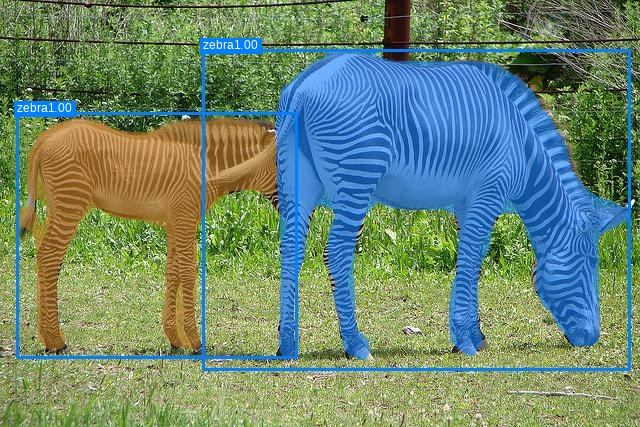}
        \end{subfigure}
    \caption{Sample results at various occlusion ratios using ResNet (row 1) and TiedResNet (row 2)}
    \label{fig:occlusion_samples}
    \end{subfigure}
\vspace{-3pt}
\caption{We evaluate TiedResNet and ResNet \textbf{performance on object detection task of MS-COCO with different occlusion ratio $r$}. AP (a) and AP at IoU = 0.75 (b) are reported. When $r$ = 0.8, TiedResNet increases by \textit{8.3\%} at AP$^{75}$ and \textit{5.9\%} at AP, much more effective at handling highly overlapping instances. (c) TiedResNet has much fewer false positive proposals, and has a significantly better instance segmentation quality. We use Mask R-CNN as the detector.}
\end{figure*}
}

%%%%%%%%%%%%%%%% VOC and Cityscapes
\def\tabVOC#1{
\begin{table*}[#1]
\tablestyle{2.5pt}{1.05}
% \scriptsize
\begin{tabular}{l||c|c|c}
\shline
Framework & Backbone & Params (M) & mAP (\%)  \\ [.1em]
\hline\hline
SSD513 \cite{liu2016ssd} & VGG16 \cite{simonyan2014very} & 138.36 & 80.6 \\
RefineDet512 \cite{zhang2018single} & VGG16 \cite{simonyan2014very} & 138.36 & 81.8 \\
R-FCN \cite{dai2016r} & ResNet101 \cite{he2016deep} & 44.65 & 80.5 \\
DSSD513 \cite{fu2017dssd} & ResNet101 \cite{he2016deep} & 44.65 & 81.5 \\
CoupleNet \cite{zhu2017couplenet} & ResNet101 \cite{he2016deep} & 44.65 & 82.7 \\
Faster R-CNN with FPN \cite{lin2017feature} & ResNet50 \cite{he2016deep} & 25.56 & 80.9 \\
Faster R-CNN with FPN \cite{lin2017feature} & ResNet101 \cite{he2016deep} & 44.65 & 82.1 \\
\hline
Faster R-CNN with FPN \cite{lin2017feature} & TiedResNet50-S & \bf{13.91} & 81.9 \\
Faster R-CNN with FPN \cite{lin2017feature} & TiedResNet50 & 22.03 & 82.6 \\
% \hline
Faster R-CNN with FPN \cite{lin2017feature} & TiedResNet101-S & 23.98 & 82.9 \\
Faster R-CNN with FPN \cite{lin2017feature} & TiedResNet101 & 39.43 & \bf{83.8}\\
\shline
\end{tabular}\vspace{-5pt}
\caption{Detection performance compared with state-of-the-art models on Pascal VOC 07 \textit{test} using voc-style evaluation metrics. With only 31$\%$ parameters, TiedResNet50-S reaches comparable performance with ResNet101.}
\label{table:voc_detection}
\end{table*}
}

\def\tabCityscapes#1{
\begin{table}[#1] % [!ht]
\tablestyle{2.5pt}{1.05}
% \scriptsize
\centering
\setlength{\tabcolsep}{6pt}
\begin{tabular}{l||c|c|c}
\shline
framework & backbone & \#params (M) & AP$^{\text{mask}}$  \\ [.1em]
\hline\hline
Mask R-CNN & ResNet50 & 25.6 & 31.5 \\
% Mask R-CNN $\dagger$ \cite{he2017mask} & ResNet50\cite{he2016deep} & 25.6 & 32.5 \\
Mask R-CNN & TiedResNet50-S & 13.9 & 32.5 \\
Mask R-CNN & TiedResNet50 & 22.0 & \bf{33.6} \\
\shline
\end{tabular}\vspace{-5pt}
\caption{Comparison on \textbf{instance segmentation task of Cityscapes} \textit{val} set and number of parameters for backbone networks, with Mask R-CNN \cite{he2017mask} as detector.}
% $\dagger$ denotes our re-implemented results.}
\label{table:cityscape_detection}
\end{table}
}

\def\tabSE#1{
\begin{table}[#1]
\tablestyle{2.5pt}{1.05}
% \scriptsize
\centering
\setlength{\tabcolsep}{6pt}
\begin{tabular}{l||c|c|c|l}
\shline
model & $B$ & top-1 (\%) & top-5 (\%) & \#params (ratio)\\ [.1em]
\hline\hline
% ResNet-50 & - & 76.15 & 92.87 & 25.56 \\
\multicolumn{5}{l}{\textit{SEResNet-50}, model params = 28.1M} \\
\hline
w/ SE & - & 76.71 & 93.38 & 2.53M {\scriptsize(100\%)} \\
w/ SE \(\ddagger\) & - & 77.08 & 93.51 & 2.53M {\scriptsize(100\%)} \\
\hline
w/ TiedSE & 2 & 77.07 & \bf{93.53} & 0.64M {\scriptsize(25\%)}\\
w/ TiedSE & 4 & \bf{77.11} & 93.52 & 0.16M {\scriptsize(6.4\%)}\\
w/ TiedSE & 8 & 77.09 & 93.52 & \bf{0.04M {\scriptsize(1.6\%)}}\\
\hline\hline
\multicolumn{5}{l}{\textit{EfficientNet-B0}, model params = 5.3M} \\
\hline
w/ SE & - & 77.1 & 93.3 & 0.65M {\scriptsize(100\%)} \\
\hline
w/ TiedSE & 2 & \bf{77.3} & \bf{93.4} & 0.16M {\scriptsize(25\%)}\\
w/ TiedSE & 4 & 77.1 & 93.3 & \bf{0.04M {\scriptsize(6.4\%)}}\\
% w/ TiedSE & 8 & 77.09 & 93.52 & \bf{25.60{\scriptsize(+0.04)}}\\
\shline
\end{tabular}\vspace{-5pt}
\caption{Comparison on \#params of \textbf{attention module SE/TiedSE} with various backbones and their recognition accuracy on ImageNet-1k.
Performance with different hyper-parameters $B$ is investigated. Using only 1.6\% (6.4\%) of the parameters, the performance of TiedSE is better than SE on SEResNet50 (EfficientNet-B0). $\ddagger$ denotes our re-implementation results.}
\label{table:imagenet_se_settings}
\end{table}
}

\def\tabGC#1{
\begin{table}[#1]
\tablestyle{2.5pt}{1.05}
% \scriptsize
\begin{tabular}{l||c|c|c|c|c|l}
\shline
framework & $B$ & AP\(^{\text{bbox}}\) & AP\(^{\text{bbox}}_{50}\) & AP\(^{\text{mask}}\)& AP\(^{\text{mask}}_{50}\)& \#params \\ [.1em]
\hline\hline
Mask R-CNN & - & 37.3 & 59.0 & 34.2 & 55.9 & - \\
\hline
% +GC\cite{cao2019gcnet} & \bf{38.9} & - & - & \bf{35.5} & - & - & 54.4{\scriptsize(+10.0)} \\
+GCB  & - & 38.9 & \bf{61.0} & 35.5 & \bf{57.6} & 10M {\scriptsize(100\%)} \\
% +GCB w/ GC  & 2 & 38.5 & 60.6 & 35.1 & 57.1 & 5.0M {\scriptsize(50\%)} \\
\hline
+TiedGCB  & 2 & \bf{39.1} & \bf{61.0} & \bf{35.6} & \bf{57.6} & 2.5M {\scriptsize(25\%)}\\
+TiedGCB  & 4 & 38.6 & 60.8 & 35.2 & 57.2 & \bf{1.3M {\scriptsize(13\%)}}\\
\shline
\end{tabular}\vspace{-5pt}
\caption{Comparison on \#params of \textbf{attention module GCB/TiedGCB} \cite{cao2019gcnet} and their performance on object detection and instance segmentation tasks of MS-COCO \textit{val-2017}. The effects of different $B$ are studied here. Result of GCB with group convolution is also compared.}
\label{table:gcnet}
\end{table}
}

\def\figinsSeg#1#2{
\begin{figure*}[#1]
\centering
\begin{subfigure}{.19\linewidth}
  \centering
  \includegraphics[width=.99\linewidth]{figures/samples/baseline/2.jpg}
\end{subfigure}
\begin{subfigure}{.19\linewidth}
  \centering
  \includegraphics[width=.99\linewidth]{figures/samples/baseline/3.jpg}
\end{subfigure}
\begin{subfigure}{.19\linewidth}
  \centering
  \includegraphics[width=.99\linewidth]{figures/samples/baseline/6.jpg}
\end{subfigure}
\begin{subfigure}{.19\linewidth}
  \centering
  \includegraphics[width=.99\linewidth]{figures/samples/baseline/8.jpg}
\end{subfigure}
\begin{subfigure}{.19\linewidth}
  \centering
  \includegraphics[width=.99\linewidth]{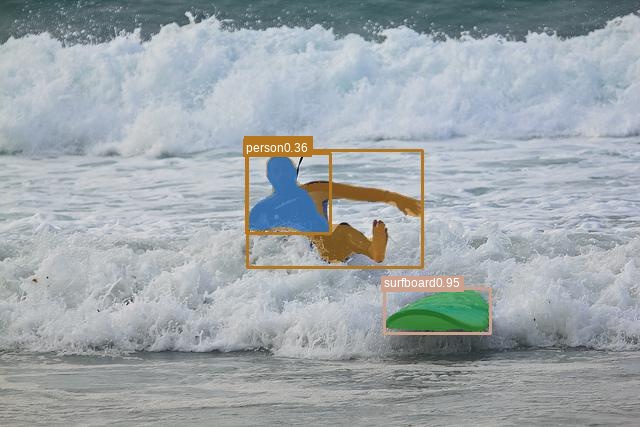}
\end{subfigure}

\begin{subfigure}{.19\linewidth}
  \centering
  \includegraphics[width=.99\linewidth]{figures/samples/ours/2.jpg}
\end{subfigure}
\begin{subfigure}{.19\linewidth}
  \centering
  \includegraphics[width=.99\linewidth]{figures/samples/ours/3.jpg}
\end{subfigure}
\begin{subfigure}{.19\linewidth}
  \centering
  \includegraphics[width=.99\linewidth]{figures/samples/ours/6.jpg}
\end{subfigure}
\begin{subfigure}{.19\linewidth}
  \centering
  \includegraphics[width=.99\linewidth]{figures/samples/ours/8.jpg}
\end{subfigure}
\begin{subfigure}{.19\linewidth}
  \centering
  \includegraphics[width=.99\linewidth]{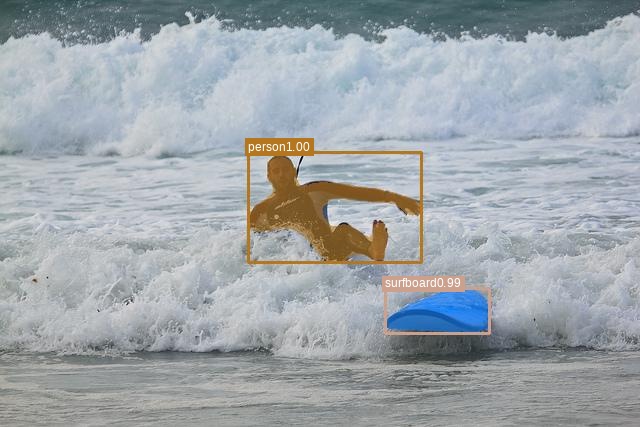}
\end{subfigure}\vspace{-3pt}
\caption{#2}
\label{fig:insSeg}
\end{figure*}
}

%****************** Experiments ***********************
%******************************************************
\section{Experimental Results}
\label{experiments}

We conduct extensive tests of TBC, TGC and TFC on major benchmarks for object recognition, object detection, instance segmentation and attention.
\tabImageNet{tbp}
\figMsCoCoAll{!t}

\subsection{ImageNet Classification}

% We implement our code in Pytorch. Both ResNet baseline and TiedResNet are trained with identical optimization schemes. We follow standard practices and perform data augmentation with random cropping using scale and aspect ratio to a size of 224 \(\times\) 224 pixels and perform random horizontal flipping as in \cite{krizhevsky2012imagenet,he2016deep}. We train the network using SGD with weight decay 0.0001 and a momentum of 0.9, and a mini-batch of 512 on 8 GPUs. The learning rate is initially set to 0.2 and divided by 10 every 30 epochs. All models are trained for 100 epochs with the same training and data argumentation strategy. Baselines are copied from Pytorch official model zoo \cite{paszke2019pytorch}.

\noindent\textbf{Implementation.} We follow standard practices and perform data augmentation with random cropping to size 224\(\times\)224 pixels \cite{he2016deep}.  We train the network using SGD with a momentum of 0.9 and a mini-batch of 256 on 8 GPUs. The learning rate is initially set to 0.1 and then decayed 10$\times$ every 30 epochs for a total of 100 epochs.

\noindent\textbf{Performance gain.} Table \ref{table:imagenet_models} compares the recognition accuracy of multiple models on ImageNet-1k \cite{deng2009imagenet} \text{validation} set. In Table \ref{table:imagenet_models}, TiedResNet50-S beats ResNet50 in terms of top-1 accuracy with only 60\% flops and 54\% parameters, likewise for TiedResNet101-S. With similar model complexity, TiedResNet50 and TiedResNet101 can beat benchmarks by more than 1.5\% and 1.4\% separately with 10\% parameter reduction. Similar tendency can be observed for TiedResNeXt and TiedSENet. To further prove the effectiveness of TBC, we integrate it with current SOTA model ResNeSt. With only 59\% of parameters and 82\% of computation cost, TiedResNeSt-50-S obtains better performance than ResNeSt-50-S on ImageNet-1k. 

% Although DenseNet161\cite{huang2017densely} can also reach better performance than ResNet101 with fewer parameters, its computational cost remains almost the same.  At a similar accuracy, our TiedResNet101-S model can outperform Densenet161 with 40\% multiply-add reduction and another 16\% parameter reduction.

% With similar model complexity, we increase the base width of our model from 18 to 32. TiedResNet50 and TiedResNet101 can beat ResNet50 and ResNet101 by more than 1.5 and 1.4\% separately with $\sim$10$\%$ parameter reduction. They also outperform ResNeXt counterparts.

% \tabDetectionTest{!t}
% \tabDetectionVal{!t}

\subsection{Object Detection and Instance Segmentation}

\subsubsection{MS-COCO} \cite{lin2014microsoft} consists of 80 object categories with 118K/5K/208K images for training (\textit{train-2017}), validation (\textit{val-2017}) and testing (\textit{test-2017}) respectively. Average Precision (AP) across IoU thresholds from 0.5 to 0.95 with an interval of 0.05 is evaluated. Detection performance at various qualities, AP$_{50}$ and AP$_{75}$, and at different scales, AP$_\text{S}$, AP$_\text{M}$ and AP$_\text{L}$, are reported.  All models are trained on \textit{train-2017} split and  results reported on \textit{val-2017}. 
%For main results, we report officially evaluated AP on \textit{test-2017}.

\noindent\textbf{Implementation.} We use baseline backbones and our TiedResNet model in PyTorch implemented \cite{mmdetection} detectors.  The long  and short edges of images are resized to a maximum of 1333 and 800 respectively without changing the aspect ratio.  Since 1$\times$ learning schedule (LS) is under-sutured, we only report results on 2$\times$ LS for both baselines and our models.

% \noindent\textbf{Optimization.}
% Since 1$\times$ learning schedules (LS) is under-sutured, we only report results on 2$\times$ LS for both baselines and our models. As in \cite{mmdetection}, initial learning rate (LR) is set as 0.02 for the first 16 epochs and decay 10 times at epoch 18 and 22 with 24 total epochs for 2$\times$ LS. Cascade R-CNN and Cascade Mask R-CNN \cite{cai2018cascade} use 20e learning rate schedule as enhanced training schedule which decay learning rate at epoch 16 and 19 with 20 total training epochs. 

% All the corresponding models' and their counterparts' hyper-parameters and data augmentation methods will remain the same as default settings in \cite{ren2015faster,he2017mask,mmdetection,lin2017focal,cai2018cascade} for fair comparison.

% \figMsCoCoDetector{t!}
% \figMaskrcnnComp{!t}

\noindent\textbf{Results.} We conduct thorough comparisons with ResNeXt and ResNet on multiple state-of-the-art frameworks including single-stage detector, RetinaNet \cite{lin2017focal}, and two-stage detectors and Mask R-CNN \cite{he2017mask} as
% , Cascade R-CNN \cite{cai2018cascade}
% , Cascade Mask R-CNN \cite{cai2018cascade} and Hybrid Task Cascade \cite{chen2019hybrid} 
in Fig.\ref{fig:mscoco_params}.
% and Table \ref{table:coco_detection_instance_seg_val}.  
Since \cite{mmdetection} re-implemented results are generally better than those  in the original papers, we report re-implemented results for fair comparisons.

\noindent\textbf{Object detection.}  As in Fig.\ref{fig:mscoco_params}, using TiedResNet as backbone, single-stage detector RestinaNet and two-stage detector Cascade R-CNN and Mask R-CNN consistently outperform baselines by 2$\%$ to $2.5\%$ in terms of box AP. TiedResNet101 on RetinaNet even greatly outperforms the much heavier-weight ResNeXt101-64$\times4$d. Detailed comparison on various frameworks and Pascal VOC \cite{Everingham15} are in appendix materials. % and 2.3$\%$ to 3.2$\%$ for AP$_\text{L}$.

\noindent\textbf{Instance segmentation.} With light-weight TiedResNet-S and comparable sized TiedResNet backbones, we observe an increase in AP$^\text{mask}$ by 1.1\% and 2.1\% respectively. No matter how strong the baseline detector is, we always observe a boost in AP, corroborating the effectiveness of TBC.

% \noindent\textbf{Results on MS-COCO \textit{val}.} With only 55$\%$ model size and 60$\%$ computation cost, our TiedResNet101-S model outperforms ResNet101 by more than 2\% AP$^\text{bbox}$ and 1.8\% AP$^\text{mask}$ in Cascade Mask R-CNN, Hybrid Task Cascade and Mask R-CNN experiments, with significantly fewer parameters. TiedResNet101-S even beats ResNeXt101-32$\times$4d on MS-COCO detection and segmentation tasks.  With TiedResNet-101, we achieve 0.7$\%$$\sim$1.1$\%$ and 2.0$\%$$\sim$2.2$\%$ AP$^\text{bbox}$ increase compared with ResNeXt101-32$\times$4d and 0.6$\%$$\sim$0.9$\%$ and 1.6$\%$ $\sim$1.8$\%
% $ AP$^\text{mask}$ compared with ResNet101.

% \subsubsection{4.2.2. Analysis.}
% \figOcclusion{t!}

\figOcclusionWsamples{!t}

\noindent\textbf{Highly occluded Instances.} Since occlusion requires the network to accurately detect the target area and distinguish different instances at the same time, the performance on images with large occlusion reveals the network's localization capabilities.  
The occlusion ratio ($r$) of each image is: 
\begin{align}
r=\frac{\text{total overlap area}}{\text{total instance area}}
\end{align}
% Fig.\ref{fig:occulusion_ap} and Fig.\ref{fig:occulusion_ap75} shows the distribution of images in COCO \textit{val-2017} in terms of the occlusion ratio. 
 % Only the images with occlusion ratio larger than $r$ are evaluated. 
%When $r=0$, the model is evaluated on all testing samples.

% \figinsSeg{t!}{Sample results at various occlusion ratios using ResNet (Row 1) and TiedResNet (Row 2).  TiedResNet has much fewer false positive proposals, and has a significantly better instance segmentation quality.}

The AP averaged over IoU 0.5 to 0.95, and 
at IoU=0.75, AP$^{75}$, are used as standard and restricted evaluation metrics respectively. Fig.\ref{fig:occulusion_ap} and Fig.\ref{fig:occulusion_ap75} shows that ResNet is greatly affected by occlusion, AP$^{75}$ drops by more than 6\% at $r=0.8$, whereas our TiedResNet only slightly decreases by 0.7\%, exceeding the baseline of \textit{8.3\%}.  Similarly, as the occlusion rate becomes larger, the improvement on AP increases from 2.8\% to \textit{5.9\%}. These quantitative results in MS-COCO indicate TiedResNet's strong capability of handling highly overlapping instances, especially on restricted evaluation metric.
Fig.\ref{fig:occlusion_samples} shows that TiedResNet has fewer false positive proposals and better segmentation quality.

\noindent\textbf{Why larger gain on single-stage detector?} 
 Fig.\ref{fig:gradcam} shows that TiedResNet localizes the target area much better than ResNet/ResNeXt, which is especially beneficial for a single-stage detector that does not has a proposal regression layer. 
 
%%%%%%%%%%%%%%%% Instance Segmentation Visualization
% \noindent\textbf{Instance Segmentation Visualization.} 
% \tabVOC{t!}
\tabCityscapes{t!}

%%%%%%%% Results on VOC. Results on Cityscapes.
\noindent\textbf{Performance on Cityscapes.} Since Cityscapes \cite{cordts2016cityscapes} is a small dataset, thus deeper networks will generally overfit it. Therefore, we only deploy experiments with 50 layers backbone for Cityscapes datasets. Table \ref{table:cityscape_detection} shows that TiedResNet50 can reach 2.1\% gain for AP$^{\text{mask}}$.

\tabSE{t}
\tabGC{t}
%%%%%%%% Lightweight Attention
\subsection{Lightweight Attention}
\fig{atten} shows our lightweight attention modules. The SE module can be seen as a special case of our TiedSE when $B\!=\!1$; likewise, GCB is TiedGCB at $B\!=\!1$.

%%%%%%%% TiedSE
\noindent\textbf{Results of TiedSE.} All experiments in Table \ref{table:imagenet_se_settings} use reduction ratio of 16 for both baseline and our model. Several hyper-parameter settings of our TFC layer are investigated.  
Since our re-implemented baseline results are better than those in \cite{hu2018squeeze}, we report our results for fair comparison.  
While SE is light weight, it still incurs  10$\%$ parameters of overall model. Table \ref{table:imagenet_se_settings} shows that,
% that our TiedSE reaches similar performance with much fewer parameters than SE.  
% The best performance is reached at $B$=4, which decreases parameters by more than 16$\times$.  
at $B$=8, with 64$\times$ parameters reduction, TiedSE still obtains comparable performance. TiedSE significantly reduces parameters without sacrificing performance not only on SEResNet but also on Mobile architecture EfficientNet \cite{tan2019efficientnet}.% compared with ResNet50 counterpart

%%%%%%%% TiedGC
\noindent\textbf{Results of TiedGCB.} Global context blocks (GCB) \cite{cao2019gcnet} enhance segmentation and detection predictions with global context modeling and long-range dependencies. 
% However, GCB increases model parameters by more than 10M, at 23$\%$ relative increase.
GCB integrated with TBC can significantly reduce the number of parameters without losing performance.  
Table \ref{table:gcnet} shows that TiedGCB achieves 1.8\% and 1.4\%  gain in AP$^\text{mask}$ and AP$^\text{bbox}$ respectively,  with 16$\times$ parameters reduction.  
Although group convolution can reduce parameters by 2$\times$, as each GC filter only sees a subset of features, the ability to model cross-channel dependencies is also reduced, losing AP$^\text{mask}$ and AP$^\text{bbox}$ by $0.4\%$.

% \figGradCam{t!}
\figGradCamSingle{!t}
% \tabSplitsFusion{!t}
\tabSplits{!t}
\tabFusion{!t}

\subsection{Ablation Studies}
\noindent\textbf{Influence of split number.} As investigated in \cite{zeiler2014visualizing,bau2017network,xu2015show}, the proportions of units/filters that correspond to various visual concepts, such as color, texture, objects, part, scene, edge and material, are different with a variety of levels of interpretability \cite{agrawal2015learning,bau2017network}.  It may be useful to group different functional filters  together for different levels of sharing.  In Table \ref{table:ablation_bconv_settings}, we split all the channels in  the 3$\times$3 convolutional layer into $s$ splits. Each split has base width of $w$, and $B$ is 1,2,4,8 separately for the four 3$\times$3 TBC layers in $4s\times32 w$ settings.  In Table \ref{table:ablation_bconv_settings}, the best performance and model complexity trade-off can be reached at $4s\times32w$.  Table \ref{table:ablation_bconv_settings} also shows the necessity of splitting input feature maps into several chunks, when there are only 2 splits, top-1 accuracy will drop 0.4\%.

\noindent\textbf{Mixer module in TiedBottleneck.} Since we split the input feature map into several splits, the inter-dependency across these splits is missed.  To track the inter-dependency, a mixer is used to aggregate cross-split information.  Several fusion methods are investigated in Table \ref{table:ablation_bconv_mixer}.  Using concatenation reaches the best accuracy, but it introduces much more parameters.  We thus choose elementwise-sum as the fusion function as a trade-off between accuracy and model size.

% \noindent\textbf{\textit{v.s.} model compression and pruning.} Several state-of-the-art model compression and pruning methods are also compared with our ResNet50-S model in Table \ref{table:vs_compression}, trick enhanced \cite{liu2018rethinking} results are also reported. Our model reduces 15$\%$ parameters with better accuracy. Compared with these pruning and model compression methods, our model is naturally smaller and therefore can be end-to-end trained from scratch, which releases the burden of manually choosing thresholds, pruning rates and sensitivity parameters or conducting longer training and fine-tuning process to reach better performance.

\figFilterSim{t!}

\noindent\textbf{Filter similarity.} We use ImageNet pre-trained ResNet50 and TiedResNet50-S to compare the cosine filter similarity at different layers. Pairwise cosine similarity between filters' guided back-propagation patterns \cite{springenberg2014striving} averaged in 1000 ImageNet $\textit{val}$ split are used to generate these histograms. As in Figure \ref{fig:hist_tiedres}, axis $x$ is the cosine similarity and axis $y$ is the probability density.  Compared with VGG\cite{simonyan2014very}, ResNet\cite{he2016deep} has less redundancy, and our TiedResNet has the least similarity and thus removes most redundancy throughout the depth layers, which validates our hypothesis and motivation.

\noindent\textbf{Grad-CAM visualization.} 
To provide a qualitative comparison among different backbone networks, we apply grad-CAM \cite{selvaraju2017grad}  using images from ImageNet.  Grad-CAM uses the gradient information flowing into the last convolutional layer of the CNN to understand each neuron. The resulting localization map highlights important regions in the image for predicting the concept and reflects the network's ability to utilize information in the target object area. 
Fig.\ref{fig:gradcam} shows TiedResNet focusing on target objects more properly than ResNet and ResNetX, suggesting that the performance boost comes from accurate attention and noise reduction of irrelevant clutters.

This property is very useful for object detection and instance segmentation, as these tasks require the network to focus more accurately on the target region and aggregate features from it.  Incorrect attention to the target area will also lead to a large number of false positive proposals (Fig.\ref{fig:occlusion_samples}).  
% See discussions on detection and segmentation next.
% We obtained \textit{8.3\%} above the baseline on the test samples with occlusion rate higher than 0.8 on MS-COCO as in Fig.\ref{fig:occulusion}.

%% file: 5summary
% \vspace{-4pt}
\section{Summary}
% \subsection{Summary}
We propose Tied Block Convolution(TBC) that shares the same thinner filter over equal blocks of channels and produces multiple responses with a single filter. 
The concept of TBC can also be extended to group convolution and fully connected layers, and can be applied to various backbone networks and attention modules, with consistent performance improvements to the baseline. TBC-based TiedResNet also surpasses baselines with much higher parameter usage efficiency and better capability of detecting objects under severe occlusion.

% In object detection and instance segmentation tasks, the TBC based TiedResNet outperforms both group convolution based ResNeXt and standard convolution based ResNet, with much more efficient parameter usage and much better ability in under heavy occlusion.

% We propose tied block convolution (TBC), tied group convolution (TGC) and tied fully connected (TFC) layers which could reduce parameters by $B^2$ times with 
% which divides $C$ input channels into $B$ blocks and produces $B$ responses with a single thinner filter applied to each block of $\frac{C}{B}$ channels.  
% Consistent performance improvements over baselines can be obtained by integrating them into various backbone networks with.  
% It reaches state-of-the-art performance on multiple major tasks and datasets; it

%% file: arxiv_sup
\renewcommand{\thefigure}{A.\arabic{figure}}
\renewcommand{\thetable}{A.\arabic{table}}

\def\tabDetector#1{
\begin{table*}[#1]
\tablestyle{4.pt}{1.0}
% \scriptsize
\setlength{\tabcolsep}{10pt}
\begin{tabular}{l|c|cccc|cccc}
\shline
\multirow{2}{*}{Backbone} & \multirow{2}{*}{$\#$param} & \multicolumn{4}{c|}{\footnotesize{LS=1$\times$}} &\multicolumn{4}{c}{LS=2$\times$/20e} \\\cline{3-10}
&& AP&AP$_{S}$&AP$_{M}$&AP$_{L}$ & AP&AP$_{S}$&AP$_{M}$&AP$_{L}$\\ [0.1em]
\shline
\multicolumn{10}{c}{RetinaNet \cite{lin2017focal}} \\
\hline
ResNet-50  & 25.6M & 35.6&20.0&39.6&46.8 & 36.4&19.3&39.9&48.9 \\
% RetinaNet\cite{lin2017focal} & TiedResNet-50-S & 12.59 & ${\color{red}TODO3}$ &  &  & - & - & - \\
TiedResNet-50-S  & 13.9M & \bf{36.8}&\bf{21.0}&\bf{40.8}&\bf{48.1} & \bf{37.5}&\bf{20.3}&\bf{41.2}&\bf{50.1} \\
\rowcolor{Gray}
\textit{vs. baseline} & $\downarrow$11.7M & {+1.2}&{+1.0}&{+1.2}&{+1.3} & {+1.1}&{+1.0}&{+1.3}&{+1.2} \\
TiedResNet-50  & 22.0M & \bf{37.7}&\bf{21.3}&\bf{41.8}&\bf{49.5} & \bf{38.6}&\bf{21.3}&\bf{42.3}&\bf{51.5} \\
\rowcolor{Gray}
\textit{vs. baseline} & $\downarrow$3.6M & {+2.1}&{+1.3}&{+2.2}&{+2.7} & {+2.2}&{+2.0}&{+2.4}&{+2.6} \\
\hline
ResNet-101  & 44.6M & 37.7&21.1&42.2&49.5 & 38.1&20.2&41.8&50.8 \\
% RetinaNet\cite{lin2017focal} & TiedResNet-50-S & 12.59 & ${\color{red}TODO3}$ &  &  & - & - & - \\
TiedResNet-101-S  & 24.0M & \bf{39.8}&\bf{22.4}&\bf{44.4}&\bf{52.2} & \bf{39.1}&\bf{20.9}&\bf{42.6}&\bf{52.1} \\
\rowcolor{Gray}
\textit{vs. baseline} & $\downarrow$20.6M & {+1.2}&{+0.6}&{+1.0}&{+1.8} & {+1.0}&{+0.7}&{+0.8}&{+1.3} \\
TiedResNet-101  & 39.4M & \bf{39.8}&\bf{22.4}&\bf{44.4}&\bf{52.2} & \bf{40.5}&\bf{22.9}&\bf{44.8}&\bf{52.7} \\
\rowcolor{Gray}
\textit{vs. baseline} & $\downarrow$5.2M & {+2.2}&{+1.4}&{+2.0}&{+2.8} & {+2.4}&{+2.7}&{+3.0}&{+1.9} \\
% \hline\hline
% \hline\hline
% \multicolumn{10}{c}{Faster R-CNN with FPN \cite{ren2015faster,lin2017feature}} \\
% % \hline
% % R-50  & 25.6M & 36.4&21.5&40.0&46.6 & - \\
% % % Faster R-CNN w& FPN\cite{lin2017feature} & TiedResNet-50-S & 12.59 & 38.8(+) & 60.6 & 42.1 & - & - & - \\
% % BR-50  & 22.0M & \bf{38.8}&\bf{23.1}&\bf{42.8}&\bf{49.9} & - \\
% % \textit{vs. baseline} & {$\downarrow$3.6M} & {{+2.4}&{+1.6}&{+2.8}&{+3.3}} & - \\
% \hline
% % X101-64\(\times\)4 & & 88.8M & 40.7&22.9&44.5&53.6 & - \\
% % X101-32\(\times\)4 & & 44.2M & 40.1&23.4&44.6&51.7 & - \\
% ResNet-101  & 44.6M & 38.5&22.3&43.0&49.8 & - \\
% TiedResNet-101  & 39.4M & \bf{41.0}&\bf{24.0}&\bf{45.4}&\bf{53.0} & - \\
% \textit{vs. baseline} & {$\downarrow$5.1M} & {+2.5}&{+1.7}&{+2.4}&{+3.2} & - \\
% \shline
\shline
\hline\hline
\multicolumn{10}{c}{Cascade R-CNN \cite{cai2018cascade}} \\
\hline
ResNet-101  & 44.6M & 40.4&21.5&43.7&53.8 & 42.5&23.7&46.1&56.9 \\
% Cascade R-CNN\cite{he2017mask}  & TiedResNet-50-S & 12.59 & 81.3(+1.4) & & & - & - & - \\
TiedResNet-101-S  & 24.0M & \bf{41.5}&\bf{22.5}&\bf{44.9}&\bf{54.7} & \bf{43.8}&\bf{24.8}&\bf{47.4}&\bf{58.4} \\
\rowcolor{Gray}
\textit{vs. baseline} & {$\downarrow$22.6M} & {+1.1}&{+1.0}&{+1.2}&{+0.9} & {+1.3}&{+1.1}&{+1.3}&{+1.5} \\
TiedResNet-101  & 39.4M & \bf{42.7}&\bf{22.8}&\bf{46.6}&\bf{57.0} & \bf{44.8}&\bf{25.3}&\bf{49.0}&\bf{59.2} \\
\rowcolor{Gray}
\textit{vs. baseline} & {$\downarrow$5.2M} & {+2.3}&{+1.3}&{+2.9}&{+3.2} & {+2.3}&{+1.6}&{+2.9}&{+2.3} \\
\shline
\end{tabular}
\caption{\#params of backbones vs. their Average Precision on object detection task of MS-COCO \textit{val-2017}. LS denotes learning schedule. Baseline are obtained from \cite{mmdetection}. We experiment TBC on various detectors, including single stage detector RetinaNet \citep{lin2017focal} and SOTA two-stage detector Cascade R-CNN \citep{cai2018cascade}. Consistent performance improvements can be observed. Reported results are used to plot Figure 6(a) and Figure 6(b) of our submission.}
\label{table:coco_detection_sup}
\end{table*}
}

\def\tabDetSeg#1{
\begin{table*}[#1]
\tablestyle{4.pt}{1.0}
% \scriptsize
\setlength{\tabcolsep}{8pt}
\begin{tabular}{l|c|c|cccc|cccc}
\shline
\multirow{2}{*}{Backbone} & \multirow{2}{*}{LS} & \multirow{2}{*}{$\#$param} & \multicolumn{4}{c|}{\footnotesize{Object Detection}} &\multicolumn{4}{c}{Instance Segmentation} \\\cline{4-11}
&&& AP&AP$_{S}$&AP$_{M}$&AP$_{L}$ & AP&AP$_{S}$&AP$_{M}$&AP$_{L}$\\ [0.1em]
\shline
\multicolumn{11}{c}{Mask R-CNN \cite{he2017mask}} \\
% \hline
% ResNet-50 & 1\(\times\) & 25.6M & 37.3&21.9&40.9&48.1 & 34.2&15.8&36.9&50.1 \\
% % BR-50-S & 1\(\times\) & \bf{13.9M} & 38.1&22.0&41.8&49.5 & 34.9&16.1&37.8&51.2 \\
% % \textit{vs. baseline} & - & {$\downarrow$11.7M} & {{+0.8}&{+0.1}&{+0.9}&{+1.4}} & {{+0.7}&{+0.3}&{+0.9}&{+1.1}} \\
% % Mask R-CNN\cite{he2017mask} & ResNeXt-50\cite{xie2017aggregated} & 25.02 & 38.6 & 22.7 & 42.0 & 48.9 & 35.5 & 16.6 & 37.9 & 50.9 \\
% TiedResNet-50 & 1\(\times\) & 22.0M & \bf{39.8}&\bf{24.3}&\bf{44.0}&\bf{51.2} & \bf{36.3}&\bf{17.9}&\bf{39.5}&\bf{52.6} \\
% \textit{vs. baseline} & - & {$\downarrow$3.6M} & {+2.5}&{+2.4}&{+3.1}&{+3.1} & {+2.1}&{+2.1}&{+2.6}&{+2.5} \\
\hline
% 2X-50 & & 25.0M & 40.0&23.1&43.0&52.1 & 35.9&17.3&38.8&52.8 \\
ResNet-50 & 2\(\times\) & 25.6M & 38.5&22.6&42.0&50.5 & 35.1&16.7&37.7&52.0 \\
TiedResNet-50-S & 2\(\times\) & \bf{13.9M} & 39.6&23.0&43.3&51.3 & 36.2&17.0&38.8&52.8 \\
\rowcolor{Gray}
\textit{vs. baseline} & - & {$\downarrow$11.7M} & {+1.1}&{+0.4}&{+1.3}&{+0.8} & {+1.1}&{+0.3}&{+1.1}&{+0.8} \\
TiedResNet-50 & 2\(\times\) & 22.0M & \bf{40.9}&\bf{24.0}&\bf{44.7}&\bf{53.6} & \bf{37.0}&\bf{17.4}&\bf{39.9}&\bf{54.6} \\
\rowcolor{Gray}
\textit{vs. baseline} & - & {$\downarrow$3.6M} & {+2.4}&{+1.4}&{+2.7}&{+3.1} & {+1.9}&{+0.7}&{+2.2}&{+2.6} \\
\hline
ResNet-101 & 2\(\times\) & 44.6M & 40.3&22.2&44.8&52.9 & 36.5&16.3&39.7&54.6 \\
TiedResNet-101-S & 2\(\times\) & \bf{24.0M} & 41.7&24.1&45.8&54.3 & 37.5&17.9&40.5&55.0 \\
\rowcolor{Gray}
\textit{vs. baseline} & - & {$\downarrow$20.6M} & {+1.4}&{+1.9}&{+1.0}&{+1.4} & {+1.0}&{+1.6}&{+0.8}&{+0.4}\\
TiedResNet-101 & 2\(\times\) & 39.4M & \bf{42.8}&\bf{24.2}&\bf{46.8}&\bf{57.2} & \bf{38.4}&\bf{18.2}&\bf{41.5}&\bf{57.0} \\
\rowcolor{Gray}
\textit{vs. baseline} & - & {$\downarrow$5.2M} & {+2.5}&{+2.0}&{+2.0}&{+4.3} & {+1.9}&{+1.9}&{+0.8}&{+3.4}\\
\shline
SENet-101 & 2\(\times\) & 49.1M & 41.1&23.3&45.6&54.5 & 37.2&17.3&40.2&55.1 \\
TiedSENet-101-S & 2\(\times\) & \bf{22.8M} & 42.4&24.9&46.6&55.7 & 38.2&18.6&41.1&56.1 \\
\rowcolor{Gray}
\textit{vs. baseline} & - & {$\downarrow$5.2M} & {+1.3}&+1.6&+1.0&+1.2 & {+1.0}&+1.3&+0.9&+1.0 \\
TiedSENet-101 & 2\(\times\) & 41.8M & \bf{43.4}&\bf{25.3}&\bf{48.2}&\bf{58.4} & \bf{38.9}&\bf{19.1}&\bf{42.2}&\bf{58.3} \\
\rowcolor{Gray}
\textit{vs. baseline} & - & {$\downarrow$7.3M} & {+2.3}&+2.0&+2.6&+3.9 & {+1.7}&+1.8&+2.0&+3.2 \\
\shline
ResNeXt-101-32$\times$8d & 2\(\times\) & 88.8M & 42.8&24.5&46.9&55.7 & 38.3&16.7&37.7&52.0 \\
TiedResNeXt-101-32$\times$8d & 2\(\times\) & \bf{64.0M} & \bf{44.0}&\bf{26.0}&\bf{47.6}&\bf{56.5} & \bf{39.2}&\bf{17.9}&\bf{38.7}&\bf{52.7} \\
\rowcolor{Gray}
\textit{vs. baseline} & - & {$\downarrow$24.8M} & {+1.2}&+1.5&+0.7&+0.8 & {+0.9}&+1.2&+1.0&+0.7 \\
\shline
\hline\hline
\multicolumn{11}{c}{Cascade Mask R-CNN \cite{cai2018cascade}} \\
% \hline
% ResNet-50 & 1\(\times\) & 25.6M & 41.2&23.3&44.5&54.5 & 35.7&16.4&38.2&52.6 \\
% TiedResNet-50 & 1\(\times\) & 22.0M & \bf{43.5}&\bf{24.8}&\bf{47.3}&\bf{57.5} & \bf{37.6}&\bf{17.4}&\bf{40.4}&\bf{55.1} \\
% \textit{vs. baseline} & - & {$\downarrow$3.6M} & {+2.3}&{+1.5}&{+2.8}&{+3.0} & {+1.9}&{+1.0}&{+2.2}&{+2.5} \\
\hline
ResNet-50 & 20e & 25.6M & 42.3&23.7&45.7&56.4 & 36.6&17.3&39.0&53.9 \\
TiedResNet-50 & 20e & 22.0M & \bf{44.7}&\bf{25.8}&\bf{47.9}&\bf{59.3} & \bf{38.4}&\bf{18.7}&\bf{40.9}&\bf{56.5} \\
\rowcolor{Gray}
\textit{vs. baseline} & - & {$\downarrow$3.6M} & {+2.4}&{+2.1}&{+2.2}&{+2.9} & {+1.8}&{+1.6}&{+1.9}&{+2.6} \\
\hline
ResNet-101 & 20e & 44.6M & 43.3&24.4&46.9&58.0 & 37.6&17.3&40.4&56.2 \\
TiedResNet-101-S & 20e & 24.0M & \bf{44.5}&\bf{25.2}&\bf{48.4}&\bf{59.0} & \bf{38.6}&\bf{18.4}&\bf{41.6}&\bf{56.8} \\
\rowcolor{Gray}
\textit{vs. baseline} & - & {$\downarrow$22.6M} & {+1.2}&+0.8&+1.6&+1.0 & {+1.0}&+1.1&+1.2&+0.6 \\
TiedResNet-101 & 20e & 39.4M & \bf{45.6}&\bf{26.4}&\bf{49.5}&\bf{60.7} & \bf{39.3}&\bf{19.1}&\bf{42.4}&\bf{58.0} \\
\rowcolor{Gray}
\textit{vs. baseline} & - & {$\downarrow$5.2M} & {+2.3}&+2.0&+2.6&+2.7 & {+1.7}&+1.8&+2.0&+2.2 \\
% \hline\hline
% \multicolumn{5}{c}{Hybrid Task Cascade} \\
% \hline
% R-101 & 20e & 44.6M & 44.9&-&-&- & 39.4&-&-&- \\
% BR-101 & 20e & 39.4M & \bf{46.7}&\bf{27.6}&\bf{50.6}&\bf{61.8} & \bf{40.7}&\bf{19.1}&\bf{44.1}&\bf{59.8} \\
% \textit{vs. baseline} & - & {$\downarrow$3.6M} & {{+1.8}&-&-&-} & {{+1.3}&-&-/-} \\
\shline
\end{tabular}\vspace{0mm}
\caption{\#params of backbones vs. their Average Precision on object detection and instance segmentation tasks of MS-COCO \textit{val-2017}. LS denotes learning rate schedule. We experiment TBC on different kinds of backbone networks, including ResNet, ResNeXt and SENet. Consistent performance improvements can be obtained. \textit{Our TiedResNet-101 (with TBC) not only outperforms ResNet-101 counterpart but also achieves comparable performance with ResNeXt-101-32-$\times$8d (with Group Convolution) with only 44\% parameters.}. Results in this table was used to plot Figure 6(c) and Figure 6(d) of our submission}
\label{tab:coco_detection_instance_seg}
\end{table*}
}

\def\fighistvgg#1{
\begin{figure*}[#1]
\begin{subfigure}{.32\textwidth}
  \centering
  \includegraphics[width=.99\linewidth]{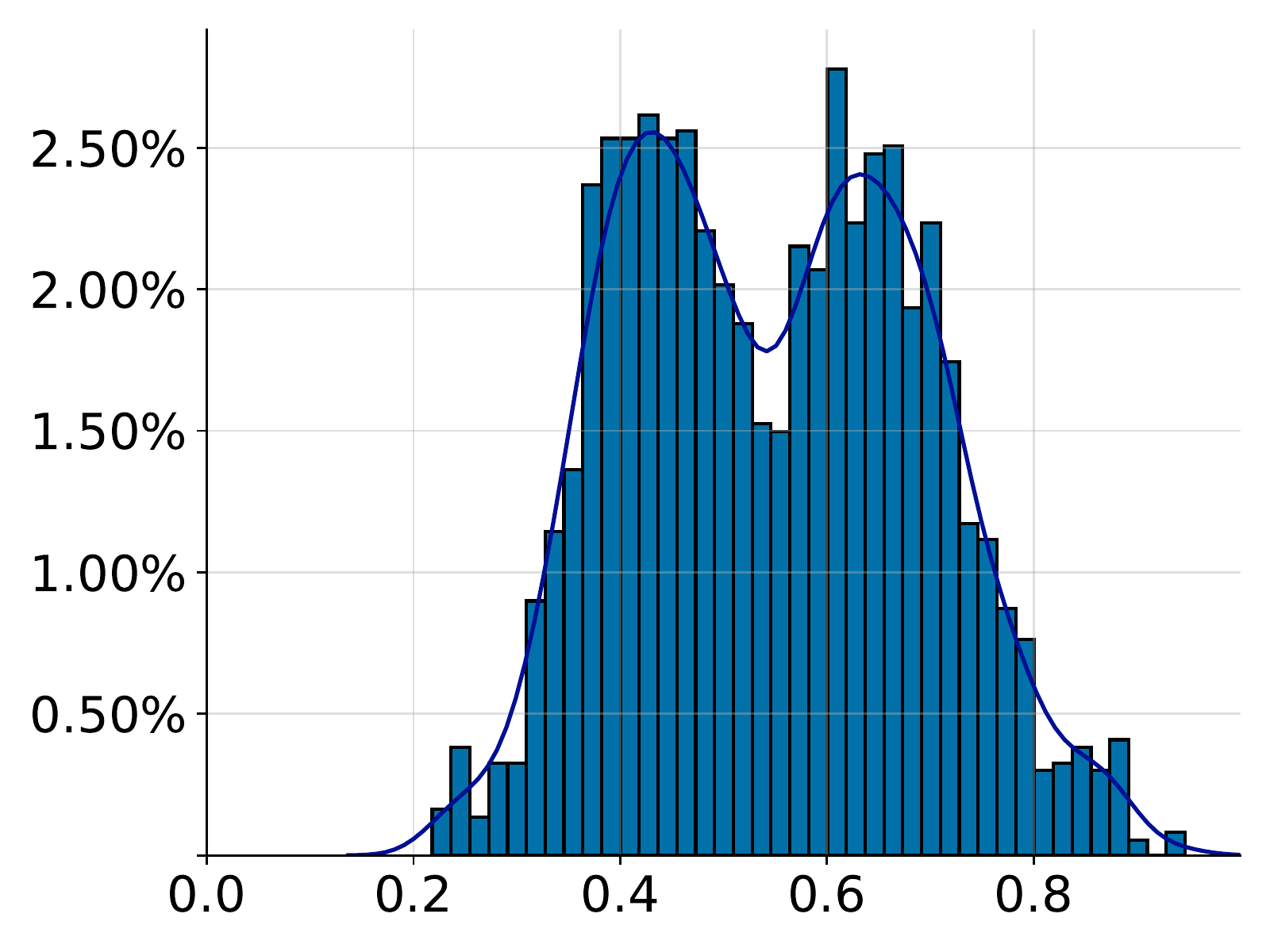}
  \caption{layer 2}
  \label{fig:sfig2}
\end{subfigure}
\begin{subfigure}{.32\textwidth}
  \centering
  \includegraphics[width=.99\linewidth]{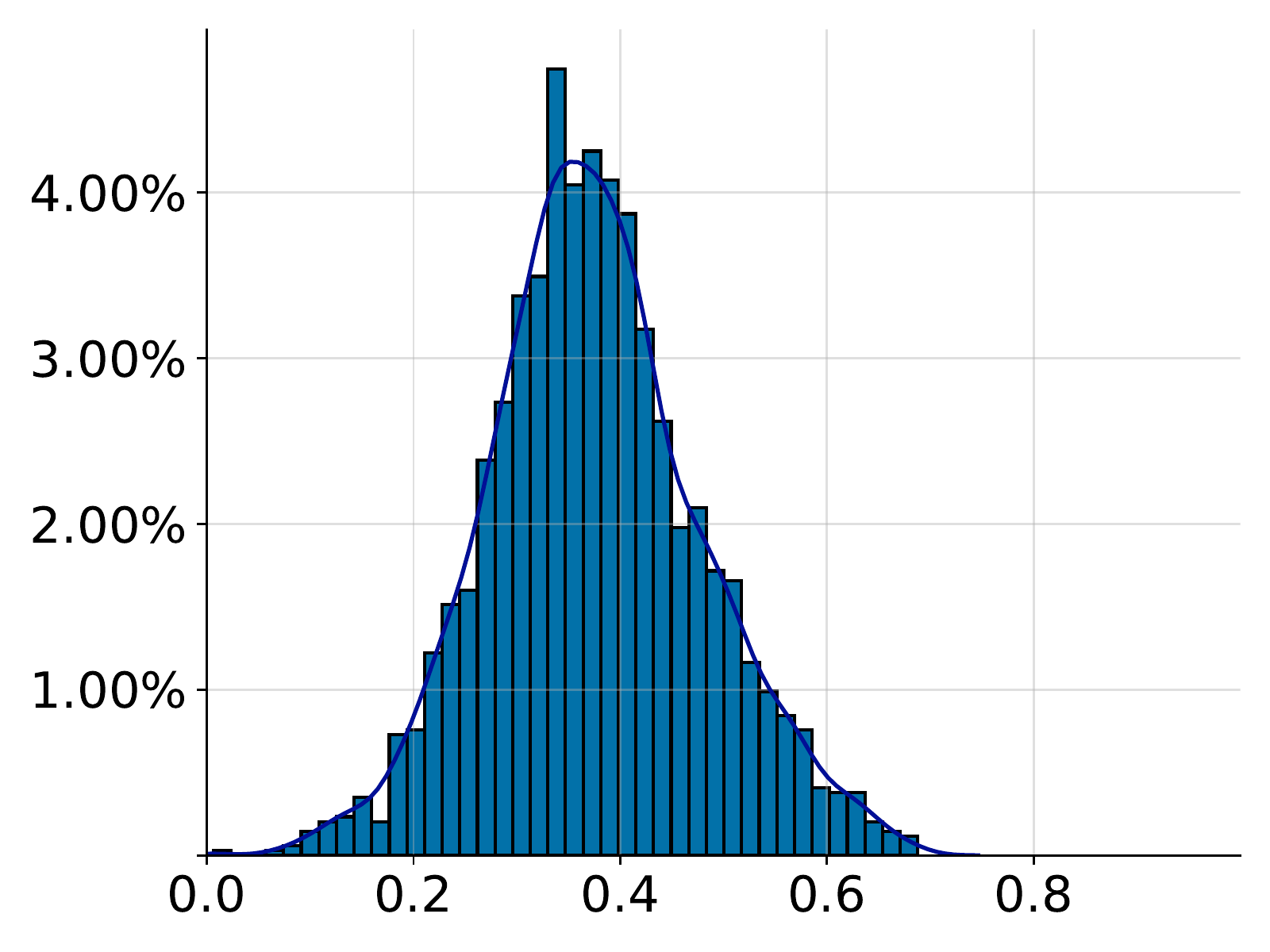}
  \caption{layer 4}
  \label{fig:sfig2}
\end{subfigure}
\begin{subfigure}{.32\textwidth}
  \centering
  \includegraphics[width=.99\linewidth]{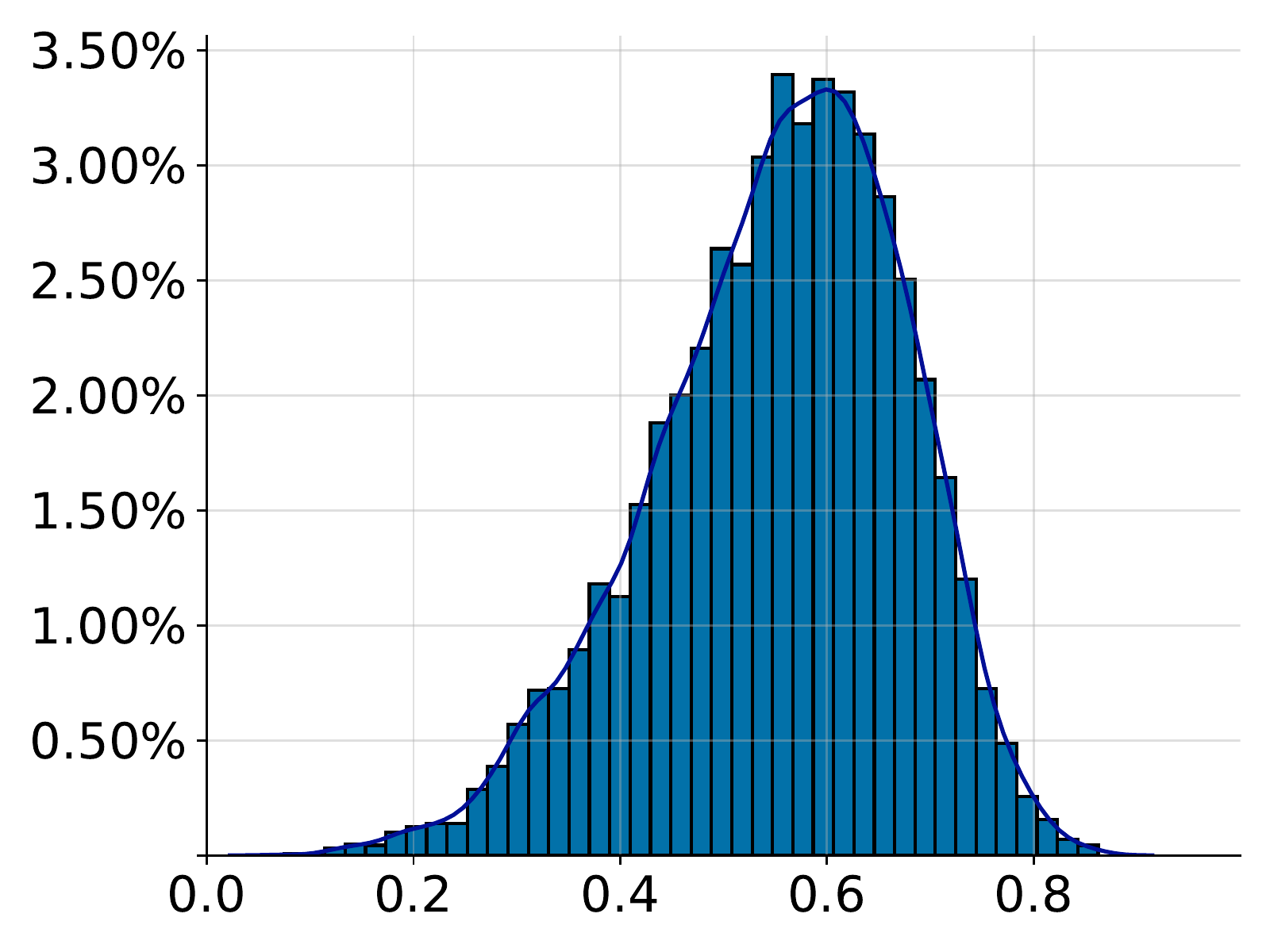}
  \caption{layer 6}
  \label{fig:sfig2}
\end{subfigure}\vspace{20pt}
\begin{subfigure}{.32\textwidth}
  \centering
  \includegraphics[width=.99\linewidth]{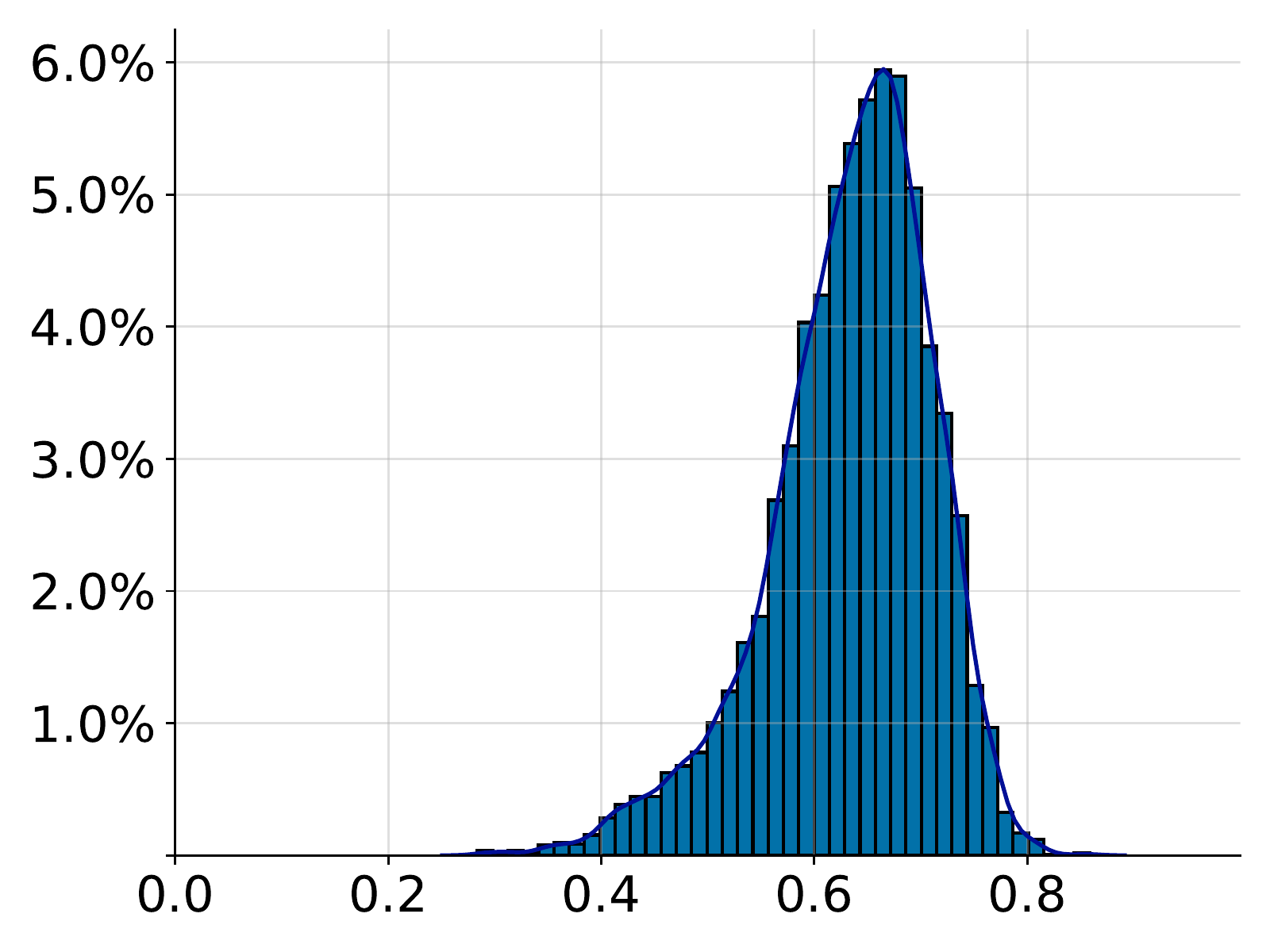}
  \caption{layer 8}
  \label{fig:sfig2}
\end{subfigure}
\begin{subfigure}{.32\textwidth}
  \centering
  \includegraphics[width=.99\linewidth]{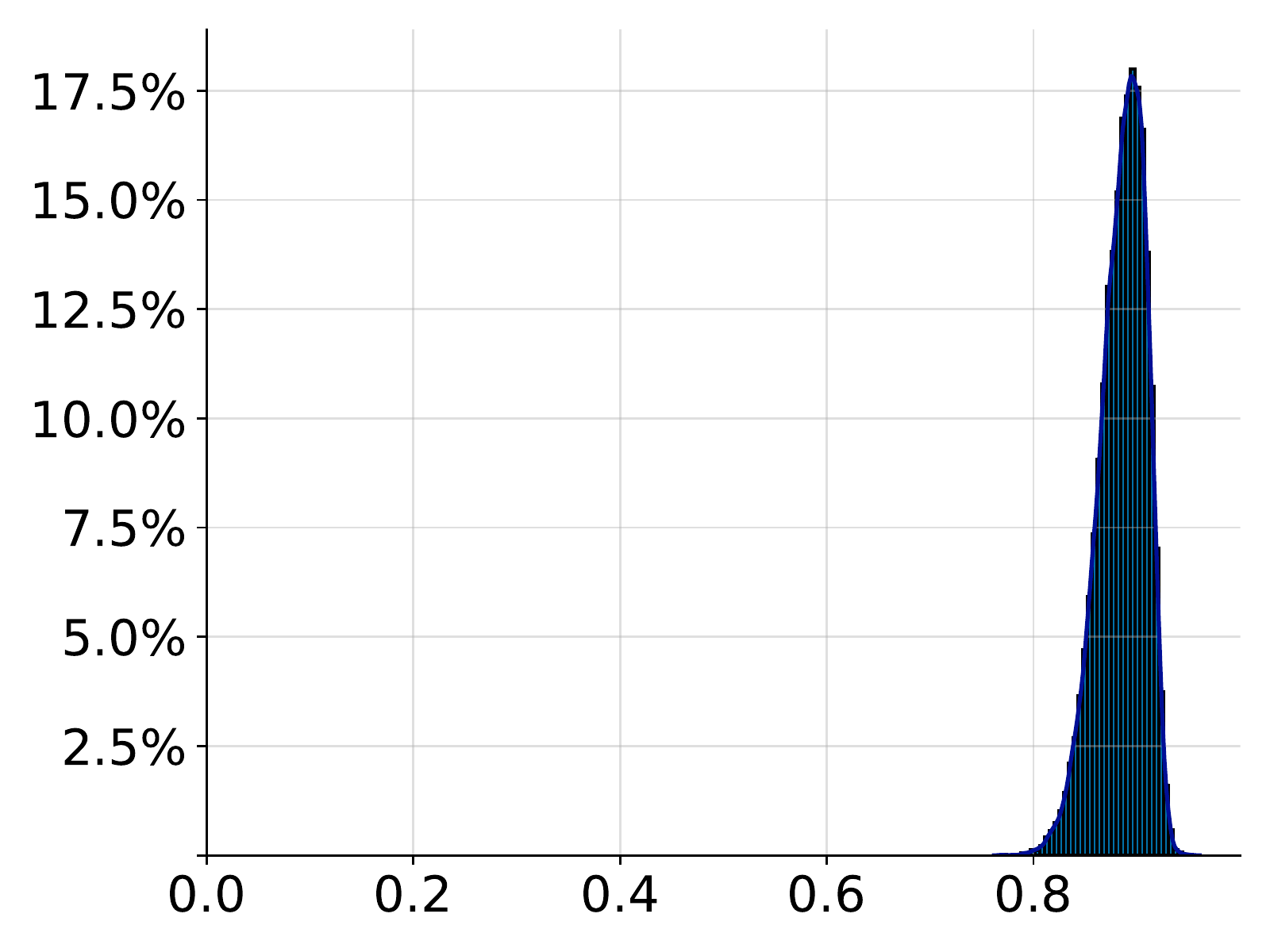}
  \caption{layer 10}
  \label{fig:sfig2}
\end{subfigure}
\begin{subfigure}{.32\textwidth}
  \centering
  \includegraphics[width=.99\linewidth]{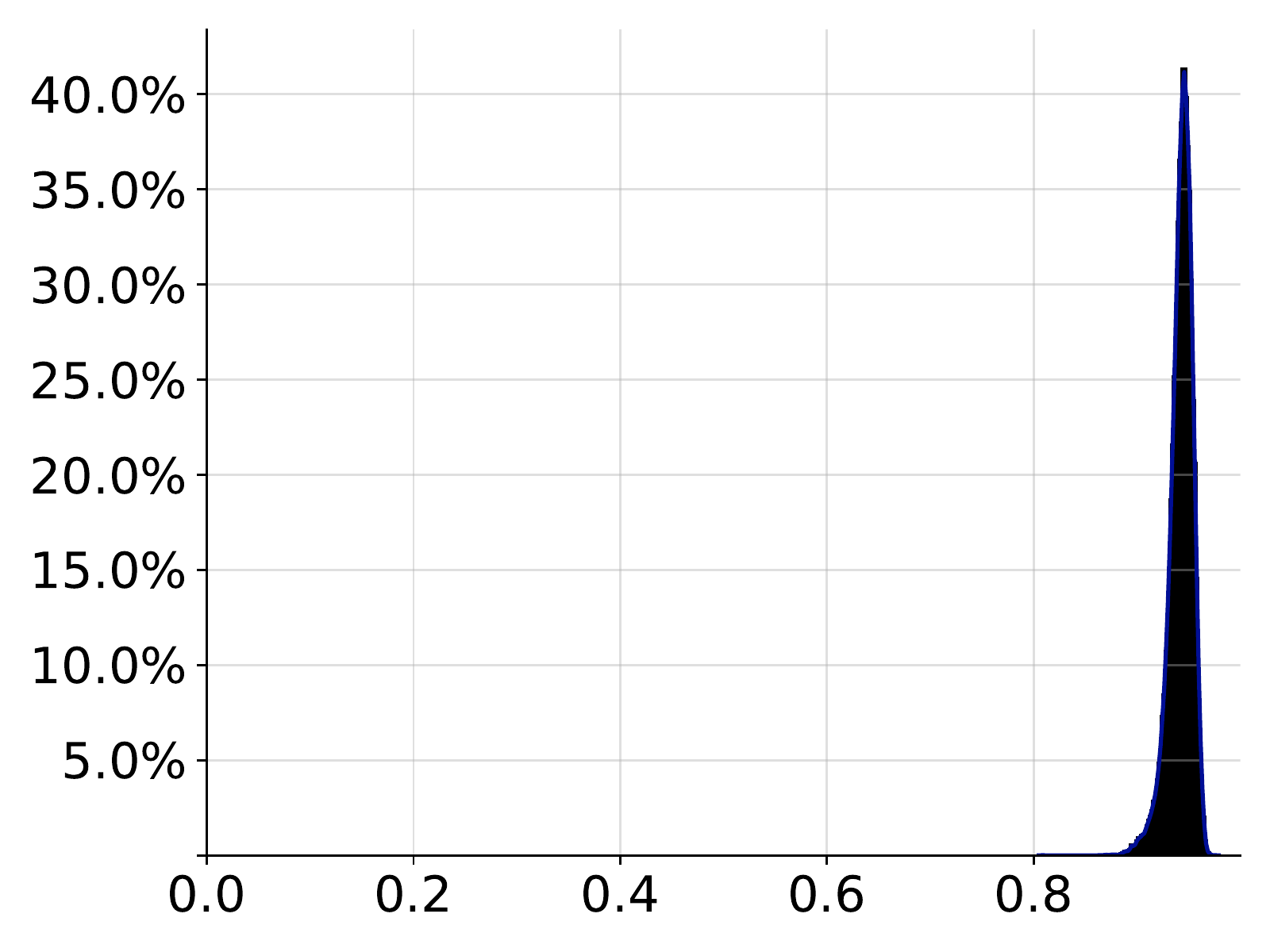}
  \caption{layer 12}
  \label{fig:sfig2}
\end{subfigure}
\caption{Histograms of pairwise filter similarities show increasing correlation among filters with depth.  At depth layer $d$ of VGG16 for ImageNet classification, we compute the similarity between two filters based on their guided back-propagation patterns on a set of images. As the number of channels increases with depth from 64 to 128 to 256, the curve shifts right and becomes far narrower, i.e., more filters become similar.}
\label{fig:histvgg}\vspace{-1pt}
\end{figure*}
}

\def\corrvgg#1{
\begin{figure*}[#1]
\begin{subfigure}{.32\textwidth}
  \centering
  \includegraphics[width=.99\linewidth]{figures/filters/corr/corr_64_2_thresh_0.06.pdf}
  \caption{layer 2}
  \label{fig:sfig2}
\end{subfigure}
\begin{subfigure}{.32\textwidth}
  \centering
  \includegraphics[width=.99\linewidth]{figures/filters/corr/corr_64_4_thresh_0.06.pdf}
  \caption{layer 4}
  \label{fig:sfig2}
\end{subfigure}
\begin{subfigure}{.32\textwidth}
  \centering
  \includegraphics[width=.99\linewidth]{figures/filters/corr&corr_64_6_thresh_0.06.pdf}
  \caption{layer 6}
  \label{fig:sfig2}
\end{subfigure}\vspace{20pt}
\begin{subfigure}{.32\textwidth}
  \centering
  \includegraphics[width=.99\linewidth]{figures/filters/corr/corr_64_8_thresh_0.06.pdf}
  \caption{layer 8}
  \label{fig:sfig2}
\end{subfigure}
\begin{subfigure}{.32\textwidth}
  \centering
  \includegraphics[width=.99\linewidth]{figures/filters/corr/corr_64_10_thresh_0.06.pdf}
  \caption{layer 10}
  \label{fig:sfig2}
\end{subfigure}
\begin{subfigure}{.32\textwidth}
  \centering
  \includegraphics[width=.99\linewidth]{figures/filters/corr/corr_64_12_thresh_0.06.pdf}
  \caption{layer 12}
  \label{fig:sfig2}
\end{subfigure}
\caption{Correlation matrix of pairwise filter similarities across 64 random filters from each layer.}
\label{fig:corrvgg}\vspace{-1pt}
\end{figure*}
}

\def\gradcam#1#2#3#4#5#6#7#8#9{
\imwh{figures/gradcam/#1/#2.png}{0.12}{0.08}&
\imwh{figures/gradcam/#1/#3.png}{0.12}{0.08}&
\imwh{figures/gradcam/#1/#4.png}{0.12}{0.08}&
\imwh{figures/gradcam/#1/#5.png}{0.12}{0.08}&
\imwh{figures/gradcam/#1/#6.png}{0.12}{0.08}&
\imwh{figures/gradcam/#1/#7.png}{0.12}{0.08}&
\imwh{figures/gradcam/#1/#8.png}{0.12}{0.08}&
\imwh{figures/gradcam/#1/#9.png}{0.12}{0.08}\\
}

\def\figGradCam#1{
\begin{figure*}[#1]
    \centering
% \tb{@{}cccccccc@{}}{0.15}{
% \gradcam{Origin}{tiger_cat}{kite}{racer}{koala}{studio_couch}{table_lamp}{television}{umbrella}
% \gradcam{ResNet}{tiger_cat}{kite}{racer}{koala}{studio_couch}{table_lamp}{television}{umbrella}
% \gradcam{ResNeXt}{tiger_cat}{kite}{racer}{koala}{studio_couch}{table_lamp}{television}{umbrella}
% \gradcam{TiedResNet}{tiger_cat}{kite}{racer}{koala}{studio_couch}{table_lamp}{television}{umbrella}
% }\vspace{18pt}

\tb{@{}cccccccc@{}}{0.15}{
\gradcam{Origin}{banana}{boxer}{kite}{zebra}{bison}{gorilla}{warthog}{umbrella}
\gradcam{ResNet}{banana}{boxer}{kite}{zebra}{bison}{gorilla}{warthog}{umbrella}
\gradcam{ResNeXt}{banana}{boxer}{kite}{zebra}{bison}{gorilla}{warthog}{umbrella}
\gradcam{TiedResNet}{banana}{boxer}{kite}{zebra}{bison}{gorilla}{warthog}{umbrella}
}\vspace{-8pt}

\caption{Additional Grad-CAM \cite{selvaraju2017grad} visualization comparison among ResNet50, ResNeXt50 and TiedResNet50 in Rows 2-4 respectively for images in Row 1.  The grad-CAM is calculated for the last convolutional output.}
\label{fig:gradcam}
\end{figure*}
}

\def\seg#1#2#3#4#5#6#7{
\imwh{figures/demo/#1/#3.#2}{0.2}{0.1}&
\imwh{figures/demo/#1/#4.#2}{0.2}{0.1}&
\imwh{figures/demo/#1/#5.#2}{0.2}{0.1}&
\imwh{figures/demo/#1/#6.#2}{0.2}{0.1}&
% \imwh{latex/figures/demo/#1/#7.#2}{0.2}{0.09}&
\imwh{figures/demo/#1/#7.#2}{0.2}{0.1}\\
}

\def\segCity#1#2#3#4#5#6#7{
\imwh{figures/demo/#1/#3.#2}{0.23}{0.1}&
\imwh{figures/demo/#1/#4.#2}{0.23}{0.1}&
\imwh{figures/demo/#1/#5.#2}{0.23}{0.1}&
\imwh{figures/demo/#1/#7.#2}{0.23}{0.1}\\
% \imwh{figures/demo/#1/#6.#2}{0.2}{0.1}&
% \imwh{latex/figures/demo/#1/#7.#2}{0.2}{0.09}&
}

\def\figSamples#1{
\begin{figure*}[#1]
    \centering
\tb{@{}cccccc@{}}{0.2}{
\seg{city}{png}{2}{9}{8}{4}{1}
\seg{city}{png}{3}{5}{6}{7}{10}

\seg{voc}{jpg}{8}{5}{9}{4}{2}
\seg{voc}{jpg}{1}{3}{6}{7}{10}
\seg{voc}{jpg}{11}{12}{13}{14}{15}

\seg{coco}{jpg}{3}{2}{13}{7}{5}
\seg{coco}{jpg}{1}{4}{6}{8}{9}
\seg{coco}{jpg}{10}{11}{0}{12}{14}
}
\caption{Sample results of object detection and instance segmentation on Cityscapes \textit{val} \citep{cordts2016cityscapes} (ROW 1, 2), Pascal VOC \textit{test-2007} \citep{Everingham15} (ROW 3, 4, 5) and MS-COCO \textit{val-2017} \citep{lin2014microsoft} (ROW 5, 6, 7) splits. We choose Mask R-CNN with TiedResNet-50 as a detection framework for Cityscapes and MS-COCO datasets and use Faster R-CNN with TiedResNet-50 as a detector for Pascal VOC dataset. All positive proposals with confidence scores greater than 0.05 are visualized here. Although many instances are highly overlapping with each other, our network can still distinguish them clearly and make high-quality bounding box proposals and segmentation masks.}
\label{fig:sample_results}
\end{figure*}
}

\setcounter{table}{0}
\setcounter{figure}{0}

% \fighistvgg{t}
% \figcorrvgg{h}

% {\twocolumn \centering \huge \bf Tied Block Convolution: \\
% Leaner and Better CNNs with Shared Thinner Filters \\
% Supplementary Materials}

\subsection{Detailed Results on Object Detection and Instance Segmentation}
Here we provide \textbf{detailed results of experimented backbones and frameworks on object detection and instance segmentation tasks} of MS-COCO \citep{lin2014microsoft} in Table \ref{table:coco_detection_sup} and Table \ref{tab:coco_detection_instance_seg}. Average Precision (AP) across IoU thresholds from 0.5 to 0.95 with an interval of 0.05 and AP under various qualities and scales are reported. All experiments are conducted on mmdetection v1.0 codebase \citep{mmdetection}.

Regardless of the type and the performance of the experimented detector, TiedResNet consistently outperforms ResNet by more than 2\%, and has a higher efficiency of parameter usage. The light version of TiedResNet even increase the performance by 1.2\% with about 2 times parameter reduction.
In addition, the improvements in detection and instance segmentation tasks (about 2.5\%) are usually higher than the improvements in recognition task (about 1.5\%). In comparison, ResNeXt's improvements in recognition and detection tasks are similar, that is, about 1.4\%.
% Although TiedResNet-\{50,101\} only achieves similar performance with ResNeXt on 
This indicates that TiedResNet is a more suitable backbone for detection and has a stronger localization capability.

% For TiedResNet on detection task, the improvements on large scale instances are generally larger than the increase on small and medium scale instances. 

\tabDetector{!b}

% \subsection{Integrating TBC/TGC/TFC with Multiple Backbones}
We also experiment our \textbf{TBC/TGC/TFC on multiple backbones}, with Mask R-CNN as a detector, to prove the effectiveness and universality of proposed operators on detection and instance segmentation tasks. All these backbones and their counterparts are pretrained on ImageNet for 100 epochs to make fair comparisons. Similar to the observation in the ImageNet recognition task, by integrating TBC/TGC/TFC into multiple backbones, consistent improvements are obtained.

\tabDetSeg{h}

\subsection{Additional Grad-CAM visualization Results}
Additional visualization results with Grad-CAM \citep{selvaraju2017grad} is shown in Figure \ref{fig:gradcam}. As an algorithm to create a high-resolution class-discriminative visualization, Grad-CAM could illustrate the network’s ability to utilize information in the target object area. Figure \ref{fig:gradcam} shows TiedResNet localizing target instances more accurately than baselines, suggesting that the performance boost in object detection and instance segmentation tasks comes from precise attention and noise reduction of irrelevant clutters. 
\figGradCam{h}

\subsection{Sample Results}
The sample results of object detection and instance segmentation tasks on multiple datasets, including Cityscapes \citep{cordts2016cityscapes}, Pascal VOC \citep{Everingham15} and MS-COCO \citep{lin2014microsoft}, are visualized in Figure \ref{fig:sample_results}. Our TiedResNet shows strong capability of handling highly overlapping instances.
\figSamples{h}